\documentclass[10pt,journal,compsoc]{IEEEtran}

\ifCLASSOPTIONcompsoc
  \usepackage[nocompress]{cite}
\else
  \usepackage{cite}
\fi

\usepackage[utf8]{inputenc}
\usepackage[T1]{fontenc}
\usepackage{hyperref}
\usepackage{url}
\usepackage{booktabs}
\usepackage{amsfonts}
\usepackage{nicefrac}
\usepackage{microtype}
\usepackage{xcolor}

\usepackage{graphicx}
\usepackage{subcaption}
\usepackage{amsmath}
\usepackage{amssymb}
\usepackage{mathtools}
\usepackage{amsthm}
\usepackage{bm}
\usepackage{dsfont}
\usepackage{colortbl}
\usepackage{multirow}
\usepackage{makecell}
\usepackage{array}
\usepackage{enumitem}
\usepackage{pifont}
\usepackage{fontawesome5}
\usepackage{tikz}
\usetikzlibrary{shapes.geometric, arrows, positioning}
\usepackage{wrapfig}

\usepackage{algorithm}
\usepackage{algorithmic}

\newcommand{\figleft}{{\em (Left)}}

\newcommand{\figright}{{\em (Right)}}

\def\figref#1{Fig.~\ref{#1}}
\def\Figref#1{Fig.~\ref{#1}}

\def\secref#1{Section~\ref{#1}}

\def\tabref#1{Table~\ref{#1}}
\def\Tabref#1{Table~\ref{#1}}

\def\1{\bm{1}}

\def\vtheta{{\bm{\theta}}}

\DeclareMathAlphabet{\mathsfit}{\encodingdefault}{\sfdefault}{m}{sl}
\SetMathAlphabet{\mathsfit}{bold}{\encodingdefault}{\sfdefault}{bx}{n}

\newcommand{\Ls}{\mathcal{L}}
\newcommand{\R}{\mathbb{R}}

\usepackage[capitalize,noabbrev]{cleveref}
\hypersetup{
    colorlinks=true,
    linkcolor=red,
    citecolor=blue,
    urlcolor=blue
}

\theoremstyle{plain}

\theoremstyle{definition}

\theoremstyle{remark}

\definecolor{best}{rgb}{1.0, 0.65, 0.65}
\definecolor{best2}{rgb}{1.0, 0.75, 0.75}
\definecolor{best3}{rgb}{1.0, 0.90, 0.90}
\definecolor{bestlora}{rgb}{1.0, 0.80, 0.80}

\newcommand{\teaserbox}[3]{%
    \begin{tikzpicture}
        \node[inner sep=0pt] (img) {\includegraphics[width=0.14\textwidth]{#1}};
        \draw[red, thick] ([xshift=#2, yshift=#3]img.center) ++(-0.3cm, -0.3cm) rectangle ++(0.6cm, 0.6cm);
    \end{tikzpicture}%
}

\title{TransNormal-2: Geometry-Grounded Rectified Flow with Edge-Aware Decoding for Precise Normal Estimation}

\author{Mingwei~Li,
        Yi~Yang,~\IEEEmembership{Fellow,~IEEE},
        and~Hehe~Fan,~\IEEEmembership{Senior~Member,~IEEE}%
\IEEEcompsocitemizethanks{%
\IEEEcompsocthanksitem M.~Li, Y.~Yang, and H.~Fan are with the College of Artificial Intelligence, Zhejiang University. (Email: \{mingweili, yangyics, hehefan\}@zju.edu.cn)
\IEEEcompsocthanksitem M.~Li is also with Zhongguancun Academy, Beijing 100190, China.
\IEEEcompsocthanksitem This article substantially extends the authors' prior conference paper, \emph{TransNormal: Dense Visual Semantics for Diffusion-Based Transparent Object Normal Estimation}, presented at ICML 2026~\cite{li2026transnormal}. The present work reformulates the model on FLUX.2, introduces the VAE reconstruction-degradation analysis, geometry-aware pixel-space objectives, and the GRM, and broadens evaluation from transparent objects to both general and transparent scenes.
\IEEEcompsocthanksitem Corresponding author: Hehe Fan (e-mail: hehefan@zju.edu.cn).}}

\IEEEtitleabstractindextext{%
\begin{abstract}
  Diffusion-based models enable monocular geometry estimation, yet their pixel-space precision is limited by a shared, under-studied error source: \emph{VAE reconstruction degradation}.  The $8{\times}$ spatial compression in the VAE encoder-decoder degrades surface normals at object boundaries; even encoding and decoding \emph{ground-truth} normals introduces 1.3--8.5$^\circ$ of mean angular error (MAE), with edge MAE reaching $2.8{\times}$ the global MAE.  We present \textbf{TransNormal-2}, a FLUX.2-based rectified-flow framework with single-step deterministic inference that addresses this degradation on both sides of the VAE decoder: in how latent predictions are supervised during training, and in how decoded normals are corrected at inference.  First, geometry-aware pixel-space losses, including inverse rendering self-consistency, von~Mises-Fisher angular loss, and wavelet edge-aware regularization, complement latent MSE by enforcing spherical normal geometry and diffuse image-formation cues after VAE decoding.  Second, a lightweight \textbf{Geometric Refinement Module (GRM)} applies an RGB-guided residual correction to reduce boundary-localized decoding errors without freely rewriting the coarse prediction.  On general-scene benchmarks, TransNormal-2 matches or exceeds MoGe-2 on all eight reported metrics while using only $1.4\%$ as many task-specific normal annotations.  The gains are clearest for transparent objects, reducing MAE by $4.2^\circ$ on ClearGrasp and $3.1^\circ$ on ClearPose over the strongest prior baselines.  Code will be released at \url{https://longxiang-ai.github.io/TransNormal-2}.
\end{abstract}

}

\hypersetup{
    pdftitle={TransNormal-2: Geometry-Grounded Rectified Flow with Edge-Aware Decoding for Precise Normal Estimation},
    pdfauthor={Mingwei Li, Yi Yang, Hehe Fan},
    pdfsubject={Monocular surface normal estimation},
    pdfkeywords={Surface normal estimation, rectified flow, computer vision, transparent objects}
}

\begin{document}

\maketitle
\IEEEdisplaynontitleabstractindextext
\IEEEpeerreviewmaketitle

\begin{figure*}[!t]
    \centering
    \setlength{\tabcolsep}{0pt}
    \newcommand{\imgwt}{0.110\textwidth}
    \renewcommand{\teaserbox}[3]{%
        \begin{tikzpicture}
            \node[inner sep=0pt] (img) {\includegraphics[width=\imgwt]{#1}};
            \draw[red, thick] ([xshift=#2, yshift=#3]img.center) ++(-0.3cm, -0.3cm) rectangle ++(0.6cm, 0.6cm);
        \end{tikzpicture}%
    }
    \begin{tabular}{@{}c@{\hspace{0.5pt}}c@{\hspace{0.5pt}}c@{\hspace{0.5pt}}c@{\hspace{0.5pt}}c@{\hspace{0.5pt}}c@{\hspace{0.5pt}}c@{\hspace{0.5pt}}c@{\hspace{0.5pt}}c@{}}
        \includegraphics[width=\imgwt]{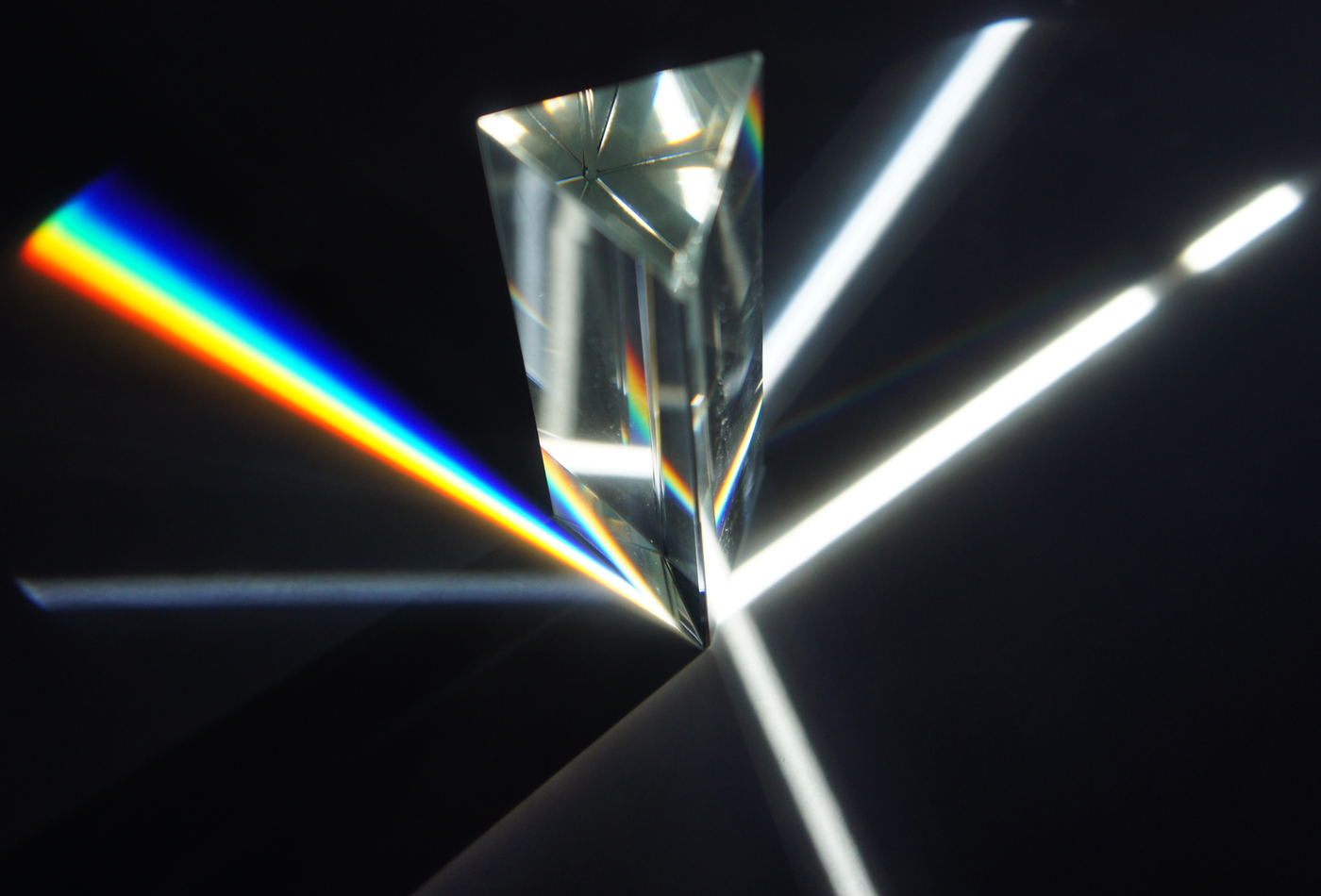} &
        \teaserbox{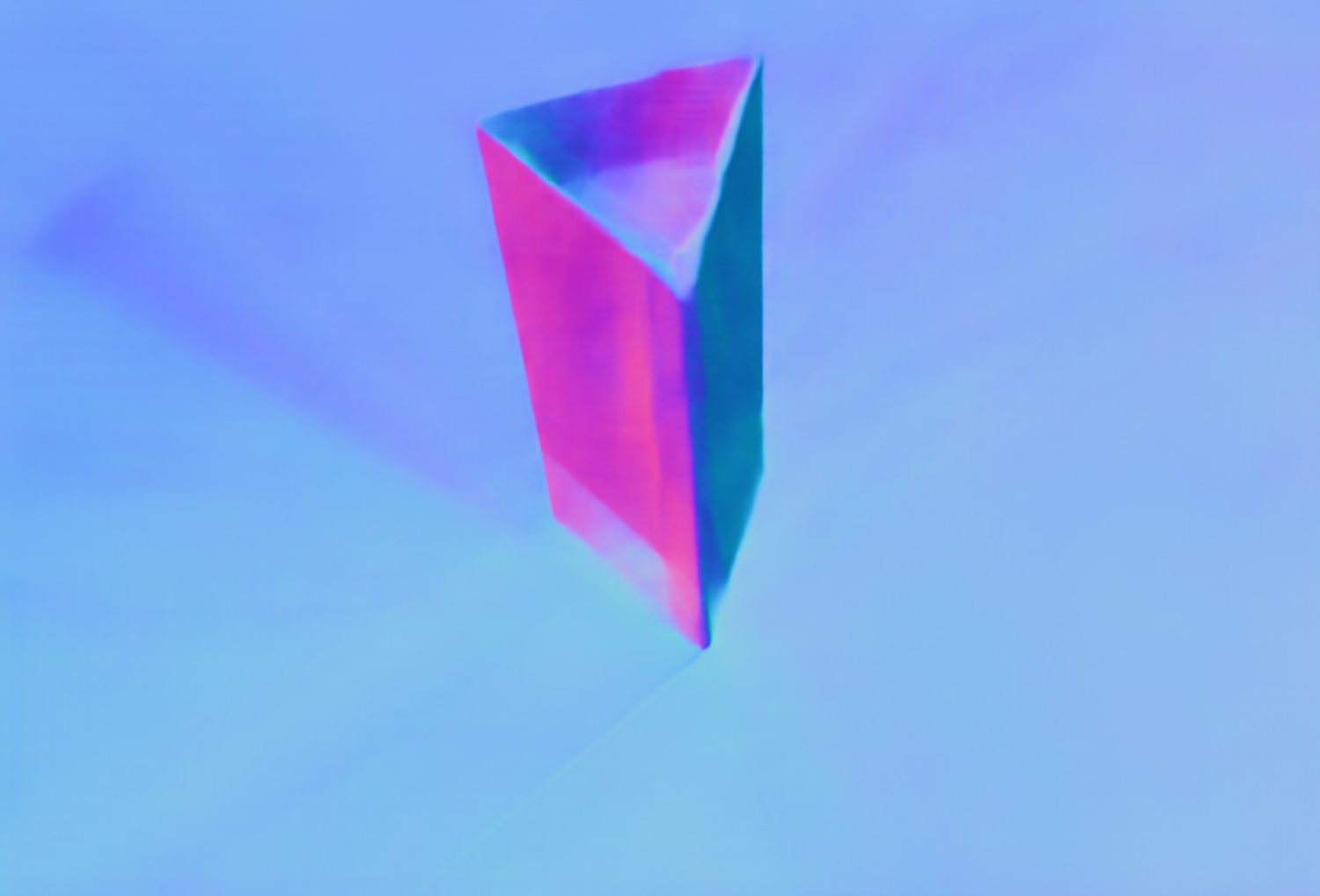}{-0.025cm}{0.35cm} &
        \teaserbox{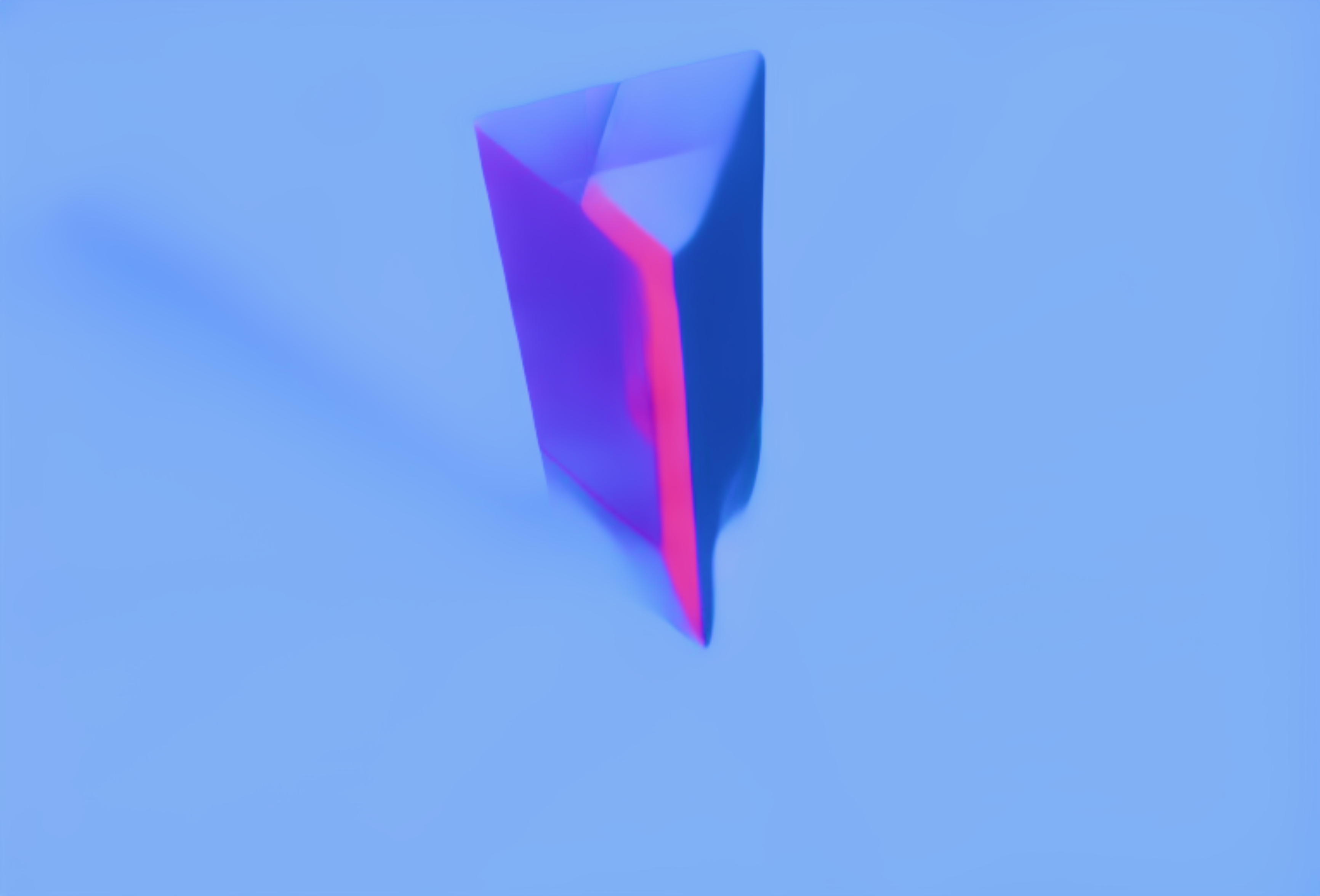}{-0.025cm}{0.35cm} &
        \teaserbox{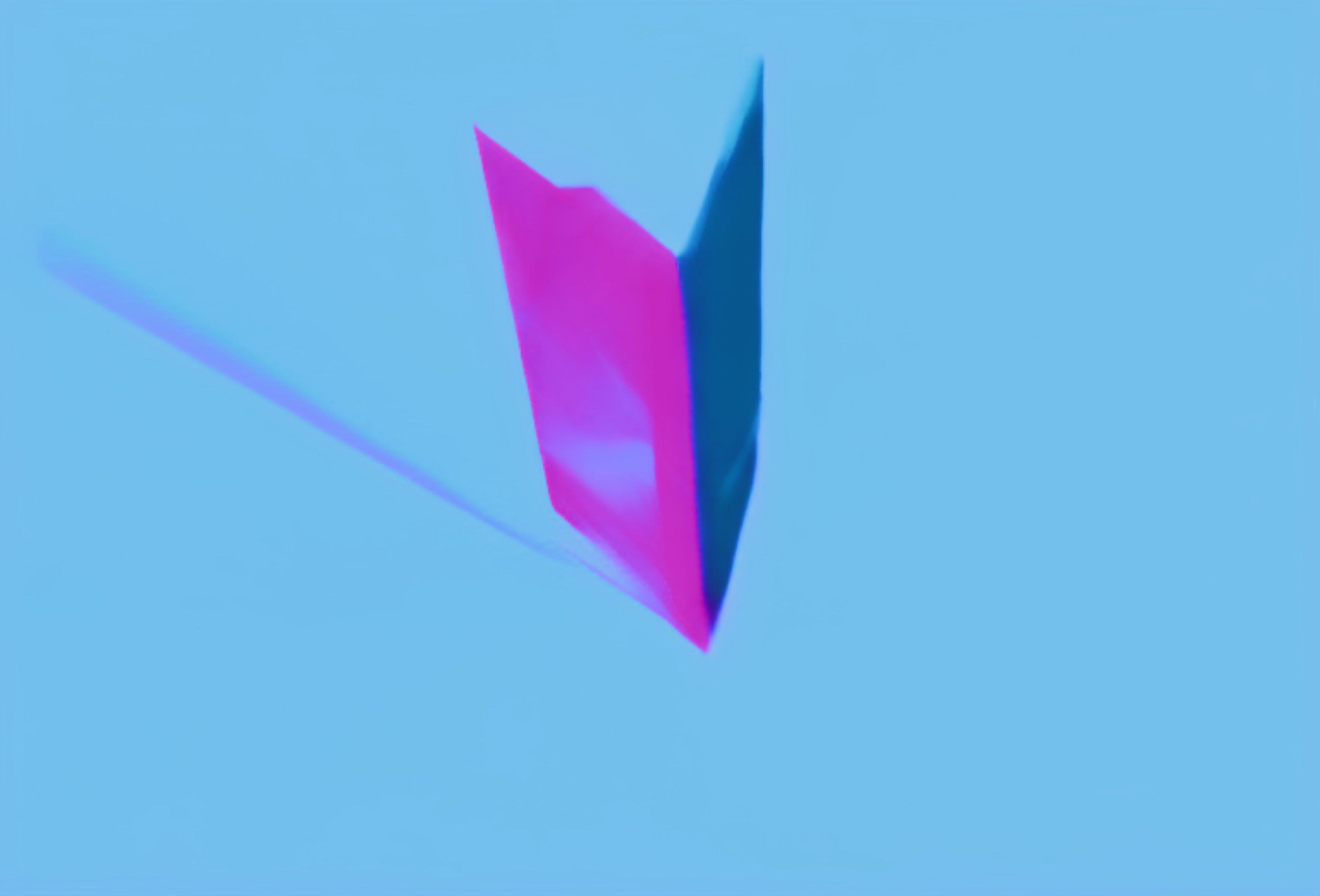}{-0.025cm}{0.35cm} &
        \teaserbox{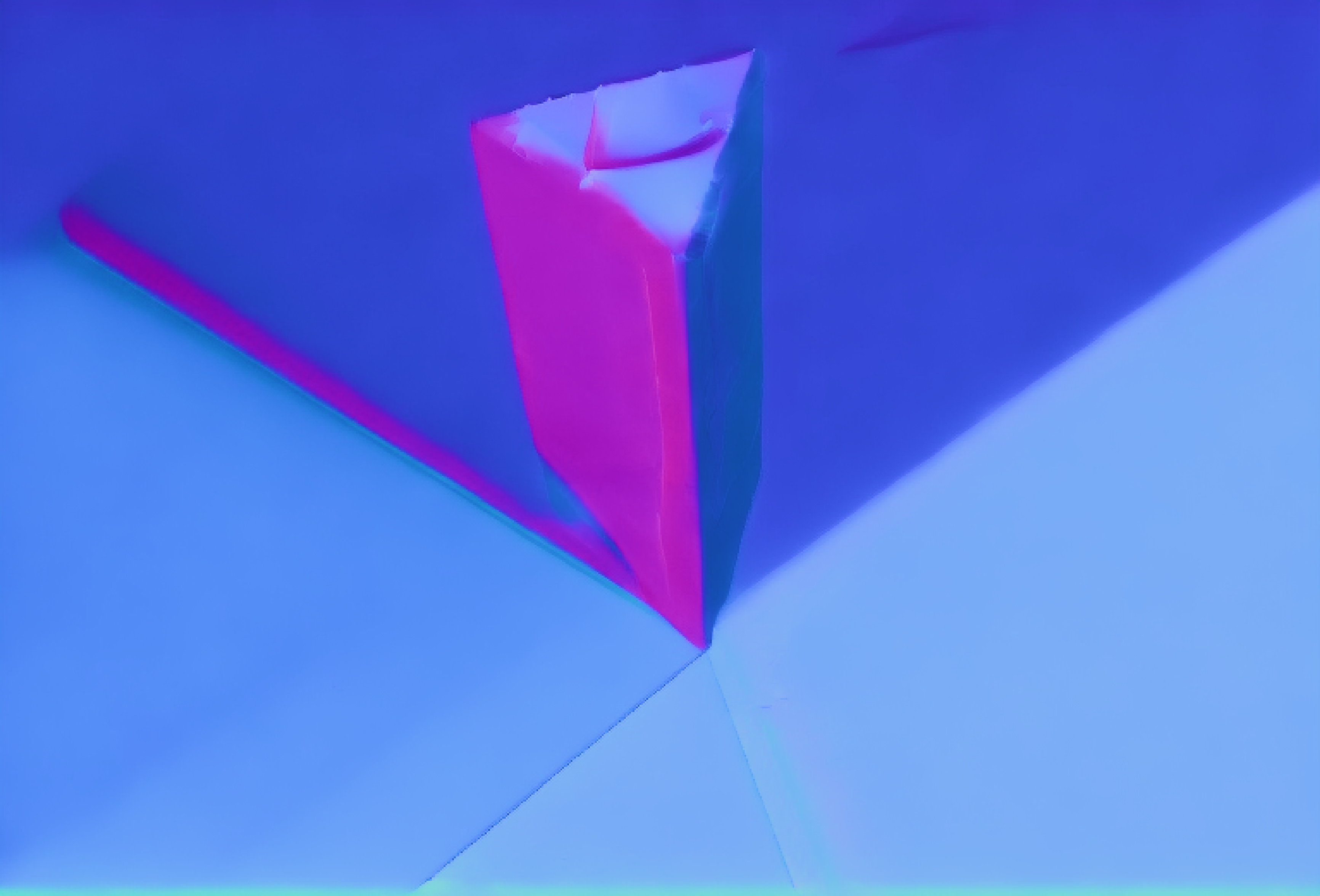}{-0.025cm}{0.35cm} &
        \teaserbox{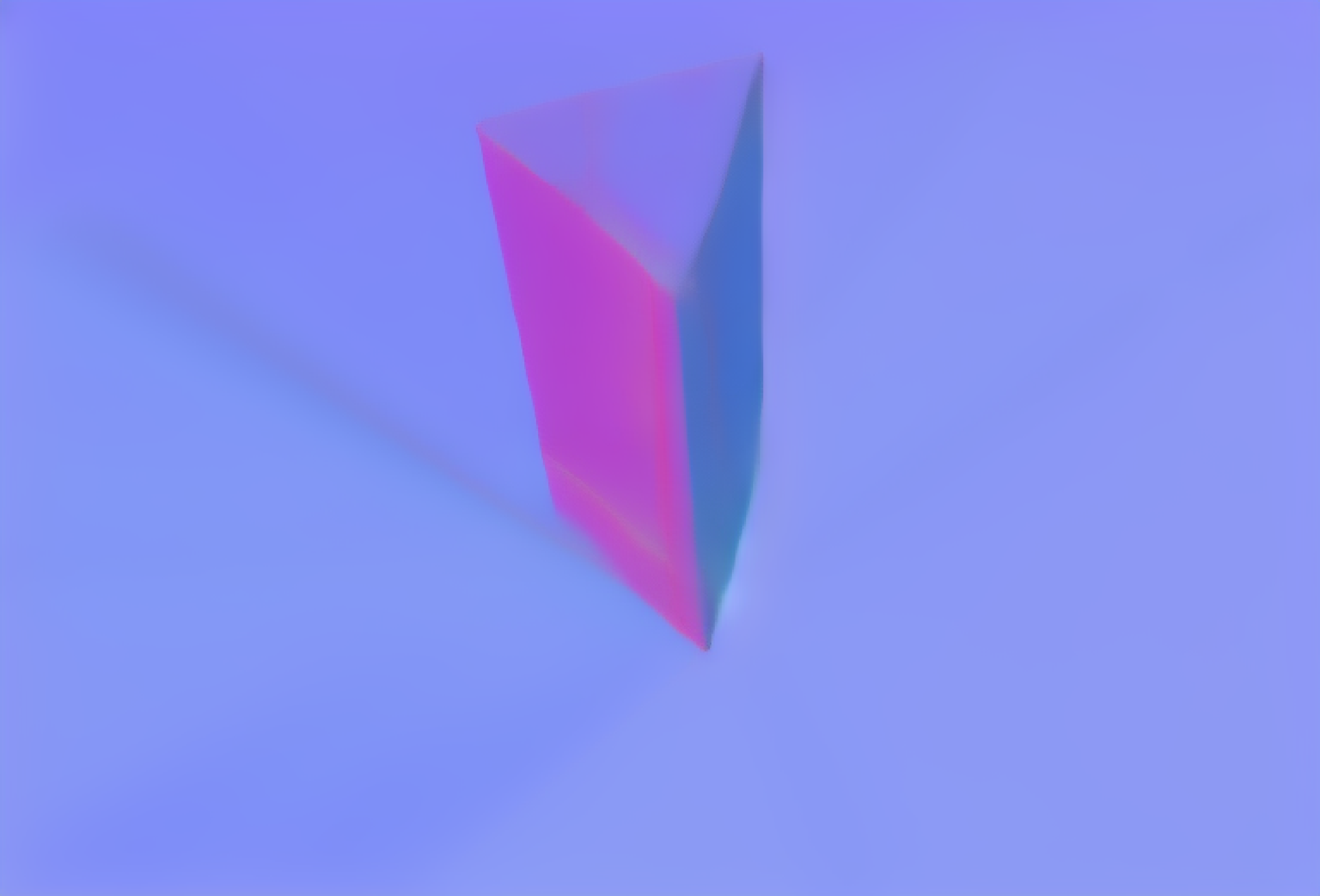}{-0.025cm}{0.35cm} &
        \teaserbox{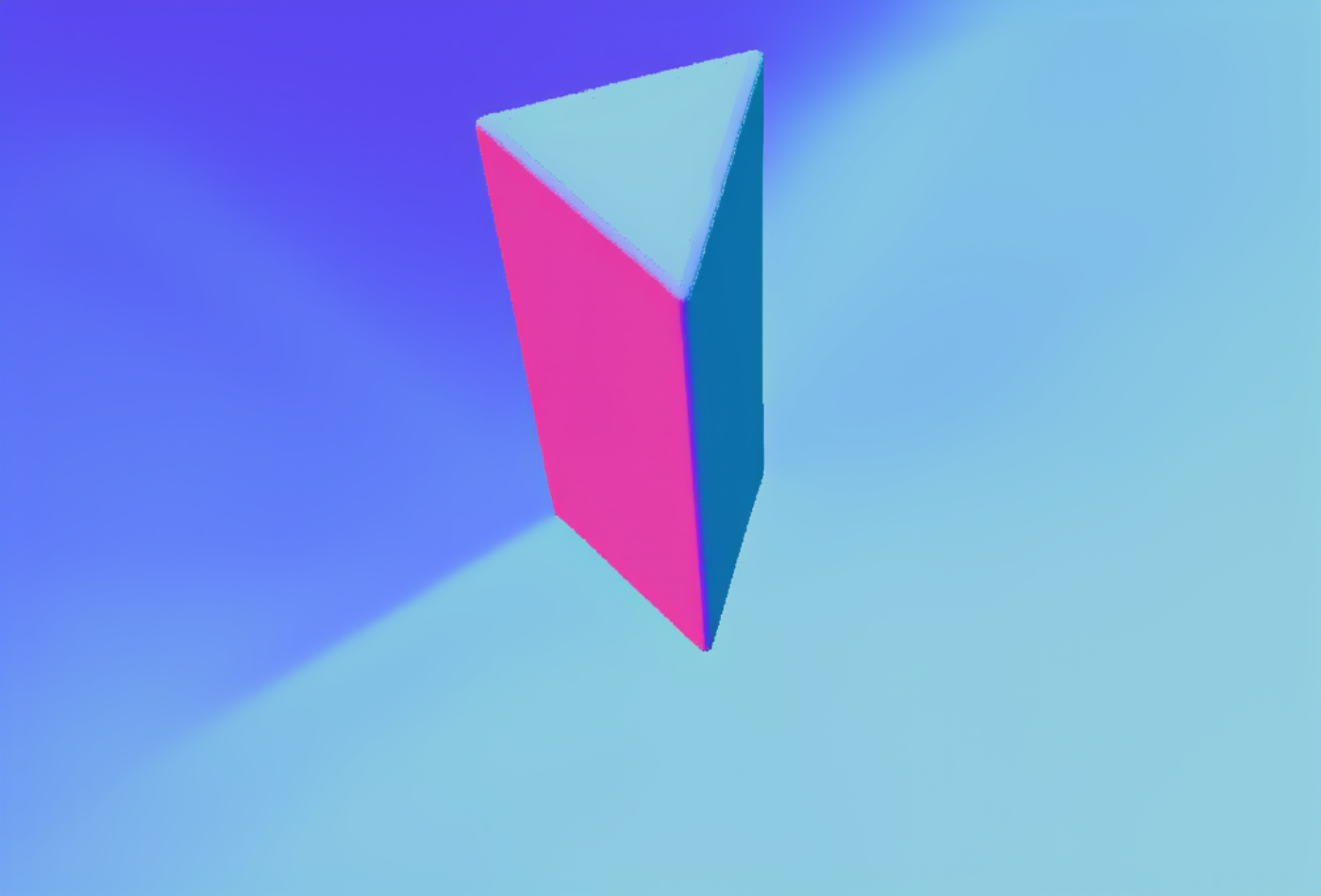}{-0.025cm}{0.35cm} &
        \teaserbox{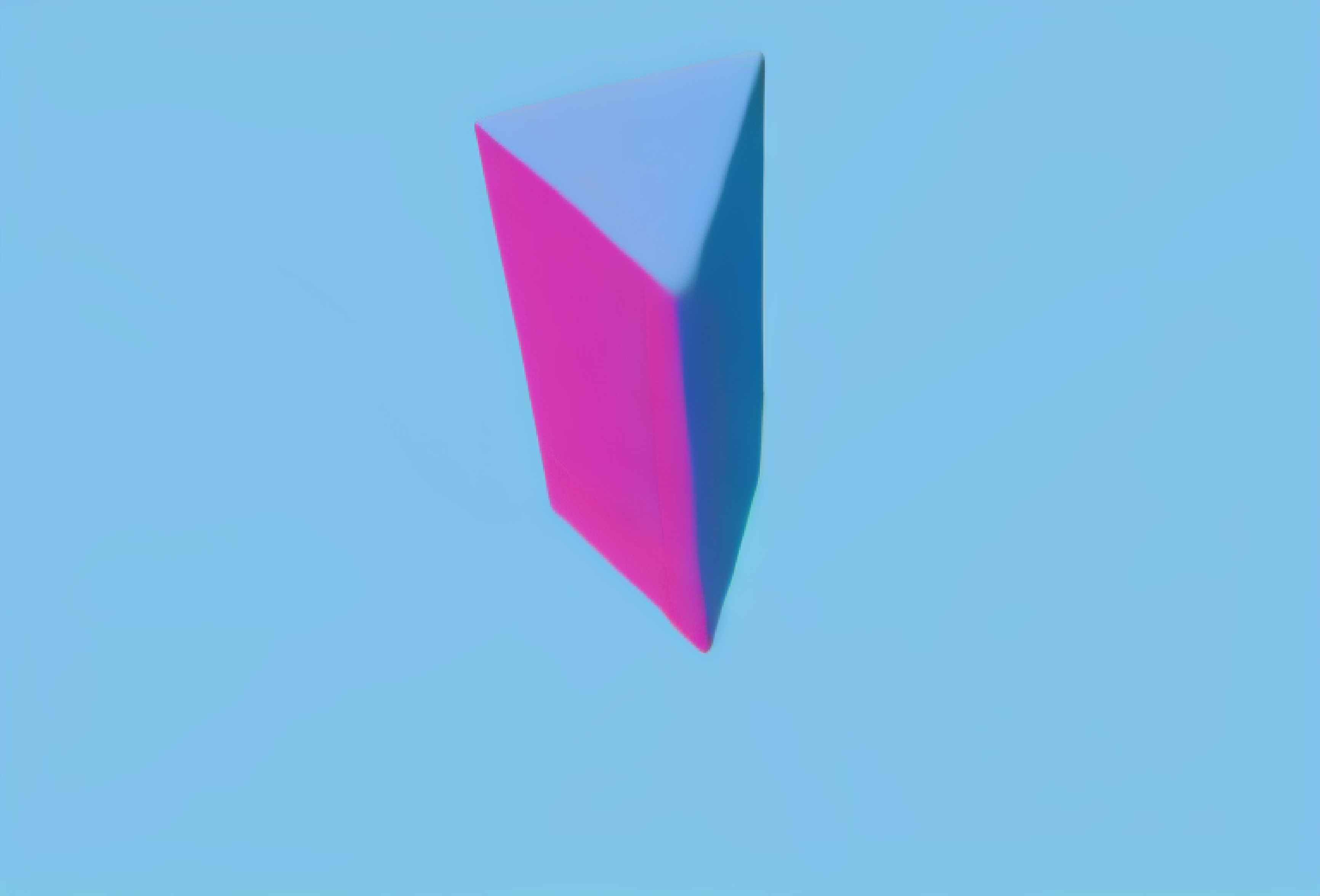}{-0.025cm}{0.35cm} &
        \teaserbox{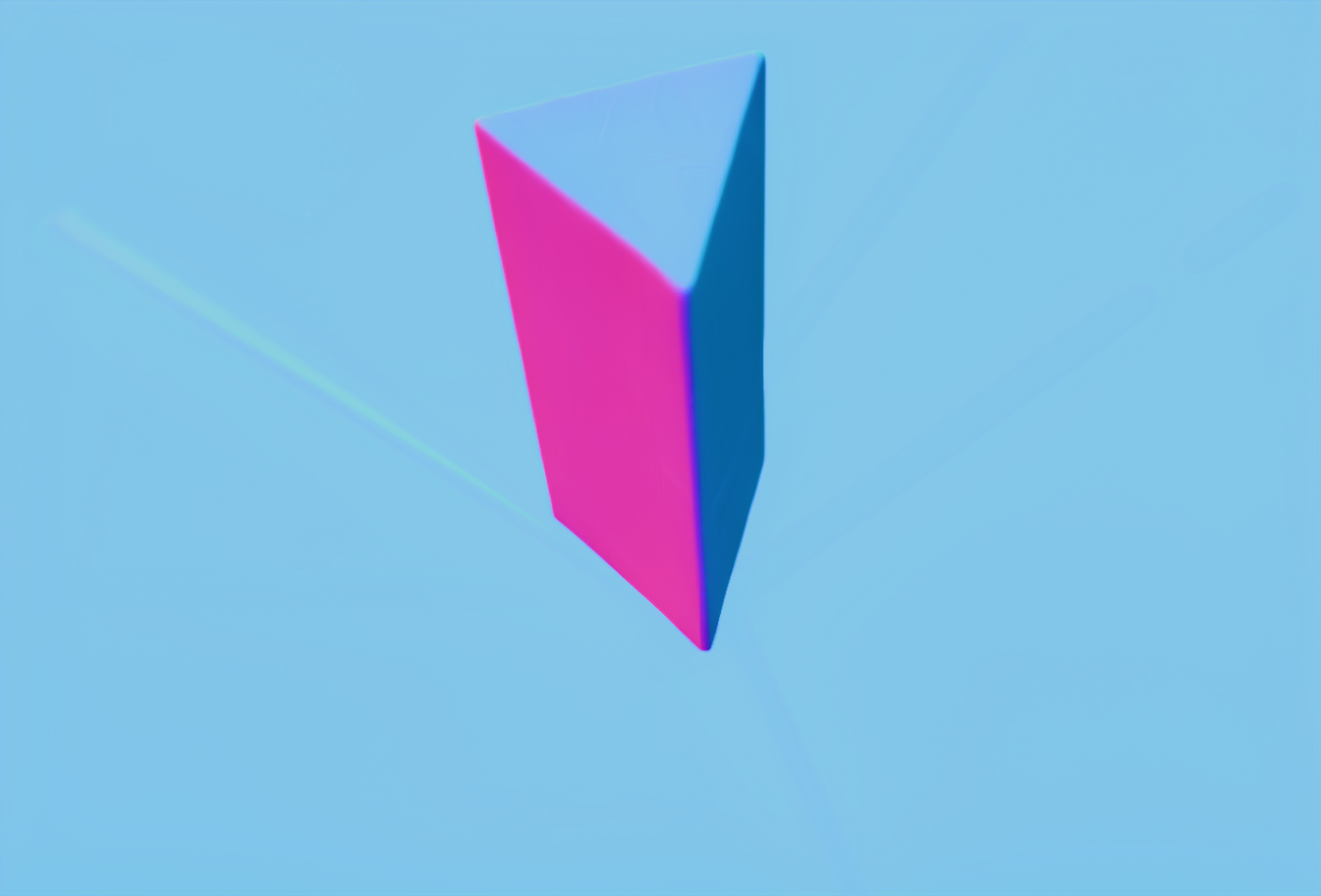}{-0.025cm}{0.35cm} \\[1pt]
        \includegraphics[width=\imgwt]{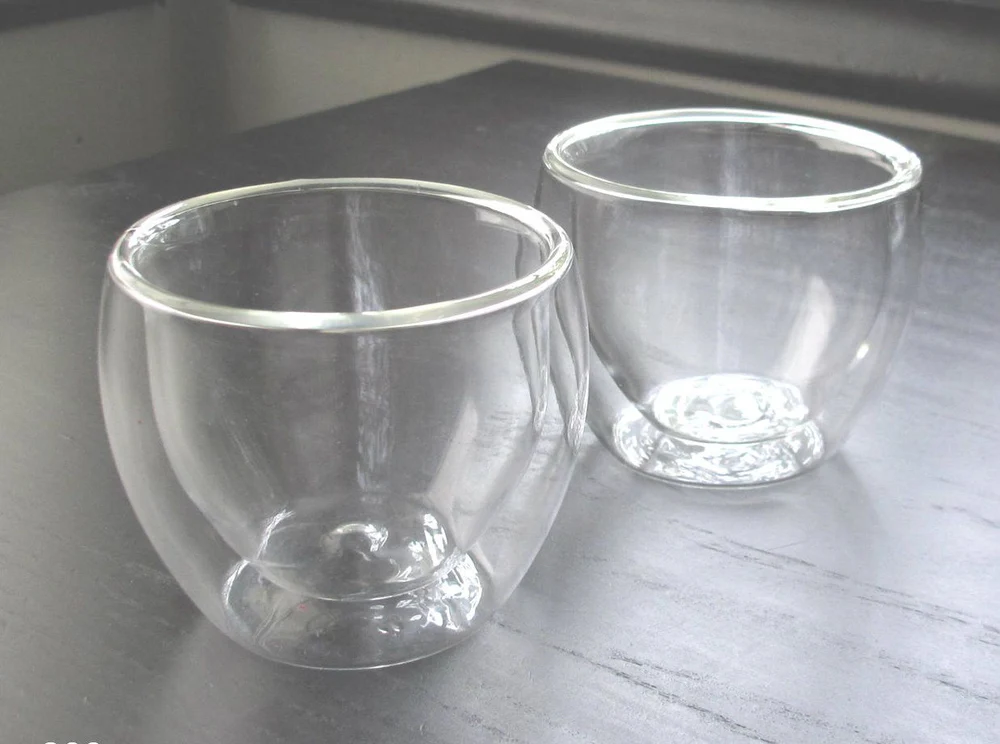} &
        \teaserbox{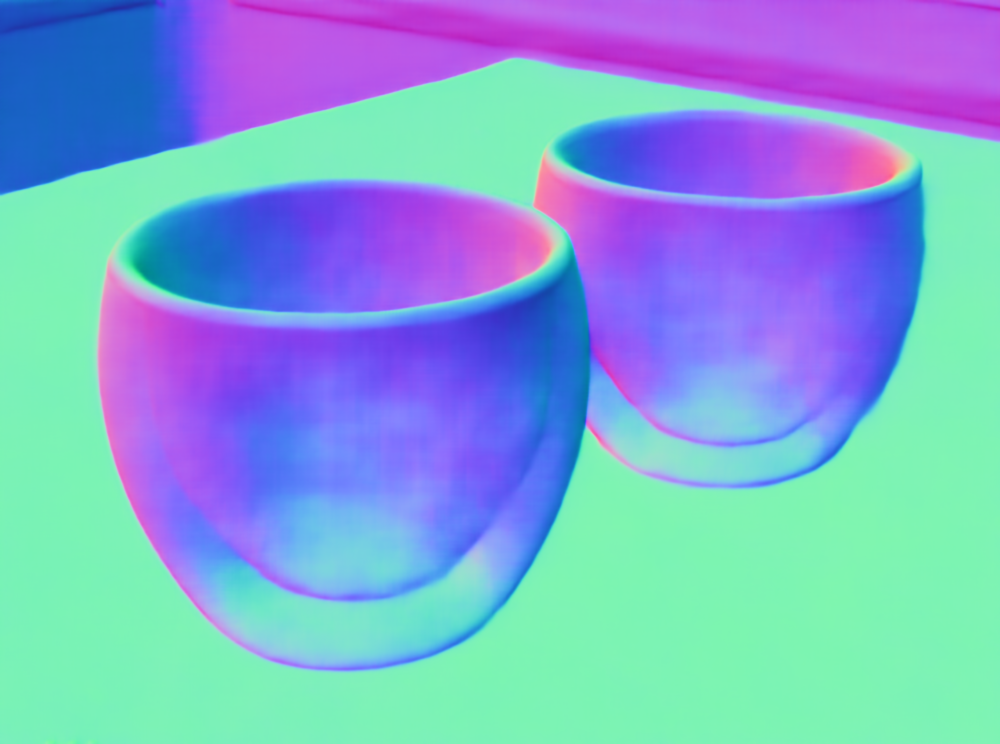}{-0.3cm}{-0.25cm} &
        \teaserbox{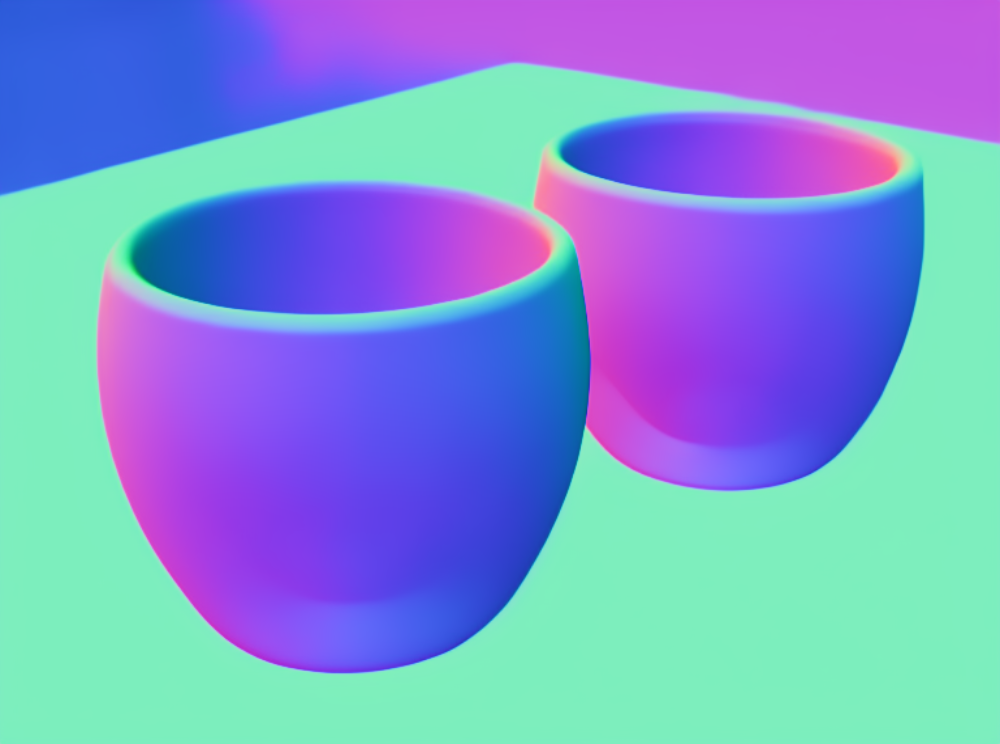}{-0.3cm}{-0.25cm} &
        \teaserbox{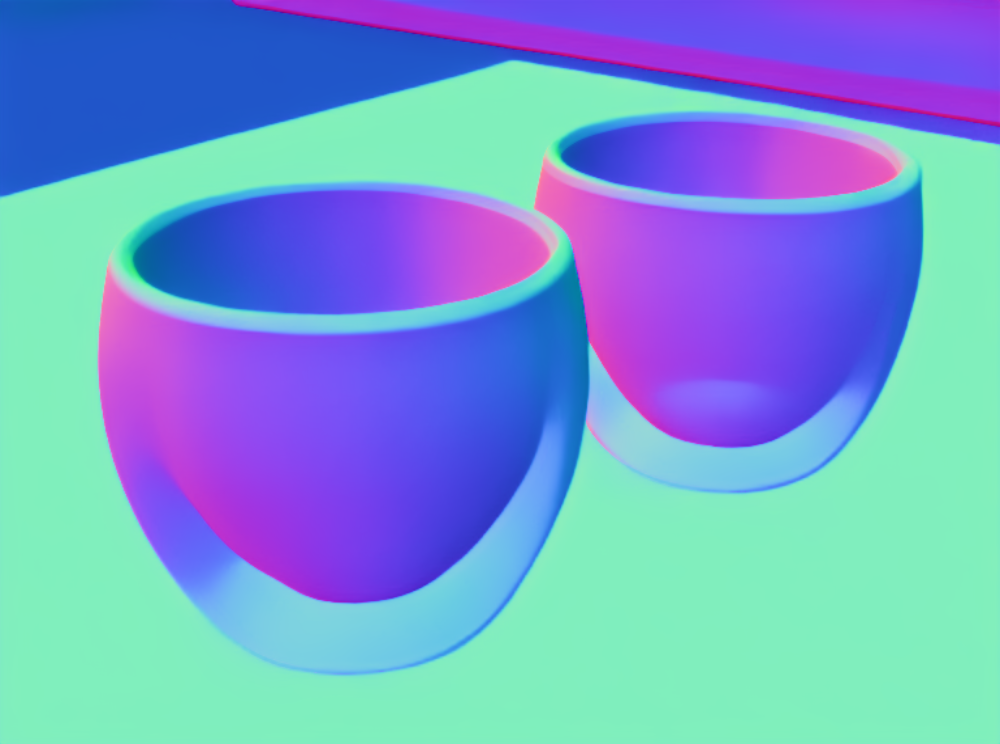}{-0.3cm}{-0.25cm} &
        \teaserbox{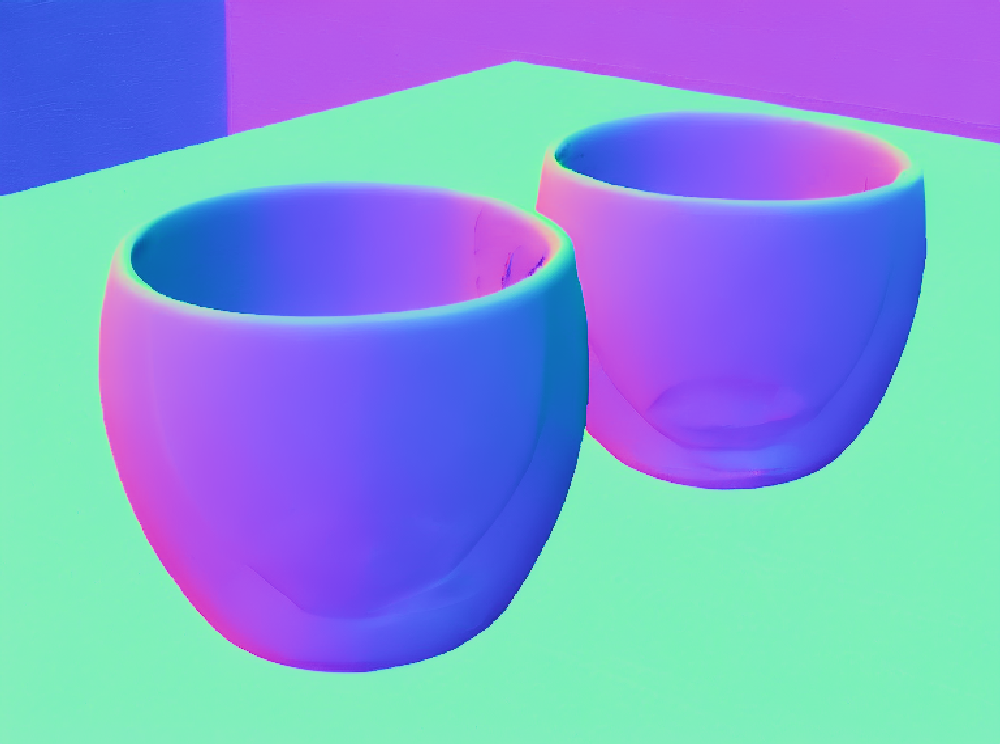}{-0.3cm}{-0.25cm} &
        \teaserbox{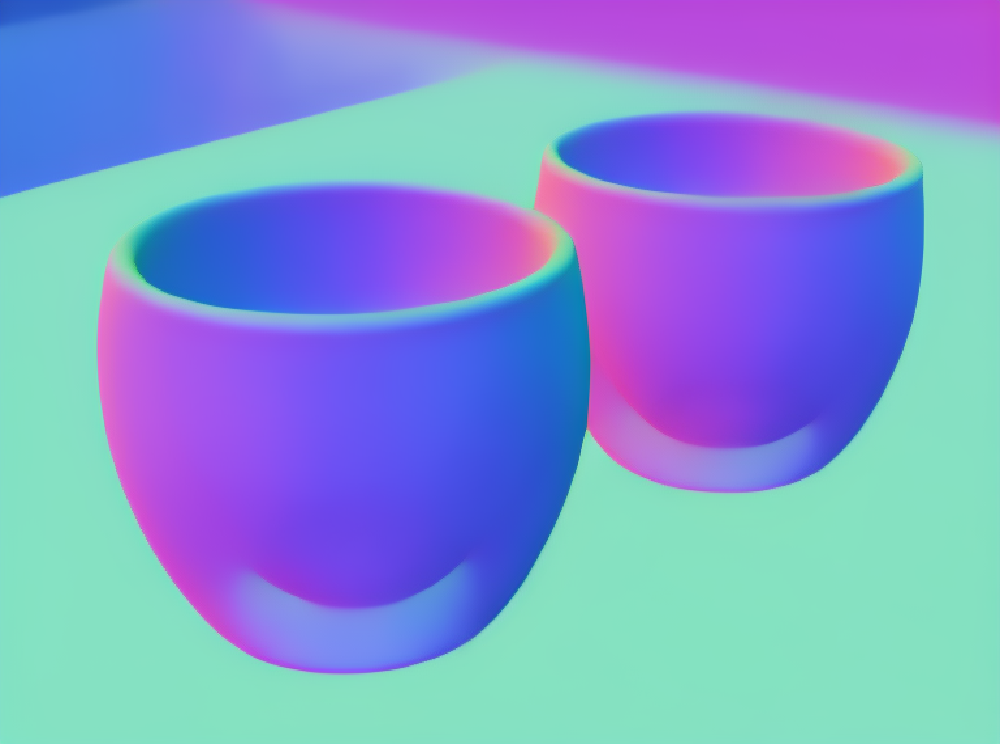}{-0.3cm}{-0.25cm} &
        \teaserbox{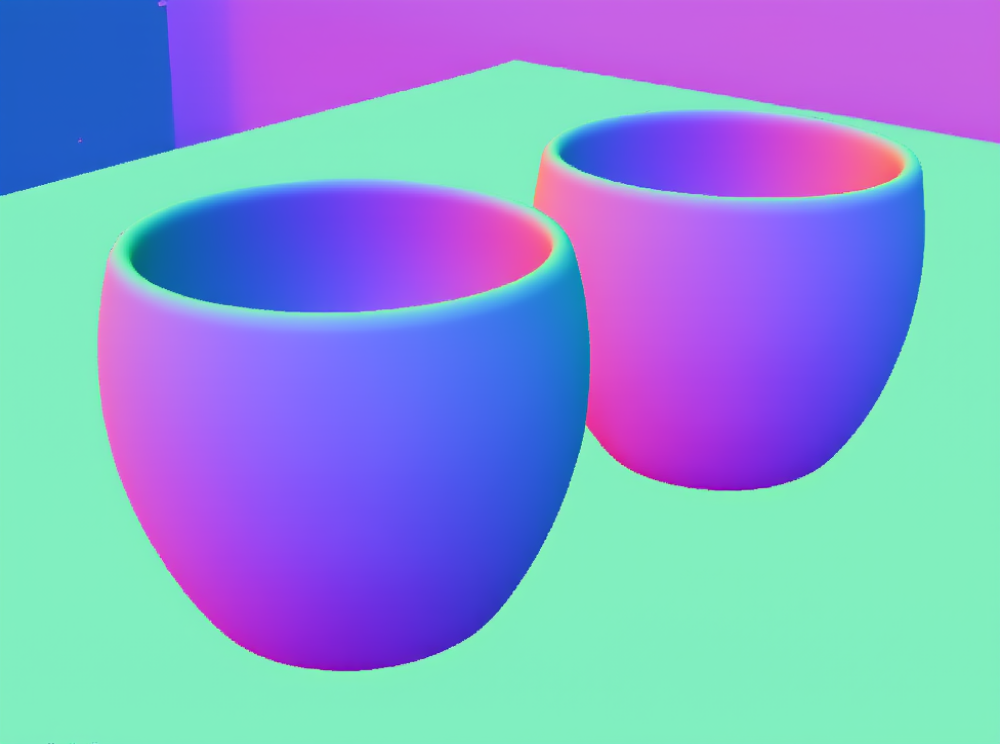}{-0.3cm}{-0.25cm} &
        \teaserbox{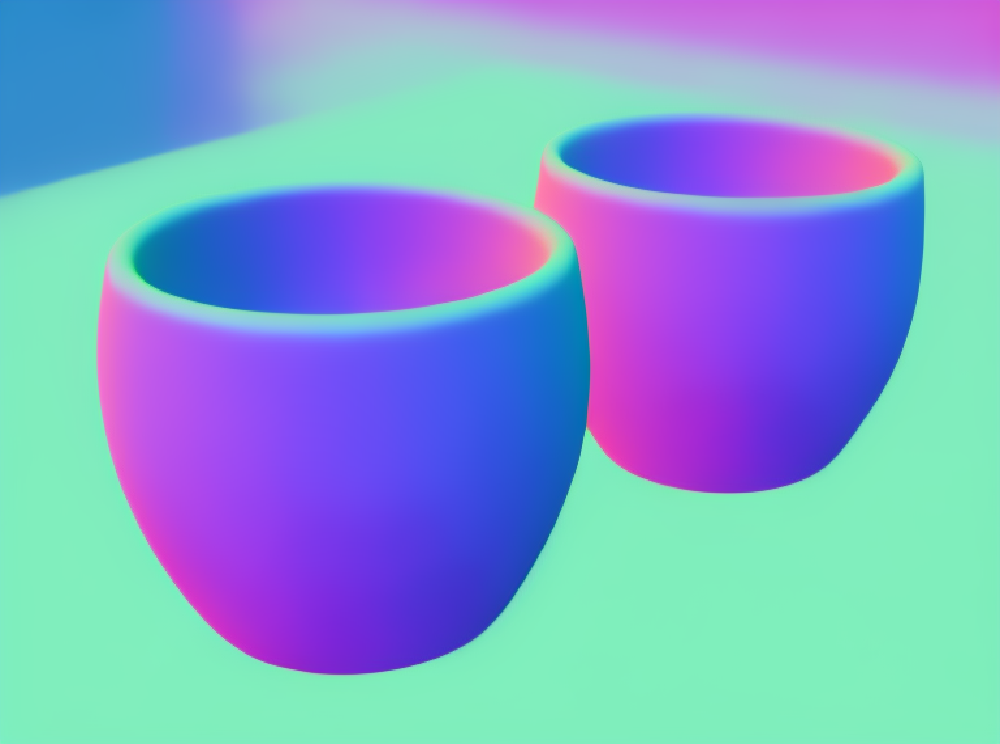}{-0.3cm}{-0.25cm} &
        \teaserbox{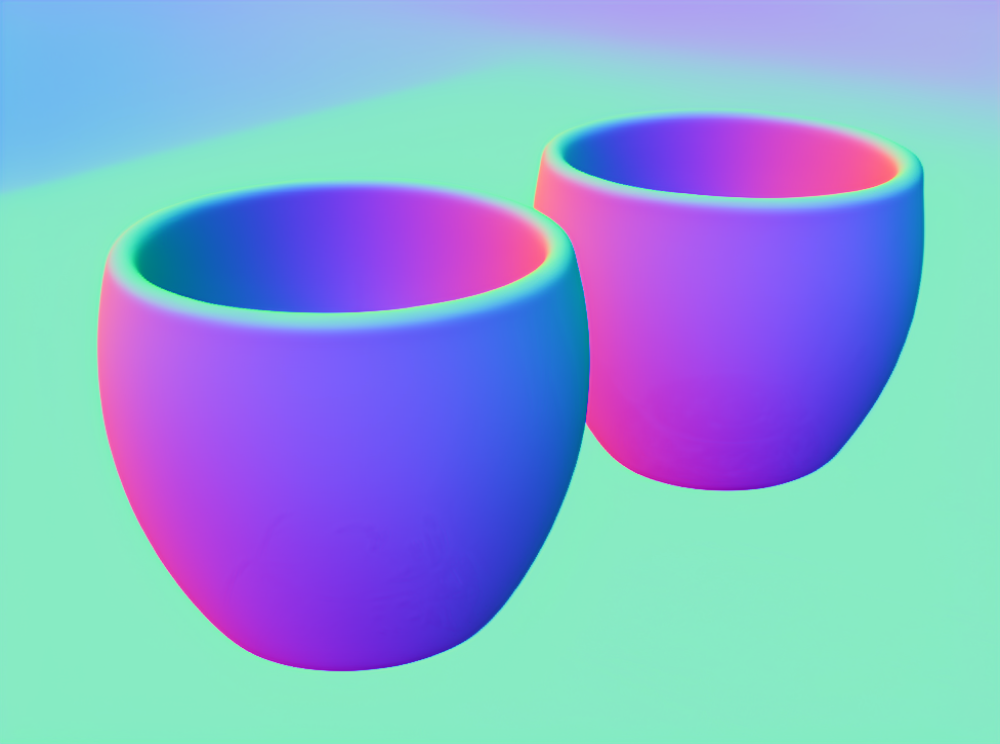}{-0.3cm}{-0.25cm} \\[1pt]
        \includegraphics[width=\imgwt]{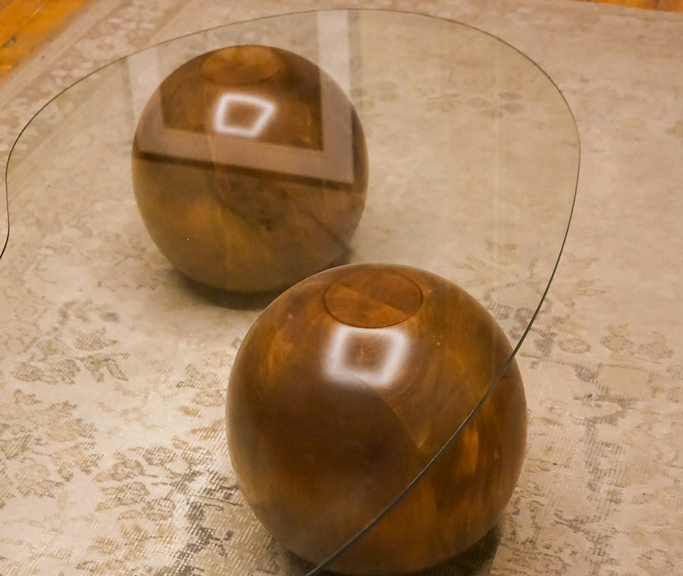} &
        \teaserbox{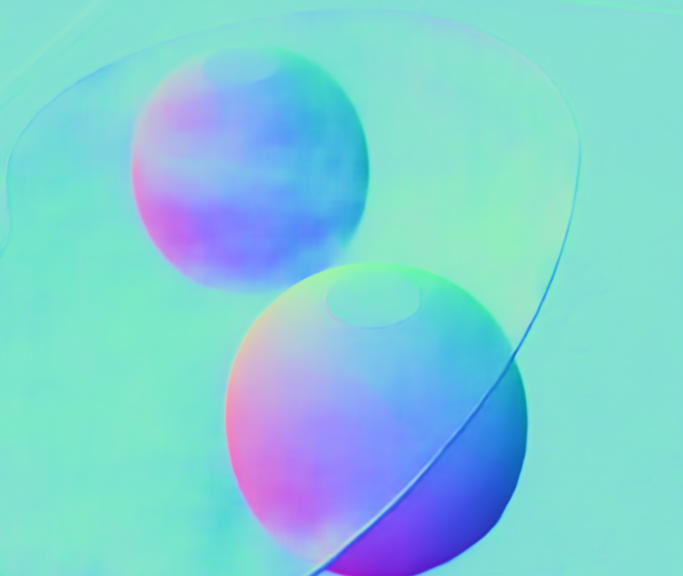}{0.25cm}{-0.20cm} &
        \teaserbox{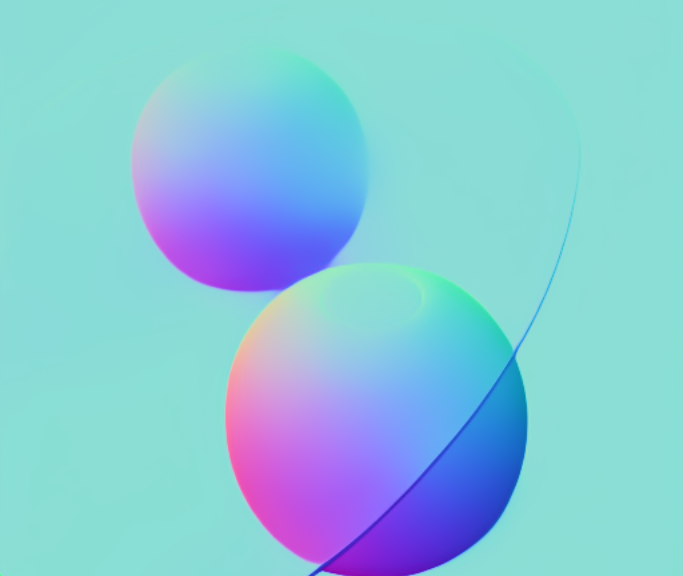}{0.25cm}{-0.20cm} &
        \teaserbox{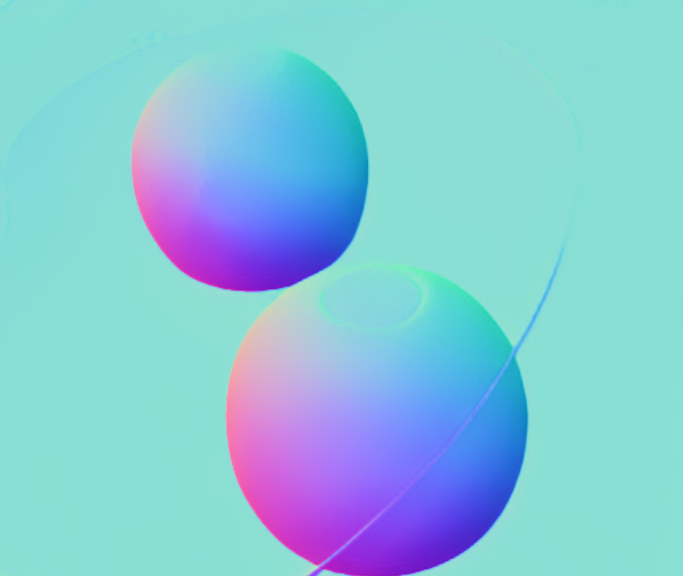}{0.25cm}{-0.20cm} &
        \teaserbox{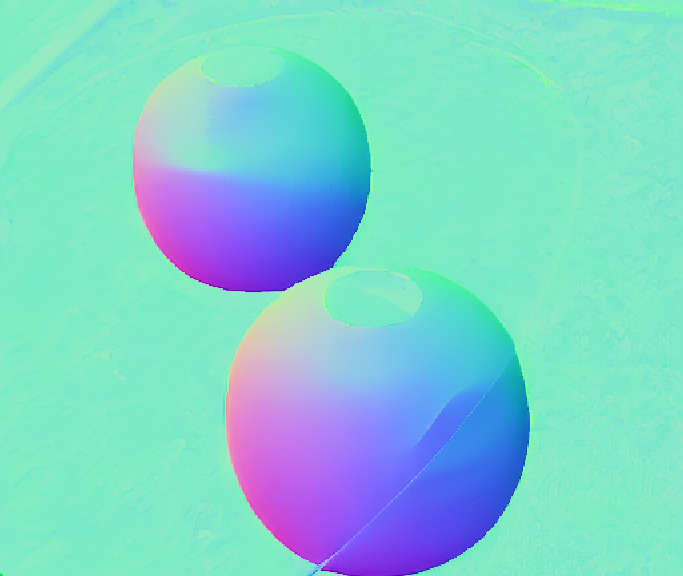}{0.25cm}{-0.20cm} &
        \teaserbox{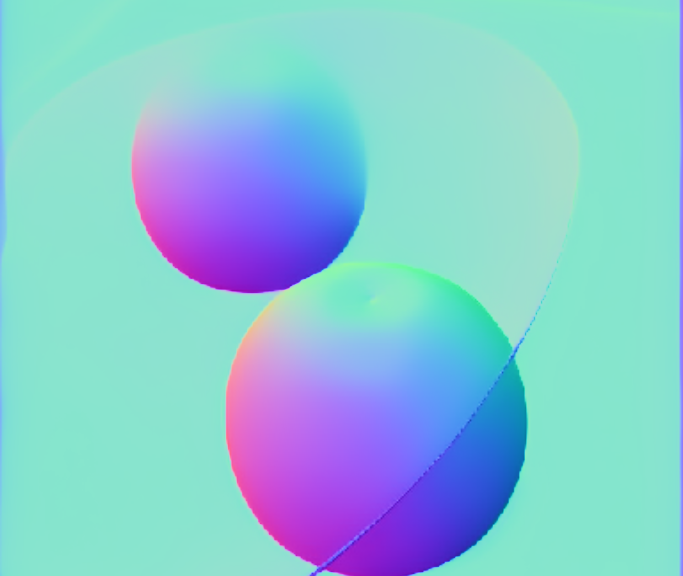}{0.25cm}{-0.20cm} &
        \teaserbox{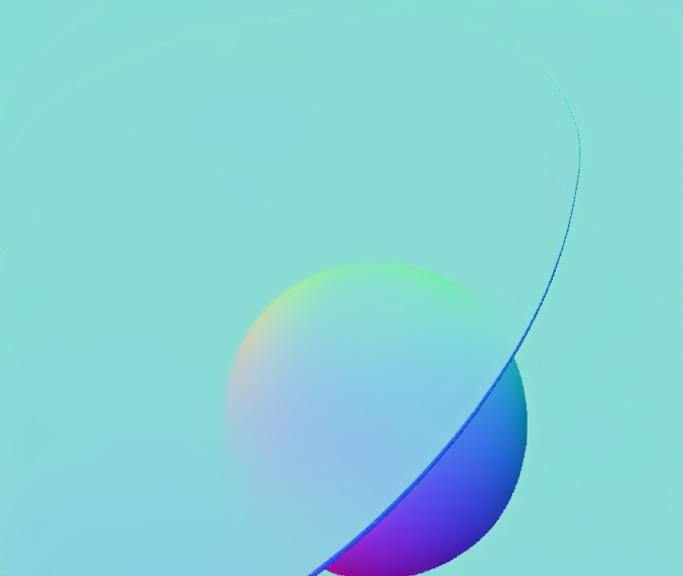}{0.25cm}{-0.20cm} &
        \teaserbox{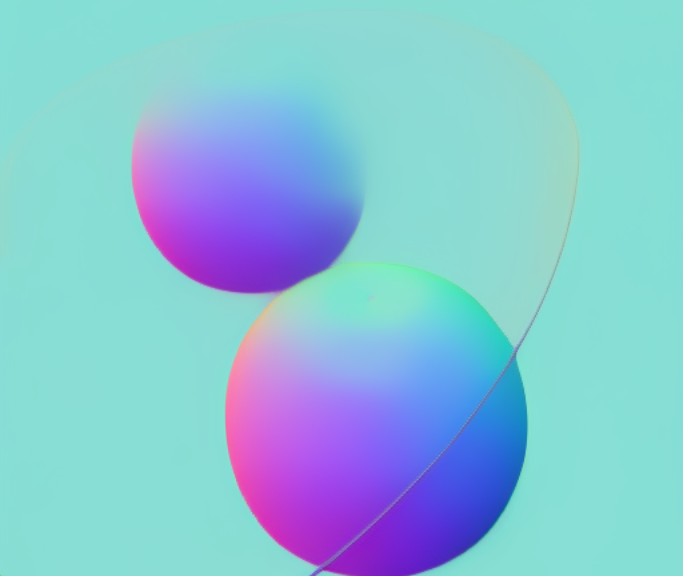}{0.25cm}{-0.20cm} &
        \teaserbox{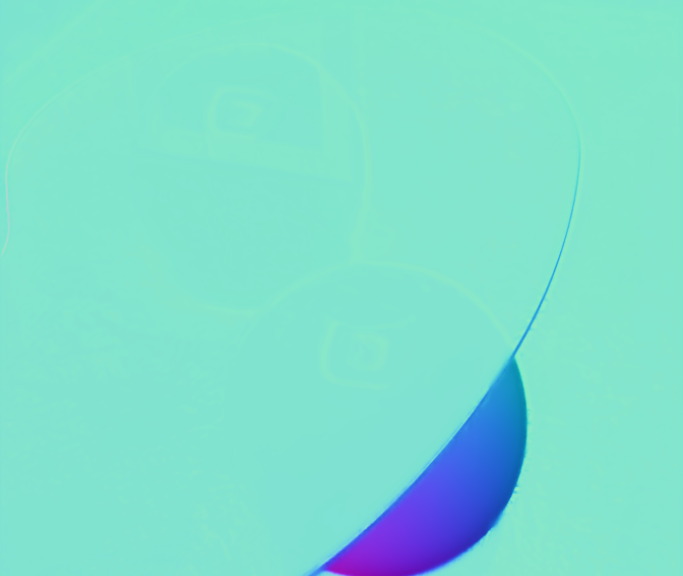}{0.25cm}{-0.20cm} \\
        {\small Input} & {\small MoGe-2} & {\small E2E-FT} & {\small Marigold} & {\small GeoWizard} & {\small Lotus-D} & {\small Lotus-2} & {\small TransNormal} & {\small \textbf{Ours}} \\
    \end{tabular}
    \caption{\textbf{In-the-wild qualitative results on surface normal estimation.} \textbf{TransNormal-2} (ours) recovers accurate surface normals on transparent objects.
    \textbf{Scene 1}: prism under extreme lighting.
    \textbf{Scene 2}: double-walled glass.
    \textbf{Scene 3}: balls viewed through a glass plate. (\S~\ref{ssec:qual_results})}
    \label{fig:teaser}
    \vspace{-3mm}
\end{figure*}
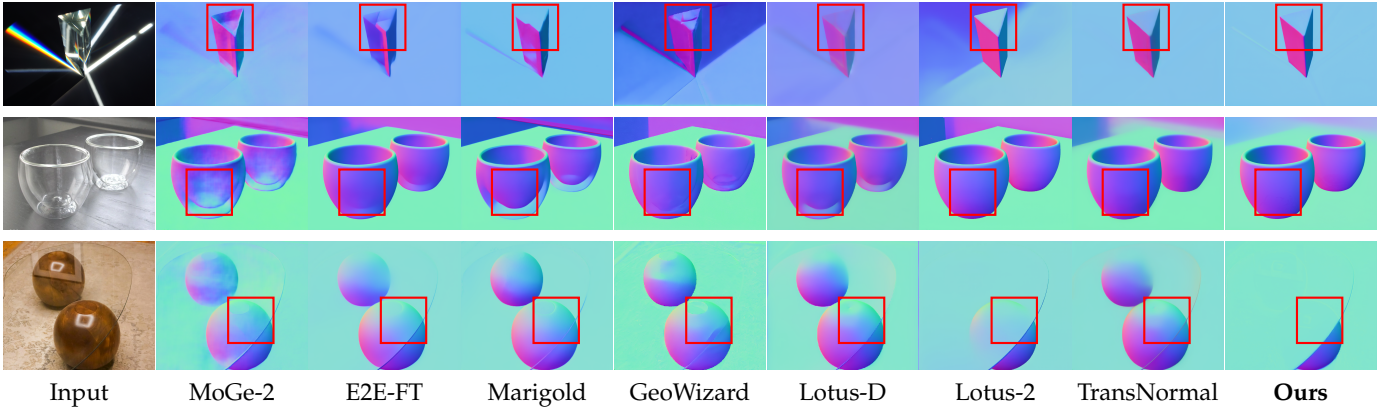

\IEEEraisesectionheading{\section{Introduction}\label{sec:introduction}}

\IEEEPARstart{R}{epurposing} pre-trained text-to-image diffusion models for dense geometric prediction has emerged as a powerful paradigm.  Pioneering works such as Marigold~\cite{ke2024repurposing} and Lotus~\cite{he2024lotus} demonstrate that these models encode rich geometric and material priors that generalize well beyond their original training distribution, while more recent efforts like Diffusion-E2E-FT (E2E-FT)~\cite{martingarcia2024diffusione2eft} and MoGe~\cite{wang2025moge} push accuracy further by end-to-end fine-tuning or discriminative reformulations.  The underlying architectures have also evolved, from U-Net-based latent diffusion~\cite{rombach2022high} to Diffusion Transformers (DiTs)~\cite{peebles2023scalable} with rectified flow training~\cite{liu2023flow, lipman2023flow}, culminating in multi-modal DiT (MM-DiT) models such as FLUX.2~\cite{flux2-2025} that offer stronger visual priors for geometry estimation.

Despite this rapid progress, we observe that VAE-based latent-diffusion geometry methods share a \emph{common but under-studied source of error}: \textbf{VAE reconstruction degradation}.  The VAE encoder-decoder compresses spatial resolution by $8{\times}$, which can average abrupt normal changes at object boundaries.  This error source is inherited by methods that represent and reconstruct geometry through such VAEs~\cite{ke2024repurposing,he2024lotus,martingarcia2024diffusione2eft}, yet, to our knowledge, it has not been quantitatively studied for geometry estimation.  Our quantitative study reveals that VAE reconstruction alone introduces non-negligible angular error even when \emph{ground-truth} normals are fed as input, with edge-region error reaching up to $2.8{\times}$ the global average (\secref{ssec:vae_bottleneck}).  Because this degradation is introduced by the encode-decode path itself, it is invisible to a latent-space training objective and cannot be removed by better latent prediction alone; because it concentrates at object boundaries, correcting it calls for full-resolution evidence that the latent representation no longer carries.

These two observations shape our design.  We present \textbf{TransNormal-2}, a framework built on FLUX.2[klein]~\cite{flux2-2025} that addresses the degradation through two complementary controls, one on each side of the VAE decoder.  The first control acts during training: geometry-aware pixel-space losses supervise the decoded normal field, so that latent predictions are not judged only by VAE-latent MSE.  A von~Mises-Fisher loss treats normals as directions on $\mathbb{S}^2$ and penalizes angular rather than Euclidean deviation; wavelet regularization concentrates high-frequency supervision on the object boundaries where the VAE degrades most; and an inverse rendering self-consistency loss draws a further constraint from the input image itself, requiring predicted normals to explain the observed shading in diffuse regions.

The second control acts after decoding: the \textbf{Geometric Refinement Module (GRM)} applies a lightweight RGB-guided residual correction to the boundary-localized decoding errors that remain.  Its readout is a gated and scaled residual, so the module is constrained to correct residual errors rather than freely rewriting the coarse normal field.  Inference remains a single deterministic forward pass without iterative sampling.

TransNormal-2 matches or exceeds MoGe-2~\cite{wang2025moge2} on all eight general-scene metrics while using only $1.4\%$ as many task-specific normal annotations as MoGe-2.  The gains are most pronounced for transparent objects, where refractive appearance makes normal estimation difficult and boundary precision remains critical, reducing mean angular error by $4.2^\circ$ on ClearGrasp and $3.1^\circ$ on ClearPose over the strongest per-dataset prior baselines; compared with Lotus-2~\cite{he2025lotus2}, the ClearPose gain is $4.3^\circ$.

Our key contributions are:
\begin{itemize}[nosep,leftmargin=1.2em]
    \item \textbf{Systematic analysis of VAE reconstruction degradation} in diffusion-based geometry estimation: we show that the VAE encode-decode process introduces a systematic spatial bias with edge regions disproportionately degraded, a limitation common to VAE-based latent-diffusion geometry pipelines.
    \item \textbf{Geometry-aware pixel-space training objectives}: an inverse rendering self-consistency loss based on Lambertian reflectance, combined with von~Mises-Fisher angular and wavelet edge-aware losses, to complement latent MSE with spherical-normal and image-formation constraints.
    \item \textbf{Geometric Refinement Module (GRM)}: a lightweight RGB-guided post-decoder module that reduces residual boundary-localized decoding errors through constrained residual learning with frozen transformer weights.
    \item \textbf{Strong results across seven benchmarks} spanning general and transparent-object scenes, matching or exceeding MoGe-2 on all eight reported general-scene metrics under far fewer task-specific annotations and achieving the strongest transparent-object results.
\end{itemize}

\begin{figure*}[t]
\centering\vspace{-4pt}%
\begin{minipage}[c]{0.34\textwidth}
    \centering
    \small
    \setlength{\tabcolsep}{2pt}
    \begin{tabular}{lccc}
    \toprule
    Dataset & Global & Edge & Ratio \\
    \midrule
    ClearGrasp & $1.29^\circ$ & $3.64^\circ$ & $2.82{\times}$ \\
    NYUv2 & $1.81^\circ$ & $2.27^\circ$ & $1.26{\times}$ \\
    ScanNet & $2.45^\circ$ & $3.29^\circ$ & $1.34{\times}$ \\
    iBims & $8.50^\circ$ & $14.61^\circ$ & $1.72{\times}$ \\
    \bottomrule
    \end{tabular}\\[2pt]
    {\small (a) VAE reconstruction error (MAE)}
\end{minipage}%
\hspace{3mm}%
\begin{minipage}[c]{0.625\textwidth}
    \centering
    \setlength{\tabcolsep}{0.4pt}
    \begin{tabular}{@{}cccc@{}}
        \includegraphics[width=0.248\linewidth,trim=145 0 10 0,clip]{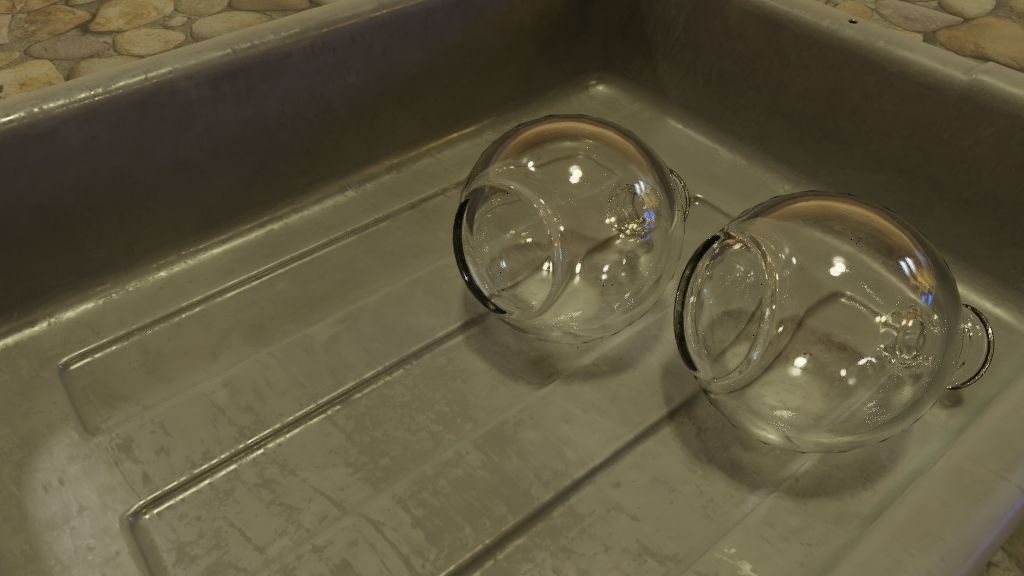} &
        \includegraphics[width=0.248\linewidth,trim=145 0 10 0,clip]{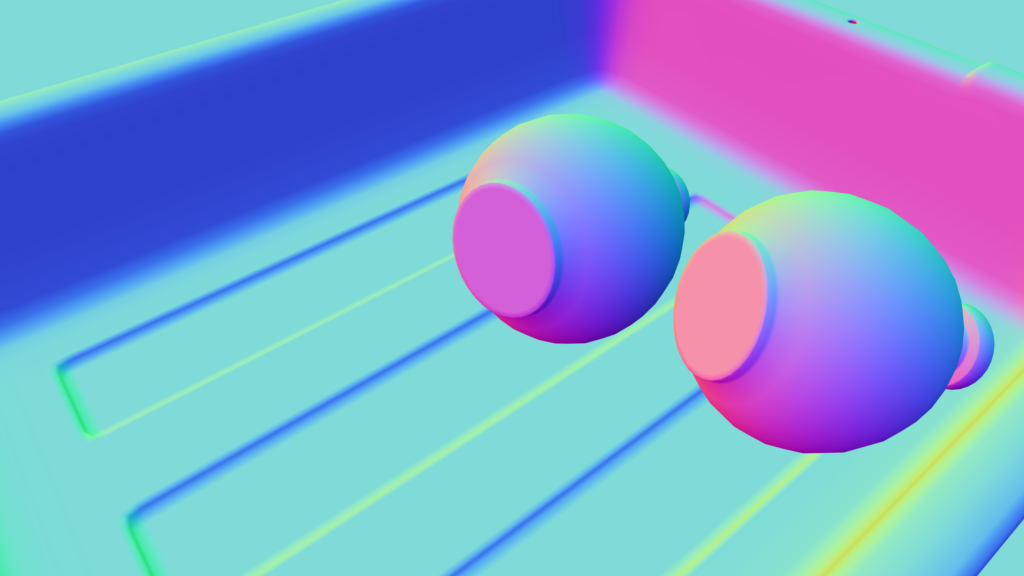} &
        \includegraphics[width=0.248\linewidth,trim=145 0 10 0,clip]{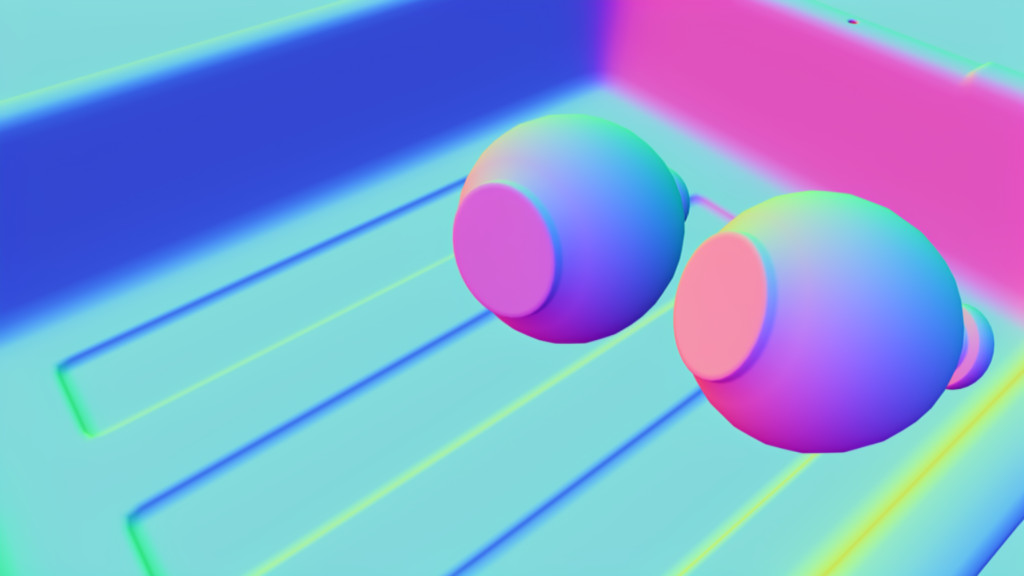} &
        \includegraphics[width=0.248\linewidth,trim=145 0 10 0,clip]{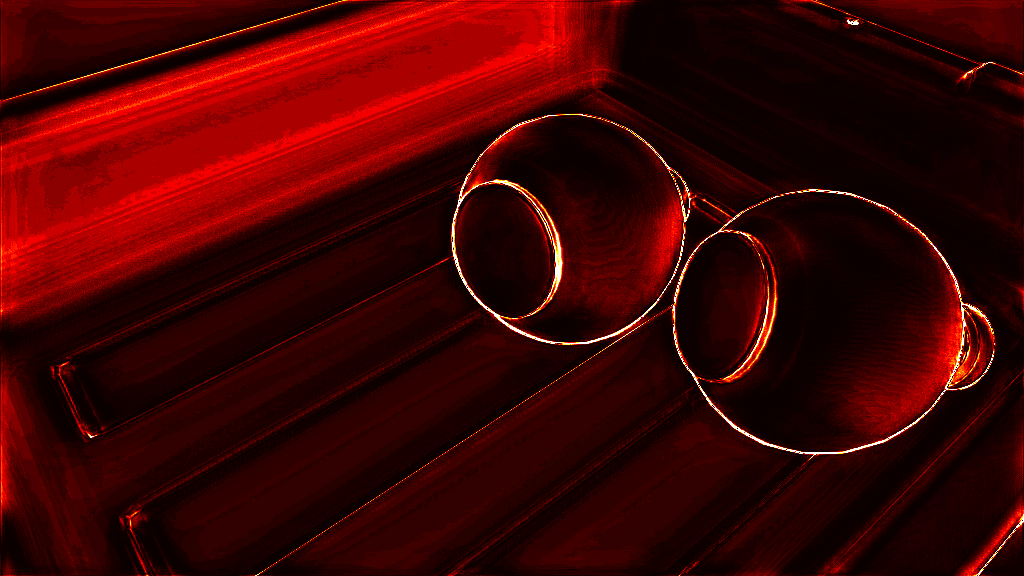} \\[-4pt]
        {\small Input RGB} & {\small GT Normal} & {\small VAE Recon.} & {\small Error Map} \\
    \end{tabular}\\[-1pt]
    {\small (b) ClearGrasp reconstruction visualization}
\end{minipage}
\vspace{-6pt}
\caption{\textbf{VAE reconstruction degradation.}  (a)~Mean angular error (MAE, $^\circ$) from encoding and decoding ground-truth normals with the frozen VAE; the Ratio column reports Edge/Global MAE.  (b)~The encode$\to$decode reconstruction and its full-frame angular error map (bright = high error) show that errors concentrate at geometric boundaries, consistent with boundary detail being blurred by the $8{\times}$ latent bottleneck.  (\S~\ref{ssec:vae_bottleneck})}
\label{fig:vae_roundtrip}
\end{figure*}

\section{Related Work}
\label{sec:related_work}

\subsection{Monocular Surface Normal Estimation}

Surface normal estimation originates from physics-based methods such as shape-from-shading~\cite{tomasi1992shape} and photometric stereo~\cite{woodham1980photometric}.  Deep discriminative approaches then dominated, spanning local canonical frames~\cite{huang2019framenet}, large-scale multi-task training~\cite{eftekhar2021omnidata}, geometric inductive biases~\cite{bae2024dsine}, joint depth-normal estimation~\cite{qi2020geonetpp, hu2024metric3dv2}, and transformer-based monocular geometry~\cite{ranftl2020towards, ranftl2021vision, dosovitskiy2020image}.

The field has recently shifted toward \emph{generative} normal estimation: Marigold~\cite{ke2024repurposing} repurposes Stable Diffusion priors for zero-shot geometry, GeoWizard~\cite{fu2024geowizard} extends this to joint depth-normal prediction, StableNormal~\cite{ye2024stablenormal} reduces diffusion variance, RoSE~\cite{li2026rose} exploits shading cues via image-to-video models, and NormalCrafter~\cite{bin2025normalcrafter} learns temporally consistent video normals through a two-stage latent-then-pixel protocol.

\subsection{Diffusion Models for Dense Geometric Prediction}

Repurposing pre-trained diffusion models~\cite{rombach2022high, ho2020denoising} for dense geometric tasks has matured rapidly: after Marigold~\cite{ke2024repurposing}, subsequent work showed that simple end-to-end fine-tuning suffices~\cite{martingarcia2024diffusione2eft}, exploited diffusion priors for generalizable dense prediction~\cite{lee2024exploiting}, systematically studied design choices~\cite{xu2024matters}, and introduced deterministic single-step prediction~\cite{he2024lotus}, with flow matching further improving efficiency~\cite{gui2025depthfm}.  Discriminative foundation models, from multi-scale CNNs~\cite{eigen2014depth} to Depth Anything~V2~\cite{yang2024depth2}, are faster and more metric but lack the generative priors useful for boundary preservation.

Backbones have evolved from U-Nets~\cite{ronneberger2015u} to Diffusion Transformers~\cite{peebles2023scalable} with rectified flow training~\cite{liu2023flow, lipman2023flow, esser2024scaling}, powering DiT-based geometry models~\cite{zhao2025diception, he2025lotus2, dens3r2026}, joint appearance-geometry modeling~\cite{krishnan2025orchid, kwon2025jointdit}, adaptations of image-\emph{editing} models whose pre-training carries stronger structural priors~\cite{wang2026fe2e, shi2025edit2perceive}, and video generative models whose cross-frame modeling transfers to cross-modal joint depth--normal prediction~\cite{yang2026geonext}.  Across this evolution, however, the spatial bias introduced by the VAE encode-decode path has received little attention.

\subsection{VAE Reconstruction Degradation in Latent Diffusion}

VAE-based latent-diffusion geometric methods represent and reconstruct predictions through a VAE~\cite{rombach2022high}, whose $8{\times}$ spatial compression can average abrupt geometric changes at object boundaries.  Recent work acknowledges this implicitly: NormalCrafter~\cite{bin2025normalcrafter} performs pixel-space fine-tuning after latent prediction, and Pixel-Perfect Depth~\cite{xu2025pixelperfectdepth} bypasses the VAE entirely by performing diffusion directly in pixel space to eliminate ``flying pixel'' artifacts.

From the VAE-architecture side, VA-VAE~\cite{vavae2025} reveals an optimization dilemma between token dimensionality and generation quality, VIVAT~\cite{vivat2025} catalogs and mitigates reconstruction artifacts, and DC-AE~\cite{chen2025dcae} pushes compression to $128{\times}$, together suggesting that the standard SD-VAE is suboptimal for geometry-sensitive tasks.

A complementary path, orthogonal to VAE-architecture changes, keeps the standard VAE and corrects residual boundary errors after decoding.  Such refinement has a long lineage, from edge-preserving filtering~\cite{he2013guided} to learned refinement driven by normal, edge, or frequency-domain cues~\cite{bae2022irondepth, qi2020geonetpp, depthmaster2025}, and most recently SharpDepth~\cite{pham2025sharpdepth}, which sharpens discriminative depth with generative boundary cues through costly iterative score distillation.  None of these works, however, quantifies the surface-normal VAE reconstruction degradation that motivates our correction; \secref{ssec:vae_bottleneck} provides this measurement.

\subsection{Geometry Estimation for Transparent Objects}

Transparent-object perception is challenged by refraction and reflection that corrupt conventional geometric cues.  Accurate geometry for such objects also underpins downstream robotic grasping and manipulation, where embodied agents must perceive and reason about objects' physical properties~\cite{liu2024reconfigurable, sun2025robotcognitive}.  Most prior work targets depth completion or estimation for transparent, mirror, and translucent surfaces from corrupted RGB-D or monocular cues~\cite{sajjan2020clear,dai2022domain,zhu2021lidf,xu2022transparenet,hong2022cluedepth,cai2023consistent,costanzino2023learning,dai2025depth,11128401}, supported by benchmarks~\cite{chen2022clearpose, fang2022transcg}; related directions include transparent segmentation~\cite{xie2020trans10k, xie2021trans10kv2, sun2023trosd} and 6D pose estimation~\cite{zhang2022transnet, jiang2024ebfa6d}.  Physics-based approaches exploit refraction~\cite{sulc2021refraction}, refractive flow~\cite{tang2024rftrans}, and polarization~\cite{shao2023polarization}, while multi-view RGB enables neural-implicit reconstruction~\cite{ichnowski2021dexnerf, li2023neto, zhou2023nvs4glass, deng2024tnsr, sun2024nunerf, li2025tsgs}.  Training data has progressed from physics-based rendering~\cite{sajjan2020clear} and multispectral capture~\cite{kim2024transpose} to generative synthesis~\cite{zhang2024transparent, agrawal2024clear} and stereo perception~\cite{cleardepth2024}.

A related development is DKT~\cite{xu2025dkt}, which repurposes \emph{video} diffusion models to internalize refraction and reflection cues for transparent-object depth and normal estimation; its video-based input-output setting differs from our single-image one, so the two are not directly comparable.

\begin{figure*}[t]
    \centering
    \includegraphics[width=\textwidth]{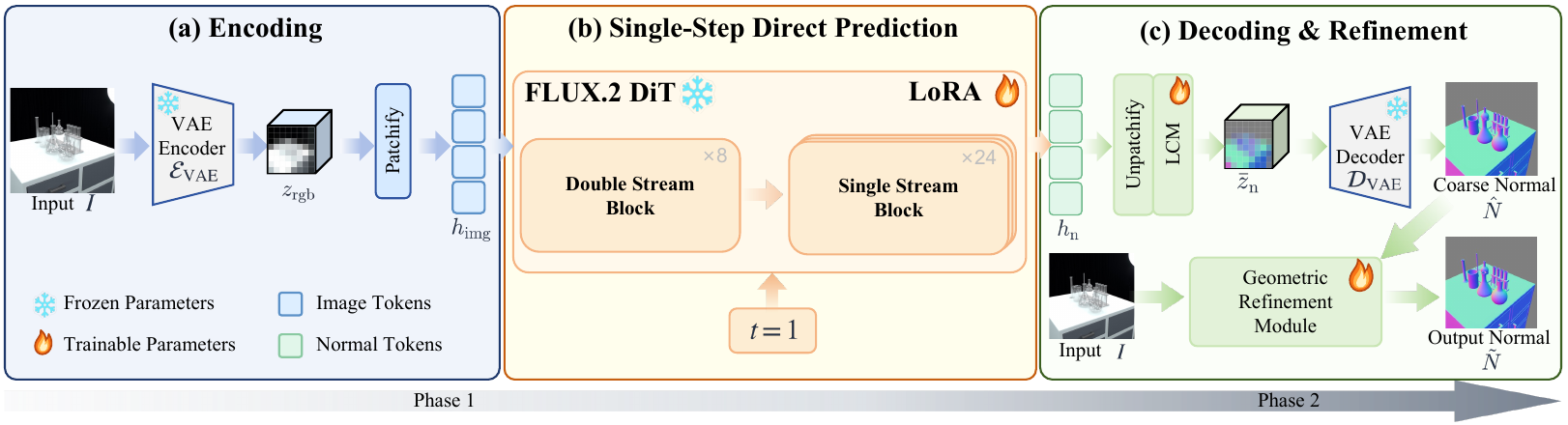}
    \caption{\textbf{Overview of the TransNormal-2 framework.}
    (a) \textbf{Encoding}: the frozen FLUX.2 VAE encoder $\mathcal{E}_{\text{vae}}$ extracts and patchifies the RGB latent $\bm{z}_{\text{rgb}}$ into image tokens $\bm{h}_{\text{img}}$;
    (b) \textbf{Single-Step Direct Prediction}: the fine-tuned FLUX.2[klein] DiT $f_{\vtheta}$ directly predicts the normal latent $\hat{\bm{z}}_{\text{n}}$ from $\bm{z}_{\text{rgb}}$ at a fixed timestep $t$ under empty-prompt conditioning, processing the tokens through the LoRA-adapted double-stream and single-stream blocks;
    (c) \textbf{Decoding \& Refinement}: the DiT output passes through a Local Continuity Module (LCM) in latent space, then the frozen FLUX.2 VAE decoder $\mathcal{D}_{\text{vae}}$ reconstructs the coarse geometry map, which is further refined by the Geometric Refinement Module (GRM) using the input RGB's edge information. Training combines latent normal MSE ($\Ls_{\text{normal}}$), wavelet edge-aware regularization ($\Ls_{\text{wavelet}}$), von~Mises-Fisher angular loss ($\Ls_{\text{vmf}}$), and inverse rendering self-consistency loss ($\Ls_{\text{render}}$). (\S~\ref{par:method_overview})}
    \label{fig:pipeline}
\end{figure*}

\section{Preliminaries and Bottleneck Analysis}
\label{sec:preliminaries}

\noindent\textbf{Setup and Notation.} TransNormal-2 builds on FLUX.2[klein]~\cite{flux2-2025}, a rectified flow DiT~\cite{peebles2023scalable, liu2023flow} that operates in a compressed latent space.  A frozen VAE provides encoder $\mathcal{E}_{\text{vae}}$ and decoder $\mathcal{D}_{\text{vae}}$, mapping between pixel space and latent space with $8{\times}$ spatial downsampling.  Following recent dense prediction works~\cite{ke2024repurposing, he2024lotus, xu2024matters}, we encode both the RGB image $\bm{I}$ and ground-truth normal map $\bm{N}$ into latents $\bm{z}_{\text{rgb}} = \mathcal{E}_{\text{vae}}(\bm{I})$ and $\bm{z}_{\text{n}} = \mathcal{E}_{\text{vae}}(\bm{N})$, and repurpose the DiT as a deterministic predictor $f_{\vtheta}$ that directly predicts a raw normal latent $\hat{\bm{z}}_{\text{n}} = f_{\vtheta}(\bm{z}_{\text{rgb}}, t, \bm{c})$ in a single forward pass, where $t$ is a fixed timestep ($t{=}1$) and $\bm{c}$ denotes empty-prompt conditioning tokens.  TransNormal-2 decodes the LCM-adjusted latent $\bar{\bm{z}}_{\text{n}}=\mathrm{LCM}_{\psi}(\hat{\bm{z}}_{\text{n}})$, yielding the coarse normal $\hat{\bm{N}} = \mathcal{D}_{\text{vae}}(\bar{\bm{z}}_{\text{n}})$; \secref{ssec:decoding} describes the LCM.  At the pixel level, we write $\bm{n}^*(p)$, $\hat{\bm{n}}(p)$, and $\tilde{\bm{n}}(p)$ for the ground-truth, coarse-predicted, and refined unit normals at pixel $p$; a full notation table is provided in the Supplementary Material.

\subsection{VAE Reconstruction Degradation}
\label{ssec:vae_bottleneck}

\noindent\textbf{Compression Failure Mode.} In this paradigm, predictions are decoded through a VAE that was originally designed for natural images.  The $8{\times}$ spatial compression ($H{\times}W \to \frac{H}{8}{\times}\frac{W}{8}$) is benign for smooth color gradients but destructive for geometric discontinuities at object boundaries, where the normal field changes abruptly.  Let $\bm{N}^{\mathrm{rec}}=\mathcal{R}_{\text{vae}}(\bm{N})=\mathcal{D}_{\text{vae}}(\mathcal{E}_{\text{vae}}(\bm{N}))$ denote the frozen VAE reconstruction.  Because the latent grid has only one spatial site for each $8{\times}8$ pixel block, details that vary within such a block cannot be faithfully represented by the latent code and must be reconstructed from learned priors.  This creates a simple failure mode for surface normals: smooth regions are usually preserved, while boundary-localized high-frequency changes are blurred or averaged during reconstruction.

\noindent\textbf{One-Dimensional Boundary Model.} This failure mode is especially visible at object boundaries, and a one-dimensional boundary model makes the consequence explicit.  Let a row of pixels cross a boundary at $x=0$, where the normal jumps from $\bm{n}^-$ to $\bm{n}^+$, and write $\Delta\bm{n}=\bm{n}^+-\bm{n}^-$ for the jump.  Reconstruction replaces the ideal step $\mathds{1}[x\ge0]$ by a blurred transition $F_\sigma(x)$ of width $\sigma$, so the reconstructed normal is $\bm{n}^-+\Delta\bm{n}\,F_\sigma(x)$ and the pointwise error is $\Delta\bm{n}\,[F_\sigma(x)-\mathds{1}[x\ge0]]$.  Integrating its magnitude along the row gives
\begin{equation}
E_{\text{edge}} = \|\Delta\bm{n}\|_2
\int\!\left|F_\sigma(x)-\mathds{1}[x\ge0]\right|\,dx
\;\propto\; \|\Delta\bm{n}\|_2\,\sigma ,
\label{eq:edge_degradation}
\end{equation}
since the integral is the area between the blurred and the ideal step.  The error is therefore confined to the boundary band and grows with the size of the normal jump and the blur width; away from boundaries, where $\Delta\bm{n}=\bm{0}$, the same blur is harmless.  Because the angular error between unit normals is monotone in their Euclidean discrepancy, the same holds for the angular error we report.

\noindent\textbf{Ground-Truth Reconstruction Test.} To measure this effect in the actual FLUX.2 VAE, we conduct a controlled \emph{VAE reconstruction} diagnostic: ground-truth normal maps are encoded into the VAE latent space and decoded back, measuring the angular error introduced by this compression alone, independent of any model prediction.  \Figref{fig:vae_roundtrip} quantifies the scale of this degradation: encoding and decoding ground-truth normals alone introduces $1.3^\circ$--$8.5^\circ$ of mean angular error (MAE) across the four benchmarks.  The error is also spatially concentrated: Edge/Global MAE reaches $2.82{\times}$ on ClearGrasp and $1.72{\times}$ on iBims, confirming that the degradation is a boundary-localized bias rather than a uniform offset.  The same diagnostic applied to other latent-diffusion VAEs shows the same edge-concentrated degradation, so the bottleneck is not specific to the FLUX.2 VAE (Supplementary Material).

\noindent\textbf{Implications.} These measurements agree with the boundary model in Eq.~\eqref{eq:edge_degradation}: VAE reconstruction degrades boundary-localized normal discontinuities more than smooth regions.  Surface normal maps are particularly vulnerable because they jump at every crease, including where depth stays continuous.  These findings motivate a post-decode correction stage (\secref{ssec:edge_refine}) that uses full-resolution RGB evidence to reduce residual boundary-localized angular error.

\section{Method}
\label{sec:method}
\noindent\textbf{Overview.}\label{par:method_overview}
Given the VAE reconstruction bottleneck identified in \secref{ssec:vae_bottleneck}, we design TransNormal-2 around two complementary controls: \emph{geometry-aware pixel-space supervision} during training (\secref{ssec:training_losses}), which makes decoded predictions respect normal geometry, and a \emph{post-decode correction} at inference (\secref{ssec:decoding}), which reduces the boundary-localized errors that remain.  The two controls divide the problem between them: the first improves what the latent can encode, the second recovers what it cannot.  Since even ground-truth normals lose $1.3^\circ$--$8.5^\circ$ in the VAE round trip (\figref{fig:vae_roundtrip}), the decoder alone cannot restore the boundary detail that the $8{\times}$ latent grid discards; recovering it requires evidence that bypasses the bottleneck.  Our backbone is FLUX.2[klein]~\cite{flux2-2025}, a 9B-parameter rectified flow DiT adapted via LoRA~\cite{hu2022lora} for single-step geometry prediction.  We first describe encoding and prediction (\secref{ssec:encoders_semantic}), then decoding and the GRM (\secref{ssec:decoding}), and finally the two-phase training objectives (\secref{ssec:training_losses}).

\subsection{Encoding and Prediction}\label{ssec:encoders_semantic}\label{sec:notation}
With the notation of \secref{sec:preliminaries}, the RGB latent $\bm{z}_{\text{rgb}}$ is patchified and linearly projected into image tokens $\bm{h}_{\text{img}}$, and the empty prompt yields a fixed set of conditioning tokens $\bm{c} \in \R^{N_c \times d}$.  Both token streams pass through the double-stream blocks of FLUX.2, where they interact through joint attention, and then through its single-stream blocks; the output normal tokens $\bm{h}_{\text{n}}$ are unpatchified into the raw normal latent $\hat{\bm{z}}_{\text{n}} = f_{\vtheta}(\bm{z}_{\text{rgb}}, t, \bm{c})$ (\figref{fig:pipeline}).  The LoRA adapters constitute the trainable DiT parameters $\vtheta$; all pre-trained weights stay frozen.

\subsection{Decoding and Geometric Refinement}\label{ssec:decoding}\label{ssec:edge_refine}
Following~\cite{he2024lotus}, the predicted latent first passes through a lightweight \emph{Local Continuity Module} (LCM) with parameters $\psi$, yielding $\bar{\bm{z}}_{\text{n}} = \mathrm{LCM}_{\psi}(\hat{\bm{z}}_{\text{n}})$, and is then decoded to pixel space via $\hat{\bm{N}} = \mathcal{D}_{\text{vae}}(\bar{\bm{z}}_{\text{n}})$.  The LCM repairs latent-level seams left by patchification; the boundary blur of \secref{ssec:vae_bottleneck} is instead a property of the $8{\times}$ latent grid itself, which no latent-side adjustment can undo.

\noindent\textbf{Geometric Refinement Module (GRM).}
Since the input RGB image never passes through the normal-latent bottleneck, its full-resolution edges provide a complementary cue for locating where VAE decoding is likely to introduce boundary-localized error.  The GRM refines the decoded coarse normal $\hat{\bm{N}} \in \R^{3{\times}H{\times}W}$ in two steps.  It first forms a parameter-free \emph{anchor} $\bm{N}_{\text{anc}}$: on opaque-domain images, the coarse normal filtered by the RGB-guided filter~\cite{he2013guided}, which transfers the full-resolution RGB edge structure to the normal map; on transparent-domain images, the coarse normal itself, because RGB edges are unreliable on refractive surfaces.  The domain is given by a per-image binary flag $\mathds{1}_{\text{T}}$ ($1$ for the transparent-object domain), set from the source dataset during training and per benchmark at evaluation (\secref{ssec:implementation_details}).  The GRM then predicts a gated residual on top of the anchor from the concatenation of $\hat{\bm{N}}$, $\bm{N}_{\text{anc}}$, the input RGB image $\bm{I} \in \R^{3{\times}H{\times}W}$, and $\mathds{1}_{\text{T}}$ broadcast as a constant channel:
\begin{equation}
    \tilde{\bm{N}} = \mathrm{normalize}\!\left(\bm{N}_{\text{anc}} + g_{\varphi} \odot s\,R_{\varphi}(\hat{\bm{N}}, \bm{N}_{\text{anc}}, \bm{I}, \mathds{1}_{\text{T}})\right),
    \label{eq:grm_readout}
\end{equation}
where $R_{\varphi}$ is the GRM residual function, $g_{\varphi} \in (0,1)^{H{\times}W}$ is a per-pixel \emph{confidence gate} broadcast over the three channels, $s$ is a global residual scale, $\varphi$ denotes all trainable GRM parameters, and $\mathrm{normalize}(\cdot)$ denotes L2 normalization to produce unit normals; the network architecture and the inference settings are given in the Supplementary Material.

\subsection{Two-Phase Optimization Objectives}\label{ssec:training_losses}
We optimize TransNormal-2 in two decoupled phases matching \figref{fig:pipeline}: Phase~1 trains the \emph{core predictor}, i.e., the DiT LoRA parameters $\vtheta$ and the LCM parameters $\psi$, with pixel-space losses backpropagating through the frozen VAE decoder; Phase~2 trains only the GRM parameters $\varphi$ (\secref{ssec:edge_refine}).  We detail the two phase objectives in turn.

\noindent\textbf{Phase 1: Core-Predictor Objective.}
The first phase uses the following terms to make the predicted latent both VAE-compatible and geometrically faithful after decoding.

\noindent\textbf{Latent MSE.} The latent objective is the mean squared error between the LCM-adjusted normal latent and the VAE-encoded ground-truth normal, $\Ls_{\text{normal}} = \| \bar{\bm{z}}_{\text{n}} - \bm{z}_{\text{n}} \|^2_2$ averaged over latent elements.

\noindent\textbf{Wavelet Edge-Aware Regularization.}\label{par:wavelet_regularization}
Following~\cite{li2026transnormal}, we apply a Haar wavelet decomposition to provide edge-selective frequency supervision (illustrated in the Supplementary Material).  The decoded and ground-truth normal maps are decomposed into a low-frequency sub-band $\mathit{LL}$ and high-frequency sub-bands $\mathit{HF}{=}[\mathit{LH};\mathit{HL};\mathit{HH}]$.  An edge mask $M_{\text{edge}}$ derived from ground-truth normal gradients restricts high-frequency supervision to object boundaries:
\begin{equation}
    \Ls_{\text{wavelet}} = \underbrace{\|\widehat{\mathit{LL}} - \mathit{LL}\|_1}_{\text{shape fidelity}} + \underbrace{\|M_{\text{edge}}\odot(\widehat{\mathit{HF}} - \mathit{HF})\|_1}_{\text{edge-selective HF}}.
    \label{eq:wavelet_loss}
\end{equation}

\begin{figure*}[t]
    \centering
    \setlength{\tabcolsep}{0pt}
    \begin{tabular}{@{}>{\centering\arraybackslash}m{0.238\textwidth}@{\hspace{0.005\textwidth}}>{\centering\arraybackslash}m{0.004\textwidth}@{\hspace{0.005\textwidth}}>{\centering\arraybackslash}m{0.746\textwidth}@{}}
        \includegraphics[height=1.50in]{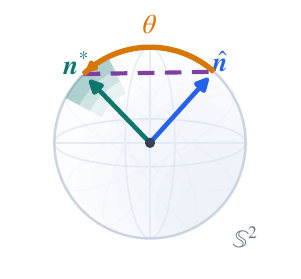} &
        {\color[HTML]{CBD5E1}\rule{0.42pt}{1.50in}} &
        \includegraphics[height=1.50in]{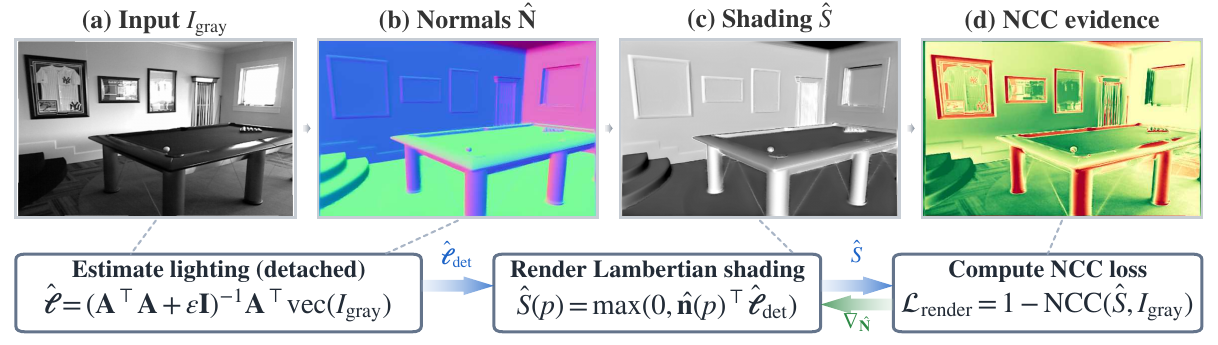}
    \end{tabular}
    \vspace{-2pt}
    \caption{\textbf{Geometry-aware pixel-space losses.} \figleft{} The vMF loss acts on decoded unit normals on $\mathbb{S}^2$: maximizing $\hat{\bm{n}}^\top\bm{n}^*=\cos\theta$ shrinks the \textcolor[HTML]{D97706}{geodesic angle} $\theta$ between prediction and ground truth, a quantity that the latent-space MSE on VAE codes never sees.  The \textcolor[HTML]{7E3FA1}{Euclidean chord} is shown for reference; for unit vectors it is a monotone function of $\theta$. \figright{} Inverse rendering self-consistency estimates a gradient-detached lighting vector $\hat{\bm{\ell}}_{\text{det}}$, renders Lambertian shading from $\hat{\bm{N}}$, and minimizes $\Ls_{\text{render}}=1-\mathrm{NCC}(\hat{S}, I_{\text{gray}})$, with panel (d) showing standardized local NCC evidence. Together, the losses enforce spherical normal geometry and diffuse image formation in pixel space. (\S~\ref{ssec:training_losses})}
    \label{fig:geometry_losses}
    \label{fig:vmf}
    \label{fig:rendering_loss}
    \vspace{-8pt}
\end{figure*}

\noindent\textbf{Von Mises-Fisher Angular Loss.}\label{par:vmf_loss}
Surface normals are unit vectors on $\mathbb{S}^2$, but the latent-space MSE compares VAE codes and never sees the direction of the decoded normal.  We therefore add a pixel-space directional loss on the decoded unit normals, derived from the von~Mises-Fisher (vMF) distribution~\cite{fisher1953dispersion}, the canonical probability model for directional data.  If each predicted normal is taken as the mean direction of a vMF distribution with fixed concentration $\kappa$, the negative log-likelihood of the ground-truth normals averaged over the valid pixels $\Omega$ is, up to an additive constant,
\begin{equation}
    \Ls_{\text{vmf}} = - \frac{\kappa}{|\Omega|} \sum_{p \in \Omega} \hat{\bm{n}}(p)^\top \bm{n}^*(p),
\end{equation}
where $\hat{\bm{n}}(p)$ and $\bm{n}^*(p)$ are the predicted and ground-truth unit normals at pixel $p$ (\secref{sec:preliminaries}).  Since $\hat{\bm{n}}^\top\bm{n}^*=\cos\theta$ for the geodesic angle $\theta$ between the two directions (\figref{fig:vmf}, left), the loss decreases monotonically as the angular error shrinks; the fixed $\kappa$ only sets the overall scale of the loss.

\noindent\textbf{Inverse Rendering Self-Consistency Loss.}\label{par:rendering_loss}
Beyond supervised normal losses, the input image itself provides a weak geometric constraint: for diffuse or approximately Lambertian regions, correct normals should explain the observed shading under a simple lighting model.  We therefore introduce a complementary self-supervised loss that exploits this diffuse image-formation cue to regularize general-scene geometry.  If the predicted normals are correct, the rendered shading $\hat{S}(p) = \max(0, \hat{\bm{n}}(p)^\top \hat{\bm{\ell}})$ under an estimated light direction $\hat{\bm{\ell}}$ should correlate strongly with the observed grayscale intensity.  We define:
\begin{equation}
    \Ls_{\text{render}} = 1 - \mathrm{NCC}(\hat{S},\, I_{\text{gray}}),
\end{equation}
where $\hat{\bm{\ell}}$ is obtained via least-squares regression from the predicted normals and observed grayscale (with gradients detached), and $\mathrm{NCC}$ denotes normalized cross-correlation (Pearson correlation).  NCC's scale and shift invariance makes the loss naturally robust to unknown albedo and ambient lighting.  Gradients flow only through the predicted normals (the light estimate is treated as fixed per iteration), yielding an EM-style alternating optimization.
The pipeline is illustrated in \figref{fig:rendering_loss} (right); a full derivation, gradient analysis, and discussion of the Lambertian assumption are provided in the Supplementary Material.

\begin{samepage}
\noindent\textbf{Phase-1 Total Loss.}
The core-predictor objective combines the latent MSE with the three pixel-space terms:
\begin{equation}
    \begin{aligned}
    \Ls_{\text{phase1}}(\vtheta,\psi) ={}& \Ls_{\text{normal}} + \lambda_{\text{wavelet}}\Ls_{\text{wavelet}} + \lambda_{\text{vmf}}\Ls_{\text{vmf}} \\
    &+ \lambda_{\text{render}}\Ls_{\text{render}}.
    \end{aligned}
    \label{eq:phase1_objective}
\end{equation}
\end{samepage}

\noindent\textbf{Phase 2: Decoupled GRM Training.}
After Phase~1 converges, we freeze the core predictor (the VAE is frozen throughout) and update only the GRM parameters $\varphi$, so that the GRM is fitted to the frozen output of the core predictor.  This decoupling is deliberate: joint training would let GRM gradients leak back through the VAE decoder into the converged DiT, and a moving backbone would give the GRM a non-stationary error distribution to fit; \secref{par:ablation_studies} confirms that joint training underperforms.

\noindent\textbf{Two-Stage GRM Objective.}
The GRM is trained in two substages, a \emph{base} substage that learns the residual and a \emph{calibration} substage that tunes the gate, both minimizing a shared geometry term plus a substage-specific regularizer,
\begin{equation}
    \Ls_{\text{phase2}}^{\text{base/cal}}
    = \Ls_{\text{geo}} + \Ls_{\text{reg}}^{\text{base/cal}}.
    \label{eq:grm_loss}
\end{equation}
The geometry term, shared by both substages, supervises the refined unit normals through their cosine to the ground truth,
\begin{equation}
    \ell_{\text{geo}}(p)
    = \lambda_{\text{geo}}\!\left[1-\tilde{\bm{n}}(p)^\top\bm{n}^*(p)\right],
    \label{eq:grm_geo}
\end{equation}
with $\Ls_{\text{geo}}=|\Omega|^{-1}\sum_{p\in\Omega}\ell_{\text{geo}}(p)$.

\emph{Base substage.}  $\Ls_{\text{reg}}^{\text{base}}$ penalizes departures of the refined normals from the anchor $\bm{N}_{\text{anc}}$, treated as a constant, so that the learned residual stays small, and adds GRM-specific wavelet and vMF terms that keep the correction boundary-faithful.

\emph{Calibration substage.}  $\Ls_{\text{reg}}^{\text{cal}}$ keeps these terms and additionally calibrates the confidence gate $g_{\varphi}$: the gate is supervised to activate on pixels whose anchor error exceeds a threshold $\tau_{g}$ while its mean activation is penalized, and $\ell_{\text{geo}}$ is re-weighted toward those hard pixels.  Inputs and anchors are detached; the complete term definitions and coefficients are provided in the Supplementary Material.

\section{Experiments}
\label{sec:experiments}

\setlength{\tabcolsep}{5pt}
\begin{table*}[!t]
\scriptsize
\caption{\textbf{Quantitative comparison on general scene normal estimation}. We evaluate on NYUv2, ScanNet, iBims, and Sintel datasets.
Metrics: mean angular error (Mean$\downarrow$) and percentage of pixels within $11.25^\circ$ ($\uparrow$).
TransNormal-2 matches or exceeds MoGe-2 on all eight general-scene metrics while using only 122K training samples ($1.4\%$ of MoGe-2's 8.9M).
The \colorbox{best}{best}, \colorbox{best2}{second best}, and \colorbox{best3}{third best} results are highlighted. $^\star$: diffusion-based; $^\dagger$: transformer-based; $^\ddag$: trained on Sintel (not zero-shot). SA: SIGGRAPH Asia. (\S~\ref{ssec:general_normal})
}
\label{tab:general}
\centering
\newcommand{\generalmethodcite}[2]{\makebox[\linewidth][l]{#1\hfill\cite{#2}}}
\resizebox{\textwidth}{!}{
\begin{tabular}{>{\raggedright\arraybackslash}p{0.19\textwidth}|c|r|cc|cc|cc|cc|c}
\toprule

\multirow{2}{*}{Method} 
& \multirow{2}{*}{Venue}
& \multirow{2}{*}{Data$\downarrow$}
& \multicolumn{2}{c|}{NYUv2} 
& \multicolumn{2}{c|}{ScanNet} 
& \multicolumn{2}{c|}{iBims} 
& \multicolumn{2}{c|}{Sintel} 
& \textcolor{black}{Avg.} \\
& &
& Mean$\downarrow$ & $11.25^\circ\uparrow$
& Mean$\downarrow$ & $11.25^\circ\uparrow$
& Mean$\downarrow$ & $11.25^\circ\uparrow$
& Mean$\downarrow$ & $11.25^\circ\uparrow$
& Rank            \\
\midrule

\generalmethodcite{Omnidata V2$^\dagger$}{kar20223d}
& CVPR 22
& 12.2M
& 17.2 & 55.5
& 16.2 & 60.2
& 18.2 & 63.9
& 40.5 & 14.7
& 12.7  \\

\generalmethodcite{DSINE}{bae2024dsine}
& CVPR 24
& 160K
& 16.4 & 59.6
& 16.2 & 61.0
& 17.1 & 67.4
& 34.9 & 21.5
& 8.9  \\

\generalmethodcite{GeoWizard$^{\star}$}{fu2024geowizard}
& ECCV 24
& 280K
& 18.9 & 50.7
& 17.4 & 53.8
& 19.3 & 63.0
& 40.3 & 12.3
& 14.5  \\

\generalmethodcite{StableNormal$^{\star}$}{ye2024stablenormal}
& SA 24
& 250K
& 18.6 & 53.5
& 17.1 & 57.4
& 18.2 & 65.0
& 36.7 & 14.1
& 13.0  \\

\generalmethodcite{E2E-FT$^{\star}$}{martingarcia2024diffusione2eft}
& WACV 25
& 74K
& 16.5 & 60.4
& 14.7 & 66.1
& 16.1 & 69.7
& 33.5 & 22.3
& 6.0  \\

\generalmethodcite{GenPercept$^{\star}$}{xu2024matters}
& ICLR 25
& 74K
& 18.2 & 56.3
& 17.7 & 58.3
& 18.2 & 64.0
& 37.6 & 16.2
& 12.8  \\

\generalmethodcite{Lotus-G$^\star$}{he2024lotus}
& ICLR 25
& 59K
& 16.5 & 59.4
& 15.1 & 63.9
& 17.2 & 66.2
& 33.6 & 21.0
& 8.9  \\

\generalmethodcite{Lotus-D$^\star$}{he2024lotus}
& ICLR 25
& 59K
& \cellcolor{best3}16.2 & 59.8
& 14.7 & 64.0
& 17.1 & 66.4
& 32.3 & 22.4
& 6.5  \\

\generalmethodcite{Lotus-2$^\star$}{he2025lotus2}
& arXiv 25
& 59K
& 16.9 & 59.0
& 14.2 & 66.8
& \cellcolor{best3}15.4 & \cellcolor{best3}70.4
& \cellcolor{best3}30.3 & \cellcolor{best}\textbf{27.6}
& 5.6  \\

\generalmethodcite{MoGe-2$^{\dagger}$}{wang2025moge2}
& NeurIPS 25
& 8.9M
& \cellcolor{best}\textbf{14.7} & \cellcolor{best2}62.3
& \cellcolor{best2}12.8 & \cellcolor{best2}68.4
& \cellcolor{best}\textbf{14.7} & \cellcolor{best3}70.4
& \cellcolor{best2}29.3 & \cellcolor{best3}24.8
& \cellcolor{best2}2.3  \\

\generalmethodcite{FE2E$^{\star}$}{wang2026fe2e}
& CVPR 26
& 71K
& 16.3 & 59.2
& \cellcolor{best3}13.8 & \cellcolor{best3}67.2
& \cellcolor{best2}15.1 & \cellcolor{best2}70.6
& 31.2 & 22.2
& 4.6  \\

\generalmethodcite{Edit2Perceive$^{\star\ddag}$}{shi2025edit2perceive}
& CVPR 26
& 178K
& \cellcolor{best2}15.5 & \cellcolor{best3}61.8
& 14.0 & 66.6
& \cellcolor{best2}15.1 & \cellcolor{best}\textbf{71.1}
& - & -
& \cellcolor{best3}3.2  \\

\generalmethodcite{GeoNeXt$^{\star}$}{yang2026geonext}
& ECCV 26
& 59K
& 16.7 & 60.0
& 16.0 & 62.8
& 16.4 & 69.2
& 33.0 & 21.5
& 7.9  \\

\generalmethodcite{TransNormal$^{\star}$}{li2026transnormal}
& ICML 26
& 122K
& 16.6 & 59.6
& 15.3 & 63.3
& 16.4 & 68.5
& 35.1 & 18.8
& 8.8  \\

\midrule

\makebox[\linewidth][l]{\textbf{TransNormal-2}$^{\star}$\hfill\textbf{(Ours)}}
& -
& 122K
& \cellcolor{best}\textbf{14.7} & \cellcolor{best}\textbf{62.5}
& \cellcolor{best}\textbf{12.7} & \cellcolor{best}\textbf{69.2}
& \cellcolor{best}\textbf{14.7} & \cellcolor{best2}70.6
& \cellcolor{best}\textbf{29.2} & \cellcolor{best2}27.5
& \cellcolor{best}\textbf{1.4}  \\

\bottomrule
\end{tabular}
}
\vspace{-2mm}
\end{table*}

\setlength{\tabcolsep}{5pt}
\begin{table*}[!t]
\scriptsize
\caption{\textbf{Quantitative comparison on transparent object normal estimation}. We evaluate on ClearGrasp, our proposed TN-Syn, and ClearPose datasets.
Metrics: mean angular error (Mean$\downarrow$, lower is better) and percentage of pixels within $11.25^\circ$ and $30^\circ$ thresholds ($\uparrow$, higher is better).
Data$\downarrow$: number of labeled training samples for the geometric task.
TransNormal-2 achieves the best result on every reported metric using only 122K samples; gains are largest on ClearGrasp and ClearPose, while TN-Syn is near-saturated. See \tabref{tab:general} for general scene results.
The \colorbox{best}{best}, \colorbox{best2}{second best}, and \colorbox{best3}{third best} results are highlighted. $^\star$: diffusion-based; $^\dagger$: transformer-based; $^\ddag$: trained on Sintel (not zero-shot). SA: SIGGRAPH Asia. (\S~\ref{ssec:transparent_normal})
}
\label{tab:transparent}
\centering
\newcommand{\transparentmethodcite}[2]{\makebox[\linewidth][l]{#1\hfill\cite{#2}}}
\resizebox{\textwidth}{!}{
\begin{tabular}{>{\raggedright\arraybackslash}p{0.19\textwidth}|c|r|ccc|ccc|ccc|c}
\toprule

\multirow{2}{*}{Method} 
& \multirow{2}{*}{Venue}
& \multirow{2}{*}{Data$\downarrow$}
& \multicolumn{3}{c|}{ClearGrasp (Synthetic)} 
& \multicolumn{3}{c|}{TN-Syn} 
& \multicolumn{3}{c|}{ClearPose (Real-World)} 
& \textcolor{black}{Avg.} \\
& &
& Mean$\downarrow$ & $11.25^\circ\uparrow$ & $30^\circ\uparrow$
& Mean$\downarrow$ & $11.25^\circ\uparrow$ & $30^\circ\uparrow$
& Mean$\downarrow$ & $11.25^\circ\uparrow$ & $30^\circ\uparrow$
& Rank            \\
\midrule

\transparentmethodcite{Omnidata}{eftekhar2021omnidata}
& ICCV 21
& 12.2M
& 36.9 & 15.1 & 49.1
& 11.3 & 80.9 & 89.3
& 48.3 & 10.8 & 33.8
& 17.1  \\

\transparentmethodcite{Omnidata V2$^\dagger$}{kar20223d}
& CVPR 22
& 12.2M
& 33.8 & 18.3 & 55.9
& 8.2 & 87.0 & 92.6
& 51.7 & 13.8 & 33.2
& 15.6  \\

\transparentmethodcite{DSINE}{bae2024dsine}
& CVPR 24
& 160K
& 25.7 & 26.4 & 68.6
& 13.2 & 70.3 & 90.7
& 40.2 & 15.9 & 46.3
& 14.2  \\

\transparentmethodcite{Marigold$^{\star}$}{ke2024repurposing}
& CVPR 24
& 74K
& 27.6 & 31.0 & 65.3
& 6.2 & 90.4 & 96.3
& 33.0 & 25.5 & 57.5
& 10.8  \\

\transparentmethodcite{GeoWizard$^{\star}$}{fu2024geowizard}
& ECCV 24
& 280K
& 31.3 & 20.8 & 59.5
& 9.4 & 78.9 & 95.0
& 36.8 & 14.2 & 49.7
& 14.8  \\

\transparentmethodcite{StableNormal$^{\star}$}{ye2024stablenormal}
& SA 24
& 250K
& 32.0 & 17.5 & 65.3
& 7.6 & 86.8 & 96.3
& 37.1 & 14.1 & 57.5
& 13.8  \\

\transparentmethodcite{E2E-FT$^{\star}$}{martingarcia2024diffusione2eft}
& WACV 25
& 74K
& 22.6 & 42.1 & 73.3
& \cellcolor{best3}5.2 & \cellcolor{best3}91.9 & 97.0
& 32.0 & 32.5 & 59.4
& 6.9  \\

\transparentmethodcite{GenPercept$^{\star}$}{xu2024matters}
& ICLR 25
& 74K
& 25.8 & 30.3 & 70.9
& 6.9 & 87.6 & 97.0
& 31.6 & 31.2 & 63.0
& 9.2  \\

\transparentmethodcite{Lotus-G$^\star$}{he2024lotus}
& ICLR 25
& 59K
& 21.7 & 39.7 & 75.4
& 8.2 & 82.3 & 96.7
& 31.8 & 28.8 & 60.4
& 9.2  \\

\transparentmethodcite{Lotus-D$^\star$}{he2024lotus}
& ICLR 25
& 59K
& 21.9 & 37.0 & 75.7
& 9.0 & 80.9 & 97.1
& 31.3 & 23.2 & 59.5
& 9.3  \\

\transparentmethodcite{Lotus-2$^\star$}{he2025lotus2}
& arXiv 25
& 59K
& \cellcolor{best2}15.5 & \cellcolor{best3}52.3 & \cellcolor{best2}87.4
& 5.5 & 91.1 & \cellcolor{best3}97.4
& \cellcolor{best3}23.4 & \cellcolor{best3}43.0 & \cellcolor{best3}72.6
& \cellcolor{best3}3.2  \\

\transparentmethodcite{MoGe-2$^{\dagger}$}{wang2025moge2}
& NeurIPS 25
& 8.9M
& 26.6 & 17.0 & 64.2
& 6.2 & 90.1 & 96.8
& 36.2 & 14.3 & 48.3
& 12.3  \\

\transparentmethodcite{Diception$^\star$}{zhao2025diception}
& NeurIPS 25
& 500K
& 29.5 & 25.8 & 65.3
& 7.1 & 88.3 & 97.3
& 31.0 & 33.8 & 63.5
& 9.0  \\

\transparentmethodcite{FE2E$^{\star}$}{wang2026fe2e}
& CVPR 26
& 71K
& 16.9 & 43.2 & \cellcolor{best3}86.3
& 21.2 & 31.0 & 77.4
& \cellcolor{best2}22.2 & \cellcolor{best2}45.4 & \cellcolor{best2}75.1
& 7.9  \\

\transparentmethodcite{Edit2Perceive$^{\star\ddag}$}{shi2025edit2perceive}
& CVPR 26
& 178K
& 20.1 & 36.1 & 79.0
& 5.9 & 90.4 & \cellcolor{best3}97.4
& 28.0 & 31.4 & 64.8
& 5.8  \\

\transparentmethodcite{GeoNeXt$^{\star}$}{yang2026geonext}
& ECCV 26
& 59K
& 21.9 & 39.7 & 75.0
& 6.2 & 91.2 & 96.9
& 32.1 & 32.3 & 59.2
& 7.9  \\

\transparentmethodcite{TransNormal$^{\star}$}{li2026transnormal}
& ICML 26
& 122K
& \cellcolor{best3}16.1 & \cellcolor{best2}52.5 & 85.6
& \cellcolor{best2}3.9 & \cellcolor{best2}93.9 & \cellcolor{best2}98.3
& 25.5 & 38.9 & 71.2
& \cellcolor{best2}3.0  \\

\midrule

\makebox[\linewidth][l]{\textbf{TransNormal-2}$^{\star}$\hfill\textbf{(Ours)}}
& -
& 122K
& \cellcolor{best}\textbf{11.3} & \cellcolor{best}\textbf{65.0} & \cellcolor{best}\textbf{94.1}
& \cellcolor{best}\textbf{3.6} & \cellcolor{best}\textbf{95.8} & \cellcolor{best}\textbf{99.1}
& \cellcolor{best}\textbf{19.1} & \cellcolor{best}\textbf{52.3} & \cellcolor{best}\textbf{80.5}
& \cellcolor{best}\textbf{1.0}  \\

\bottomrule
\end{tabular}
}
\vspace{-2mm}
\end{table*}

\subsection{Setup}
\noindent\textbf{Implementation Details.}\label{ssec:implementation_details}
We implement TransNormal-2 by fine-tuning FLUX.2[klein] 9B~\cite{flux2-2025}, a rectified flow DiT.
The FLUX.2 VAE encoder and decoder are kept frozen; the DiT is adapted via LoRA~\cite{hu2022lora} (rank $r{=}256$, $\alpha{=}256$; see \secref{ssec:ablation_lora_rank} for rank ablation). The conditioning stream uses the text encoder with an empty prompt.

We use AdamW~\cite{loshchilov2019decoupled} with random horizontal flipping for augmentation.
Training follows a two-phase strategy.
\textbf{Phase~1} trains the DiT LoRA and LCM for 35K steps in total with a staged schedule: a general-scene warmup on Hypersim and Virtual~KITTI, a continual stage that adds the transparent-object datasets, and a final stage that enables the inverse-rendering loss.
\textbf{Phase~2} freezes the transformer and LCM and trains only the GRM with the two-stage objective in Eq.~\eqref{eq:grm_loss}.  A base substage learns the geometric residual, followed by a shorter calibration substage that specializes the confidence gate and hard-pixel correction.  We refer to this as the \emph{two-stage GRM training protocol} (\secref{ssec:ablation_grm_arch}).
All models are trained on 8 NVIDIA A100 GPUs (80\,GB) with a total batch size of 32.
At inference, we predict the normal map in a single forward pass from the RGB latent and empty-prompt conditioning, followed by the GRM of Eq.~\eqref{eq:grm_readout}, whose inference settings are listed in the Supplementary Material.  The GRM domain flag $\mathds{1}_{\text{T}}$ of Eq.~\eqref{eq:grm_readout} follows the source dataset during training; at evaluation it is set to the transparent-object domain for the three transparent-object benchmarks and to the opaque domain otherwise, including for in-the-wild images.
Exact loss coefficients, substage schedules, and data mixtures are provided in the Supplementary Material.

\noindent\textbf{Training Data.}
We aim to achieve strong performance with relatively limited supervised data, so we train only on synthetic data with ground-truth normal annotations. During training, we sample from the following datasets with a ratio of \textbf{35:15:45:5}:
\ding{172}~\emph{ClearGrasp}~\cite{sajjan2020clear}: a dataset for transparent objects containing 45,454 synthetic samples with normal annotations;
\ding{173}~\emph{TransNormal-Synthetic (TN-Syn)}~\cite{li2026transnormal}: a Blender-rendered dataset of laboratory scenes with transparent glassware, providing 3,555 training and 395 testing samples with pixel-accurate normal annotations;
\ding{174}~\emph{Hypersim}~\cite{roberts2021hypersim}: a photorealistic synthetic dataset of 461 indoor scenes, from which we utilize the official training split retaining 39,648 samples after filtering, resized to $576 \times 768$;
\ding{175}~\emph{Virtual KITTI}~\cite{cabon2020virtual}: a synthetic street-scene dataset covering five urban scenes, from which we use four scenes comprising 33,580 samples, cropped to $352 \times 1216$. 

\noindent\textbf{Evaluation Data.}
\label{ssec:eval-dataset}
For general-scene evaluation, we use four standard benchmarks: \emph{NYUv2}~\cite{silberman2012indoor}, \emph{ScanNet}~\cite{dai2017scannet}, \emph{iBims-1}~\cite{koch2018evaluation}, and \emph{Sintel}~\cite{butler2012naturalistic}.
For transparent objects, we evaluate on: the synthetic test split of \emph{ClearGrasp}~\cite{sajjan2020clear} (408 samples), the held-out test set of \emph{TN-Syn} (395 samples), and \emph{ClearPose}~\cite{chen2022clearpose} (120 samples). ClearPose is a challenging real-world dataset with diverse transparent objects under varying lighting conditions; we use it for zero-shot evaluation (not included in training) to assess generalization. For ClearPose, we use the subset with available meshes and recompute normals by reprojecting the ground-truth mesh, evaluating only within the transparent object mask. On all three transparent-object benchmarks, normal metrics are computed within the transparent-object mask for fair comparison.

\noindent\textbf{Baselines.}\label{par:baselines}
We compare TransNormal-2 against representative normal-estimation methods on both general-scene and transparent-object benchmarks.
The baselines include discriminative feed-forward models (Omnidata~\cite{eftekhar2021omnidata}, Omnidata V2~\cite{kar20223d}, DSINE~\cite{bae2024dsine}, MoGe-2~\cite{wang2025moge2}) and diffusion-based dense prediction methods (GeoWizard~\cite{fu2024geowizard}, StableNormal~\cite{ye2024stablenormal}, Marigold~\cite{ke2024repurposing}, Lotus~\cite{he2024lotus}, Lotus-2~\cite{he2025lotus2}, E2E-FT~\cite{martingarcia2024diffusione2eft}, GenPercept~\cite{xu2024matters}, Diception~\cite{zhao2025diception}, FE2E~\cite{wang2026fe2e}, Edit2Perceive~\cite{shi2025edit2perceive}, GeoNeXt~\cite{yang2026geonext}).

\subsection{Qualitative Results}\label{ssec:qual_results}

\begin{figure*}[t]
    \centering
    \setlength{\tabcolsep}{0pt}
    \renewcommand{\arraystretch}{0.6}
    \newcommand{\imgwiw}{0.1085\textwidth}
    \newcommand{\wiwcell}[1]{\includegraphics[width=\linewidth]{#1}}
    \begin{tabular}{@{}>{\centering\arraybackslash}m{\imgwiw}@{\hspace{1.5pt}\vrule width 0.35pt\hspace{1.5pt}}>{\centering\arraybackslash}m{\imgwiw}@{\hspace{1pt}}>{\centering\arraybackslash}m{\imgwiw}@{\hspace{1pt}}>{\centering\arraybackslash}m{\imgwiw}@{\hspace{1pt}}>{\centering\arraybackslash}m{\imgwiw}@{\hspace{1pt}}>{\centering\arraybackslash}m{\imgwiw}@{\hspace{1pt}}>{\centering\arraybackslash}m{\imgwiw}@{\hspace{1pt}}>{\centering\arraybackslash}m{\imgwiw}@{\hspace{1pt}}>{\centering\arraybackslash}m{\imgwiw}@{}}
        {\scriptsize Input} &
        {\scriptsize MoGe-2} &
        {\scriptsize E2E-FT} &
        {\scriptsize Marigold} &
        {\scriptsize GeoWizard} &
        {\scriptsize Lotus-D} &
        {\scriptsize Lotus-2} &
        {\scriptsize TransNormal} &
        {\scriptsize \textbf{Ours}} \\[1pt]
        \wiwcell{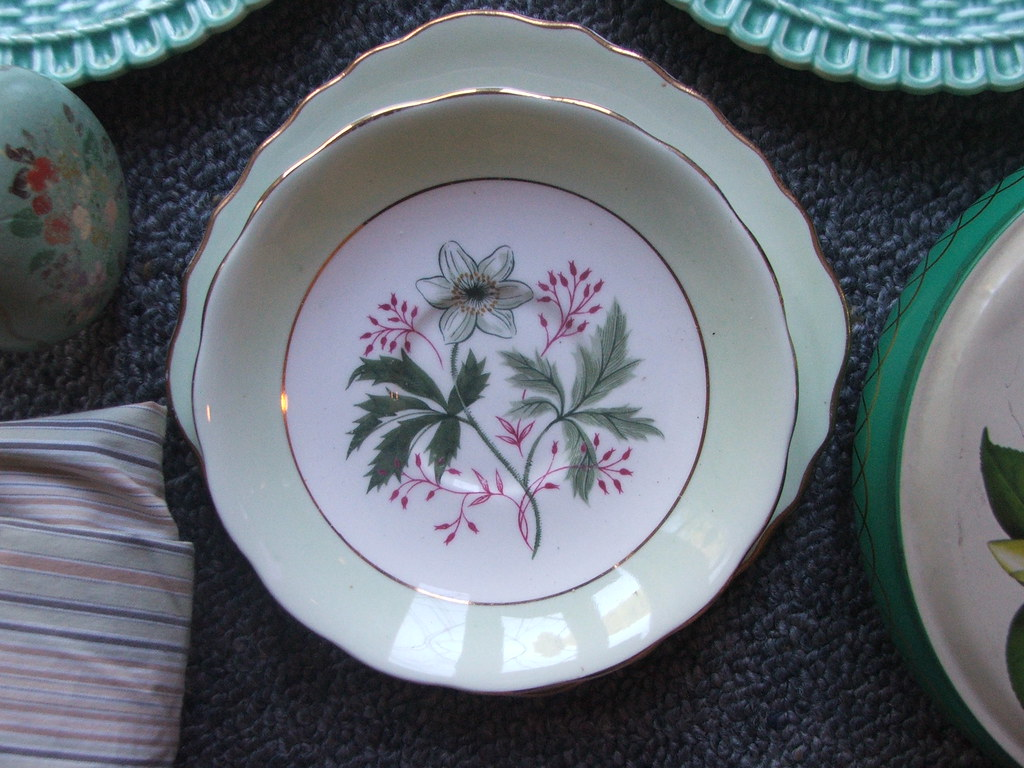} &
        \wiwcell{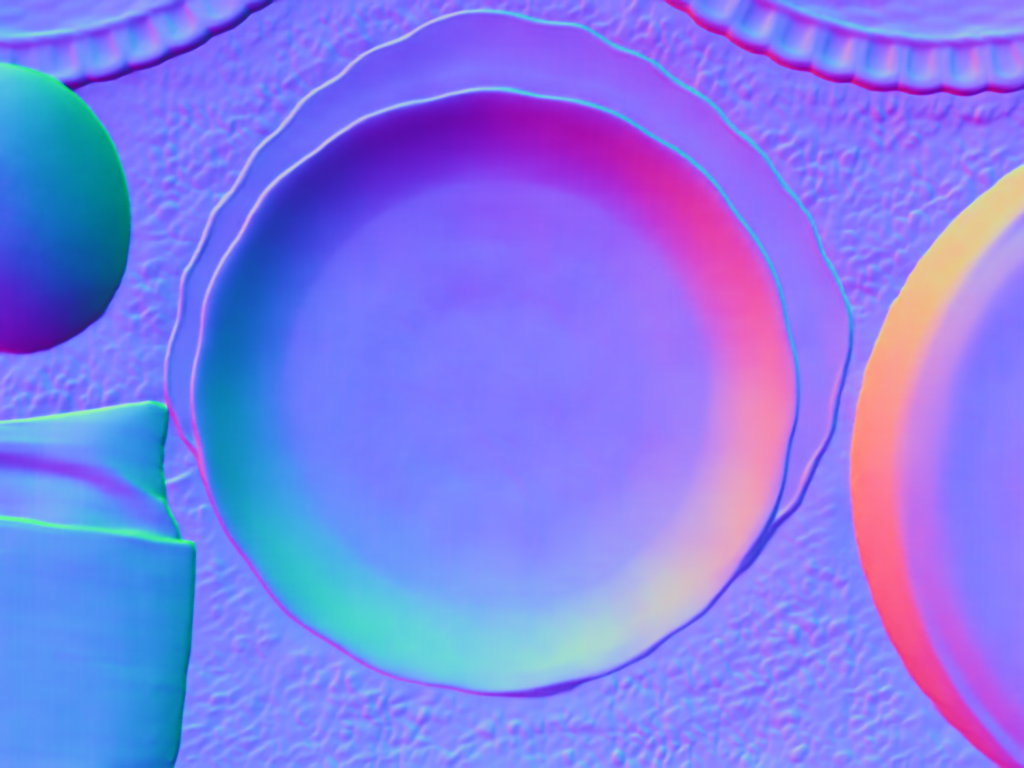} &
        \wiwcell{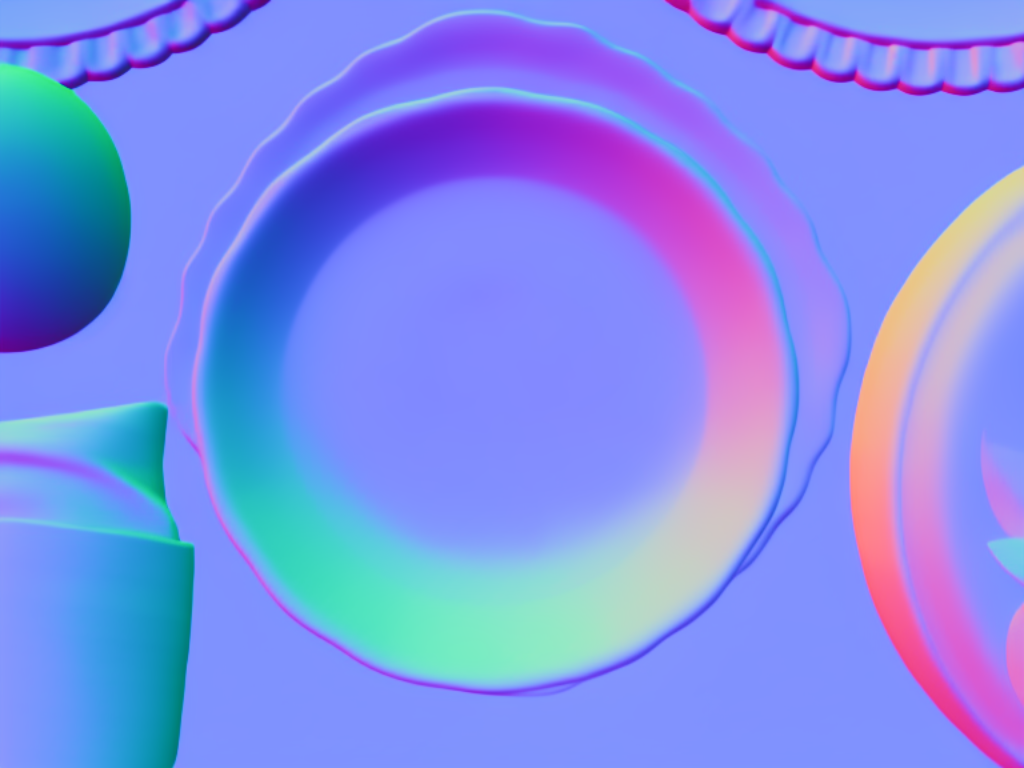} &
        \wiwcell{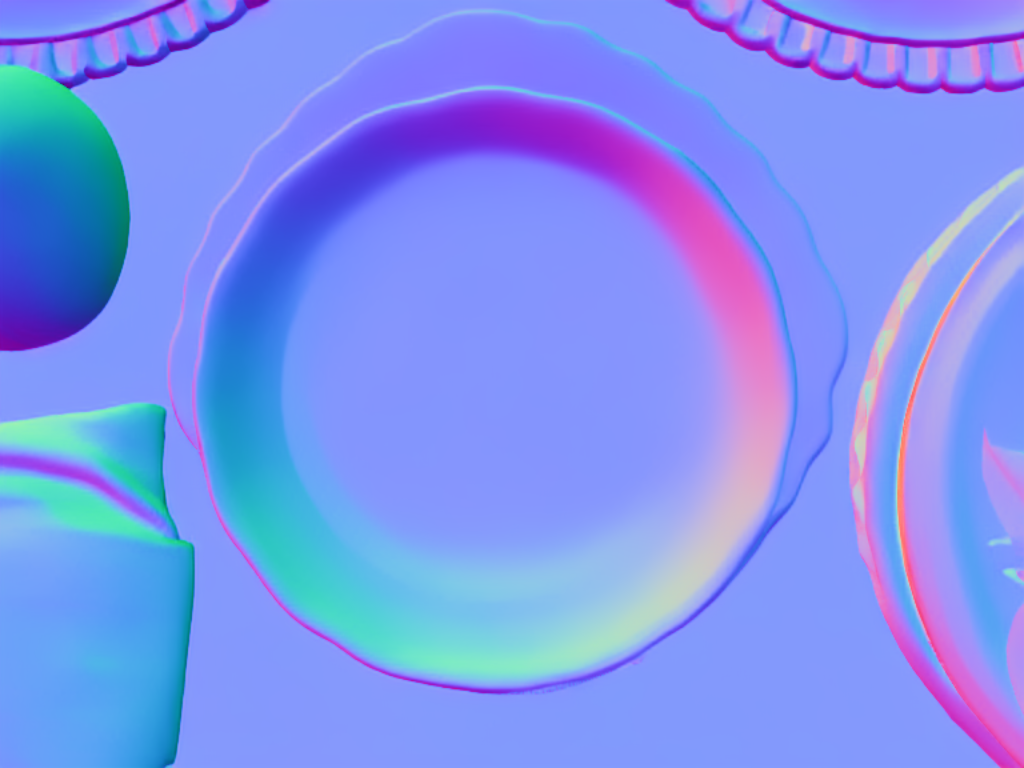} &
        \wiwcell{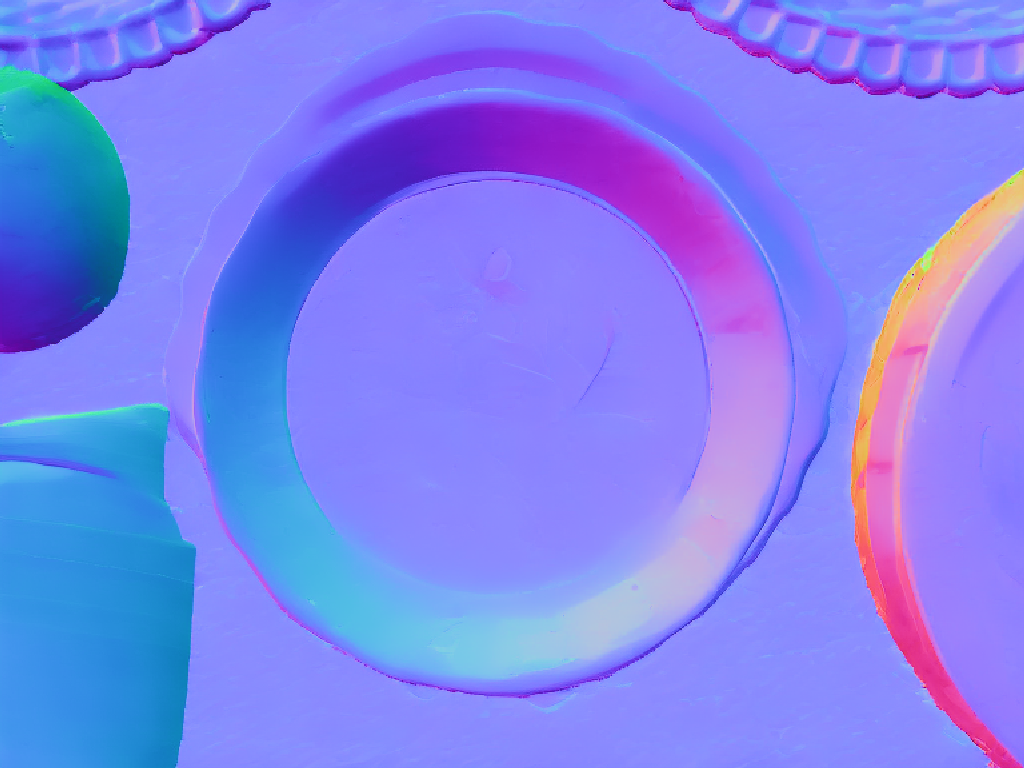} &
        \wiwcell{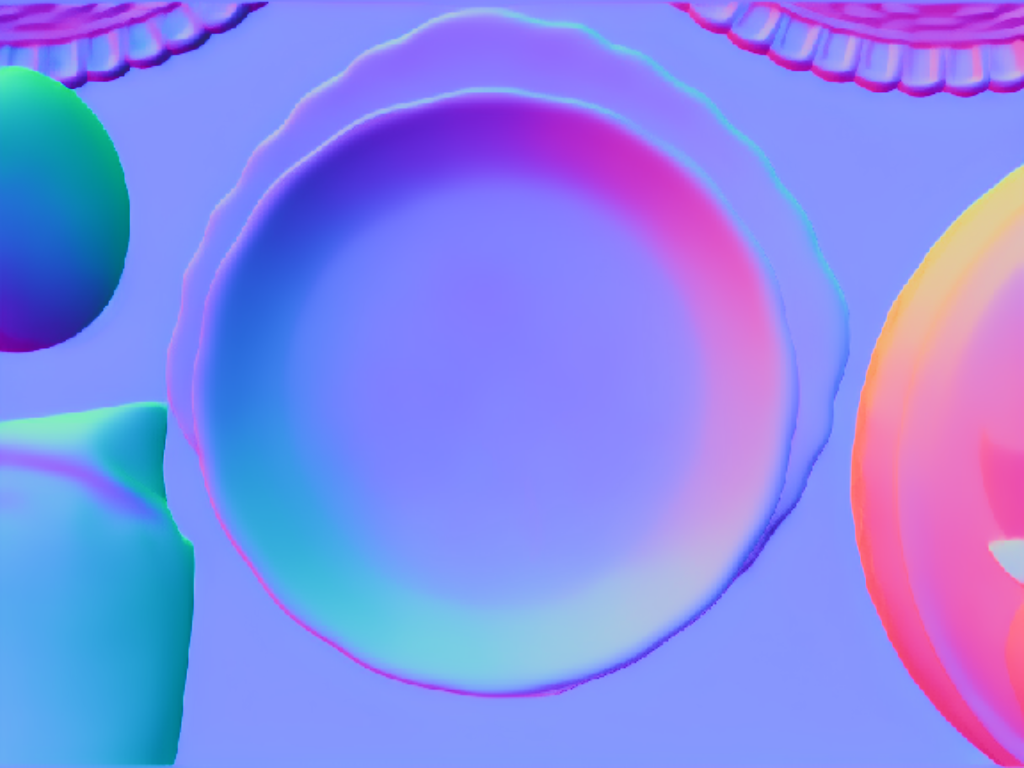} &
        \wiwcell{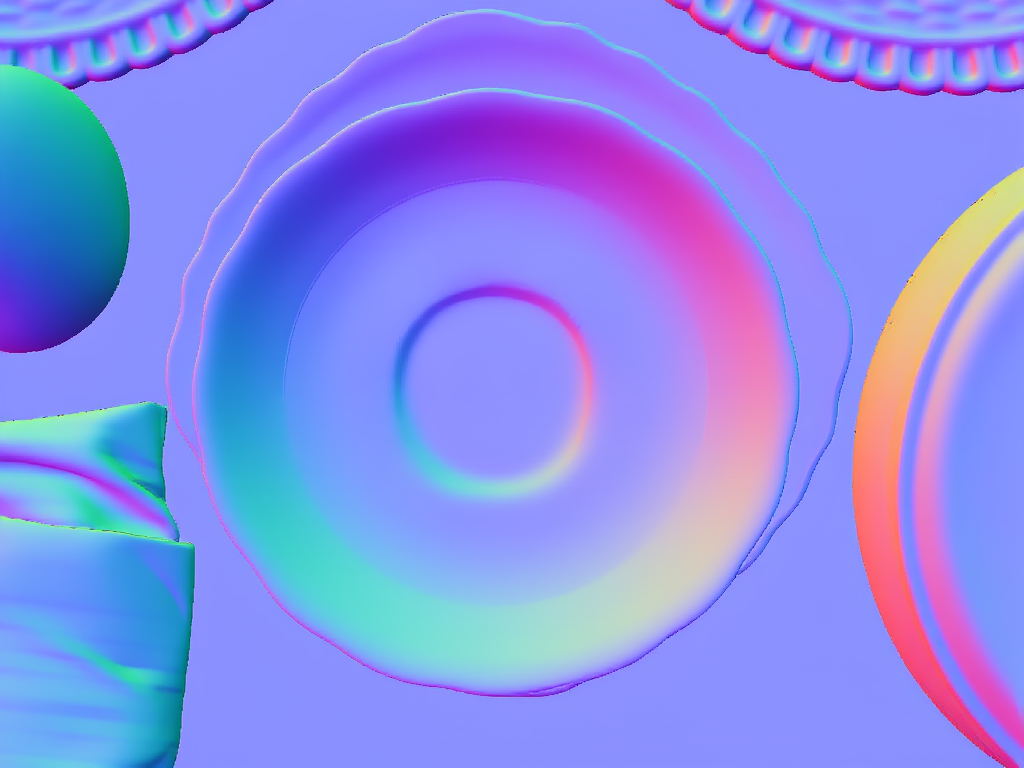} &
        \wiwcell{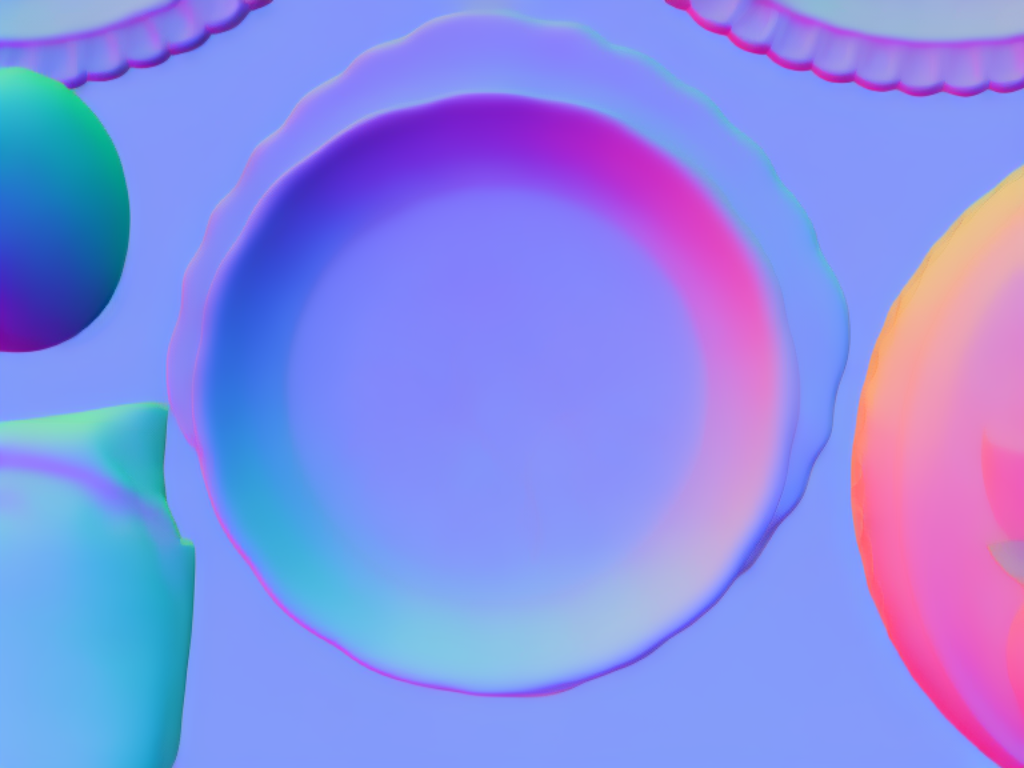} &
        \wiwcell{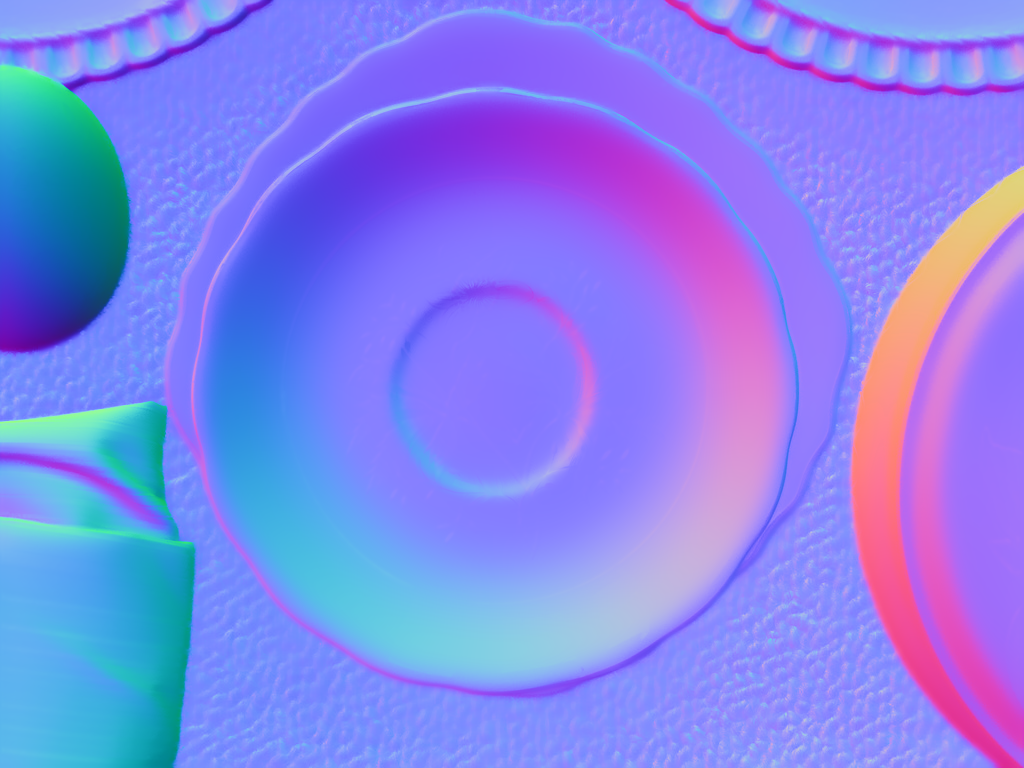} \\[0pt]
        \wiwcell{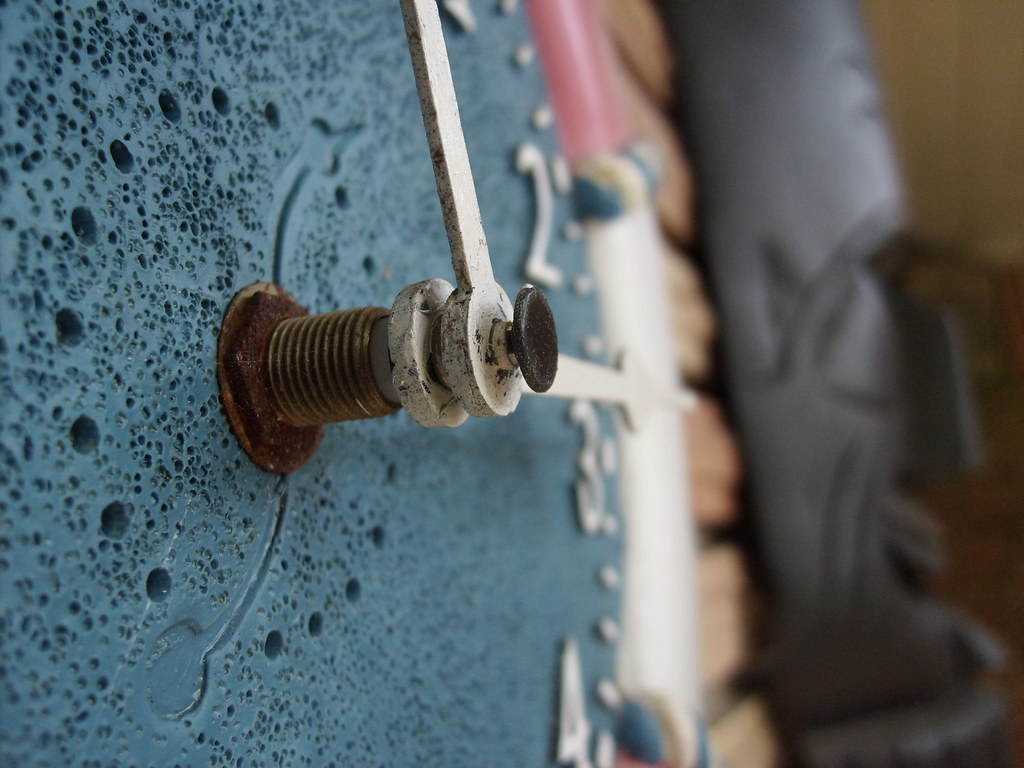} &
        \wiwcell{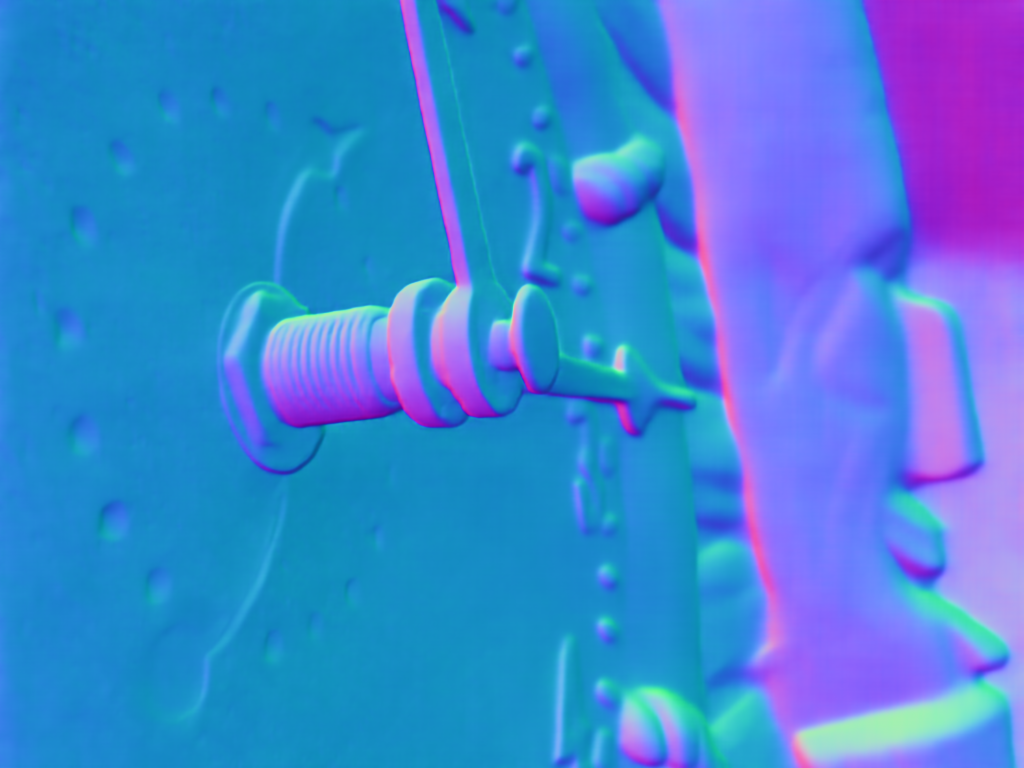} &
        \wiwcell{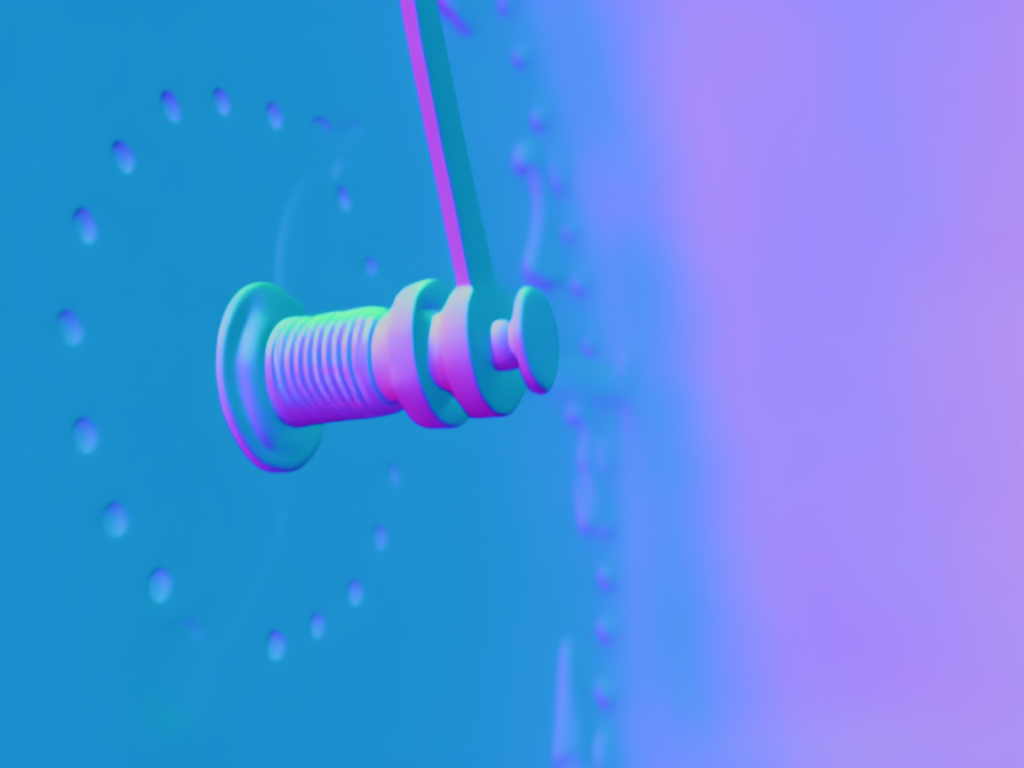} &
        \wiwcell{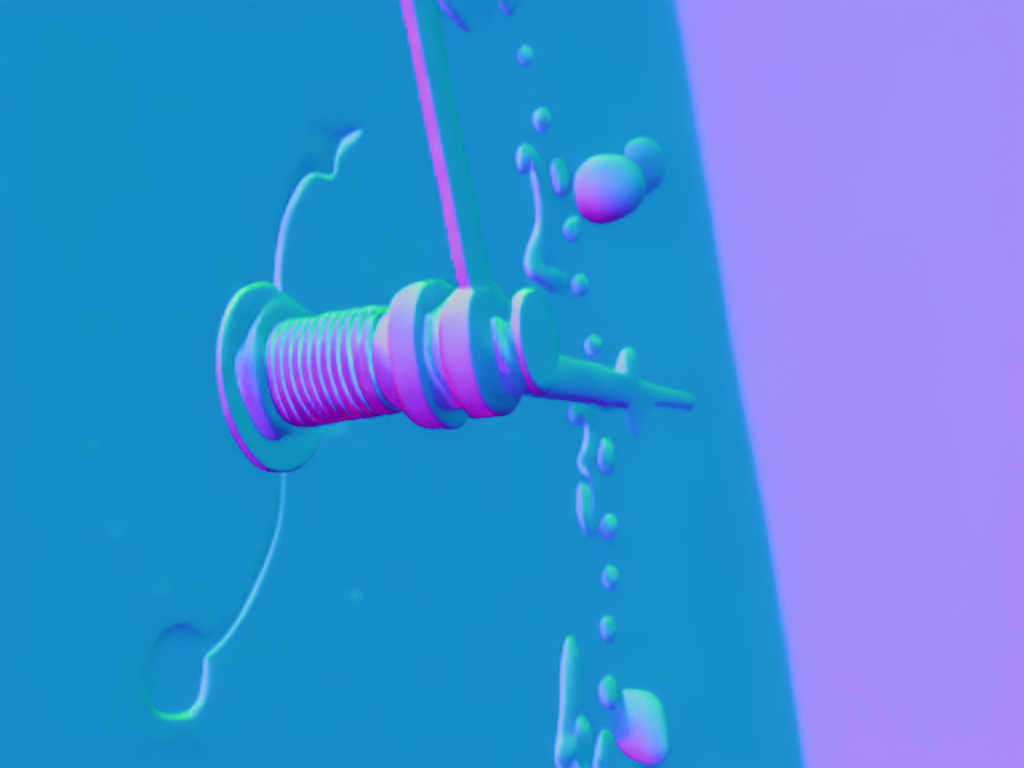} &
        \wiwcell{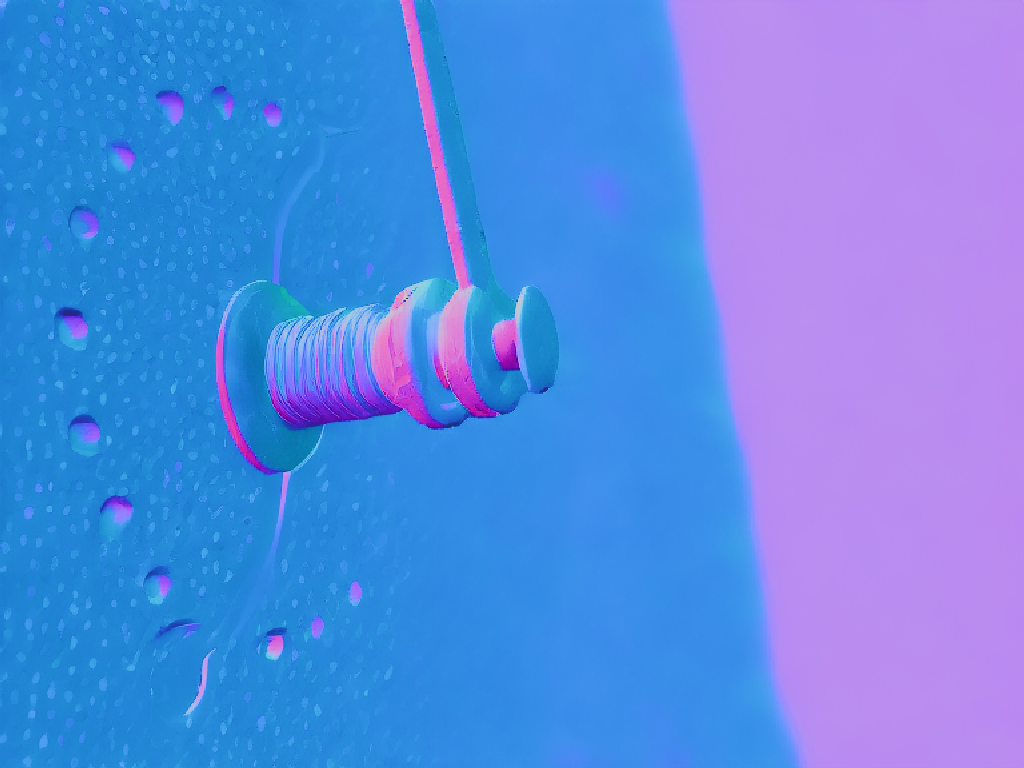} &
        \wiwcell{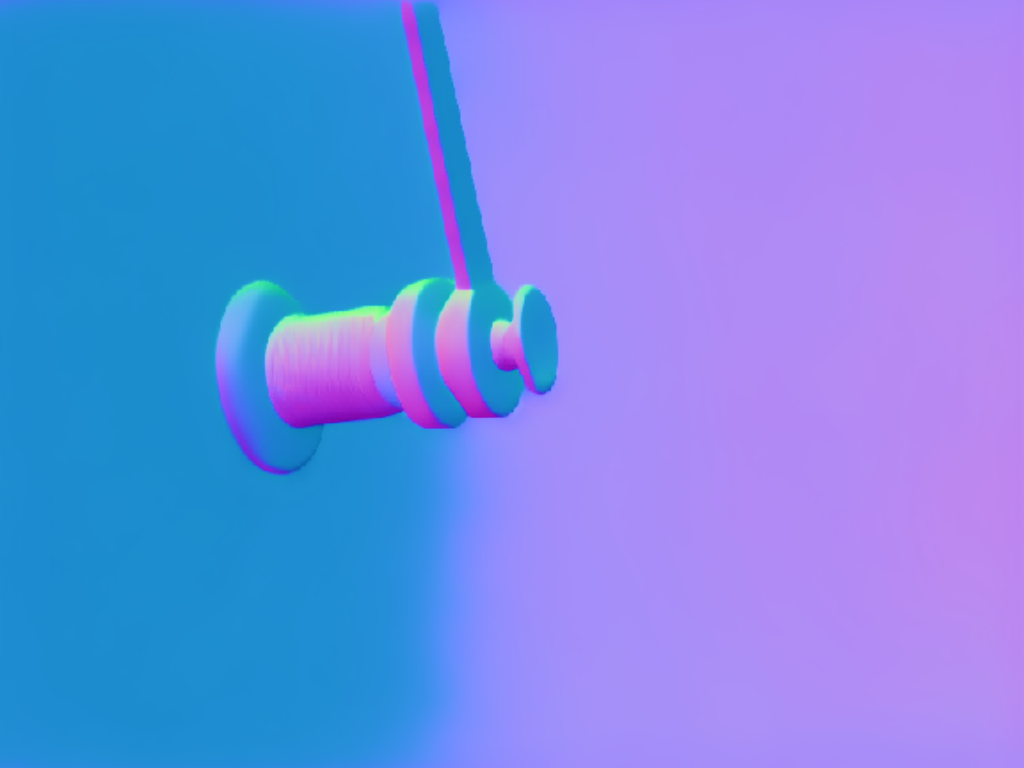} &
        \wiwcell{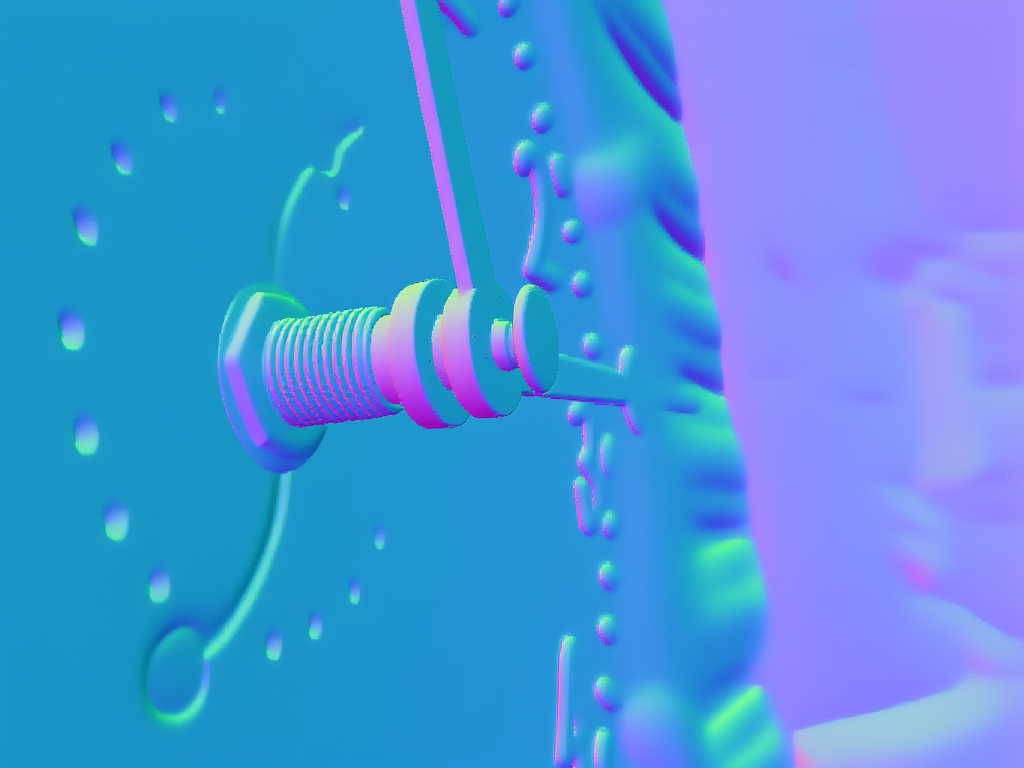} &
        \wiwcell{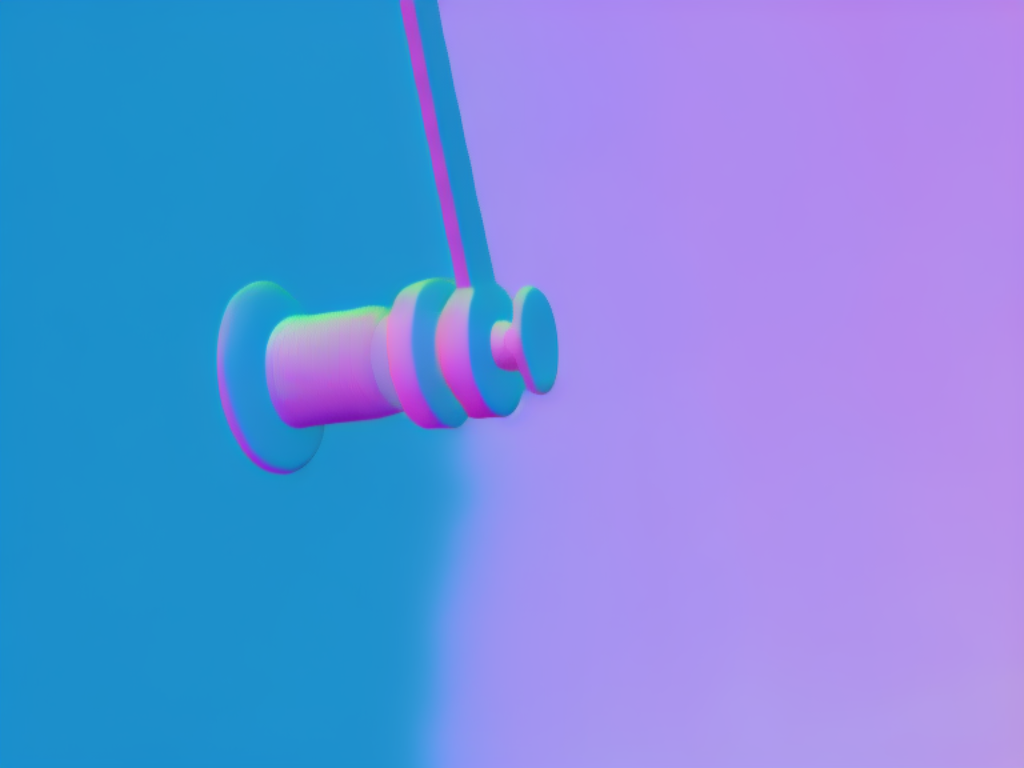} &
        \wiwcell{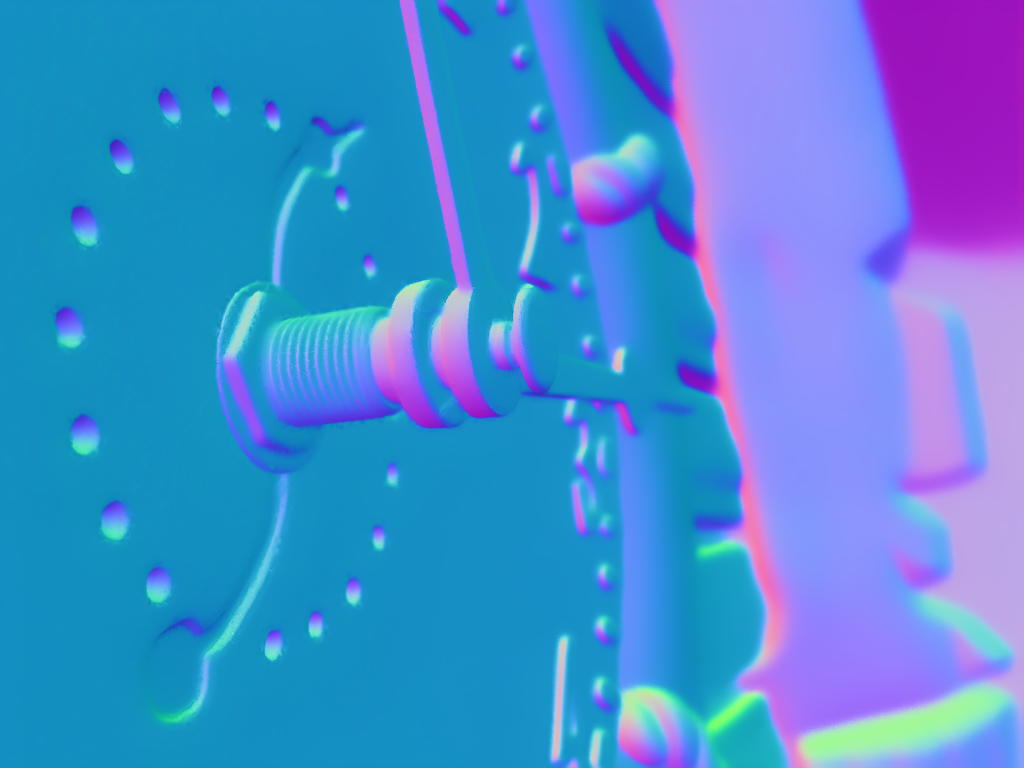} \\[0pt]
        \wiwcell{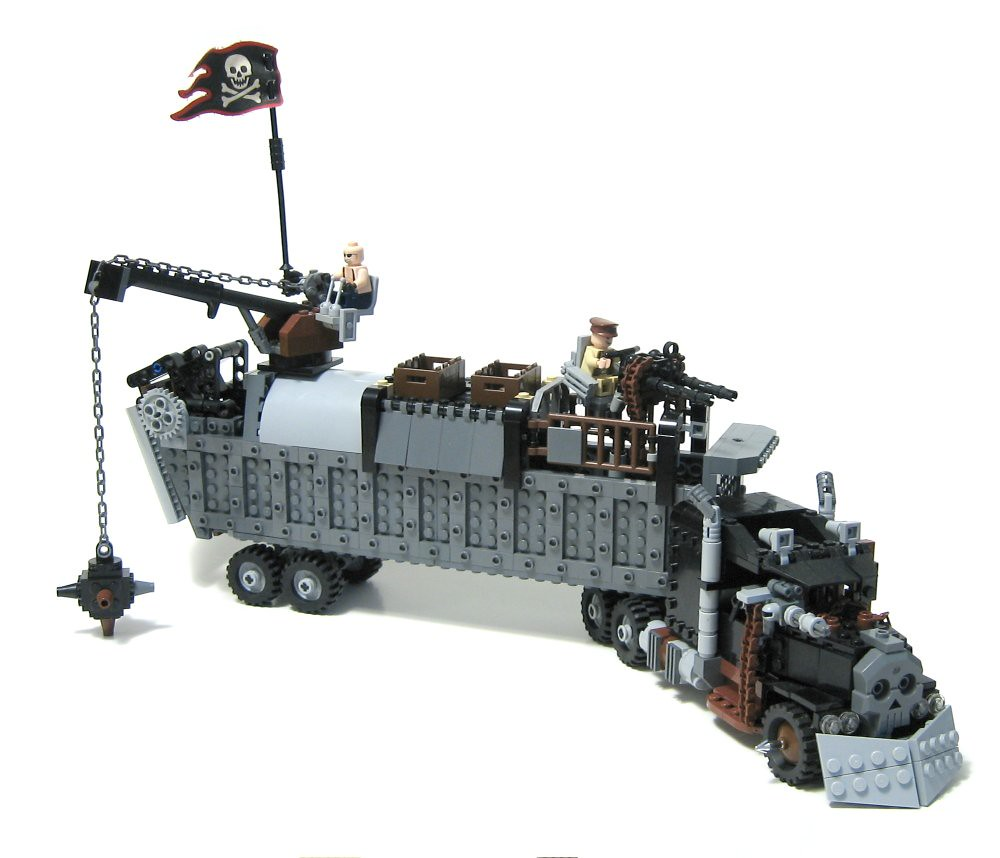} &
        \wiwcell{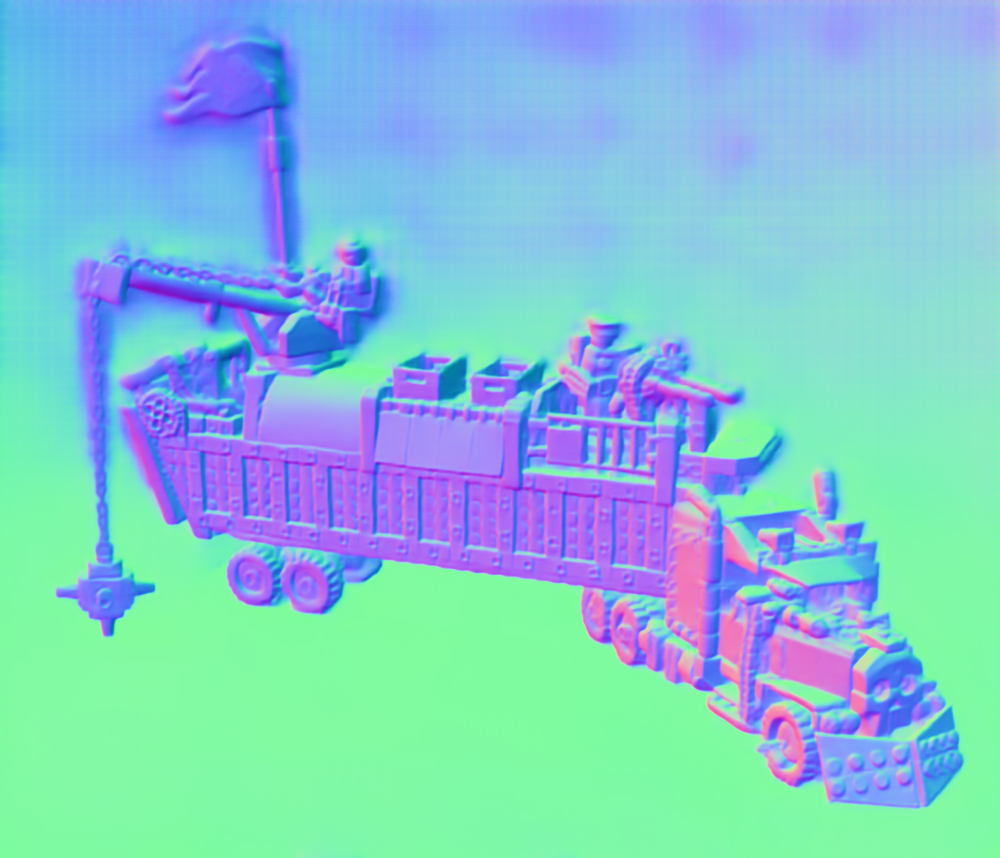} &
        \wiwcell{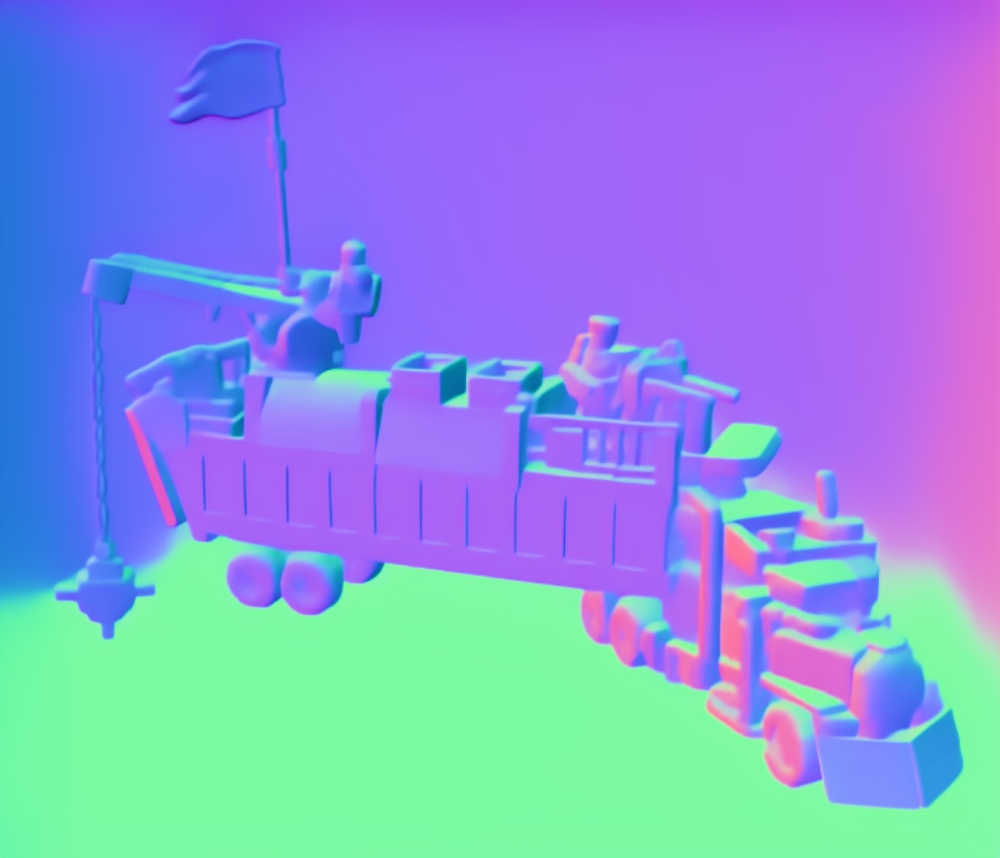} &
        \wiwcell{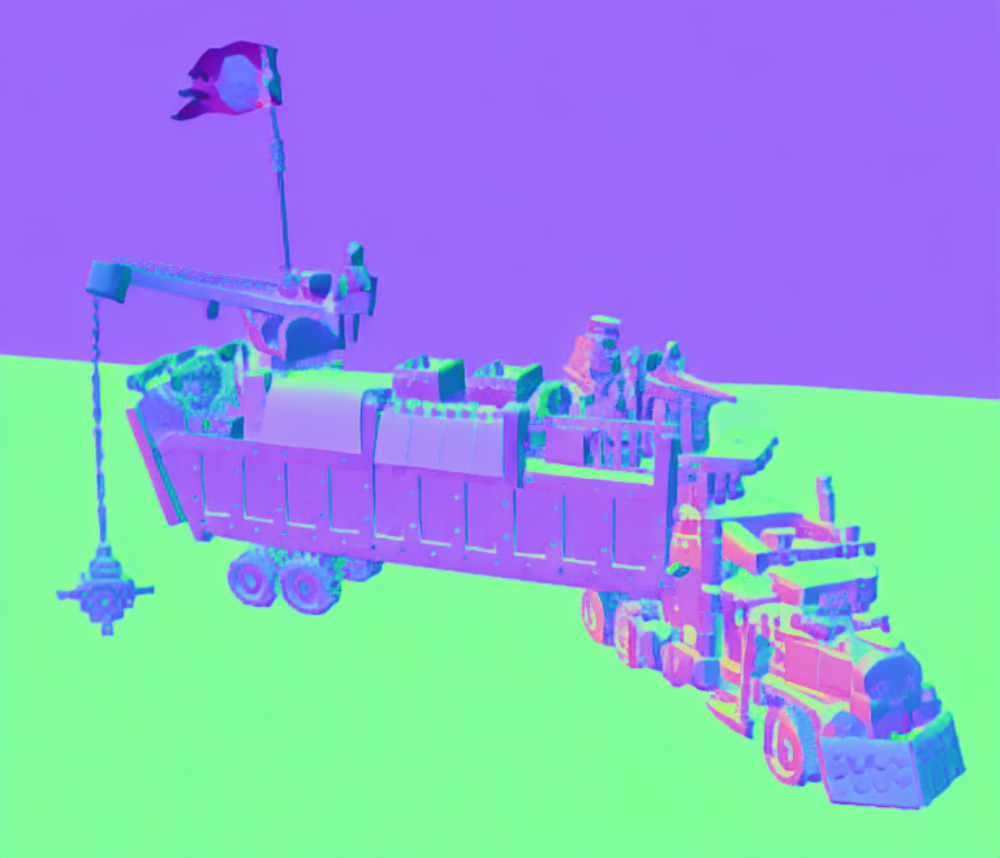} &
        \wiwcell{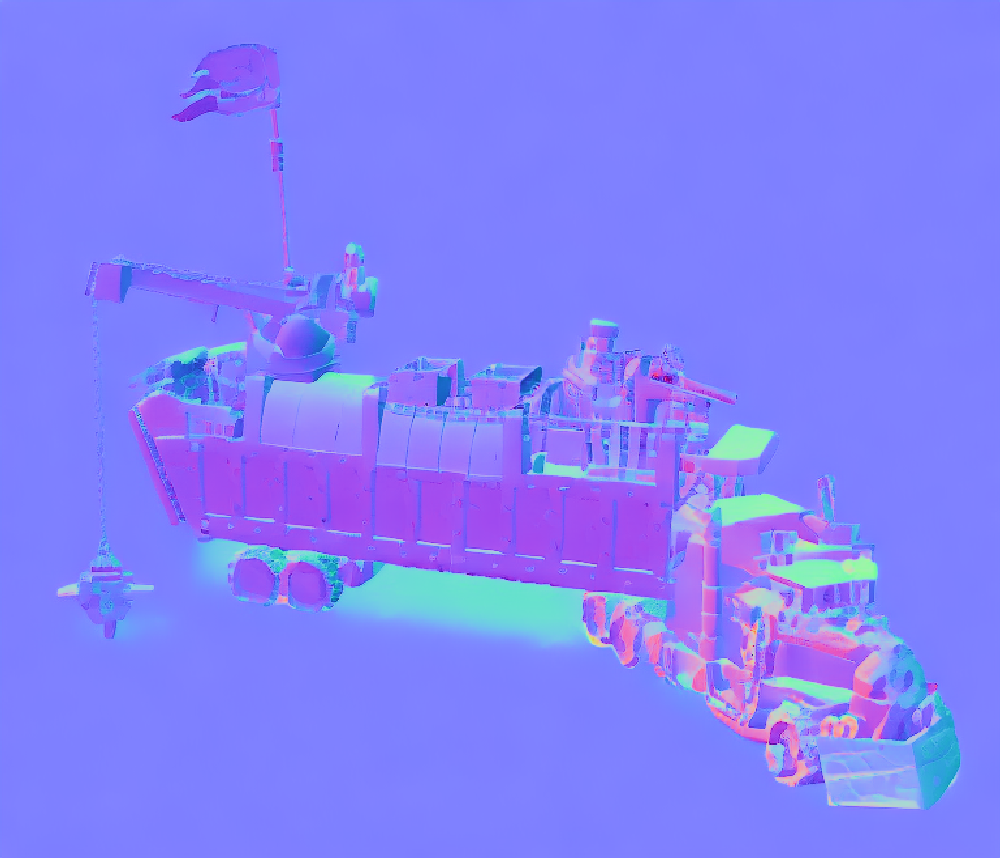} &
        \wiwcell{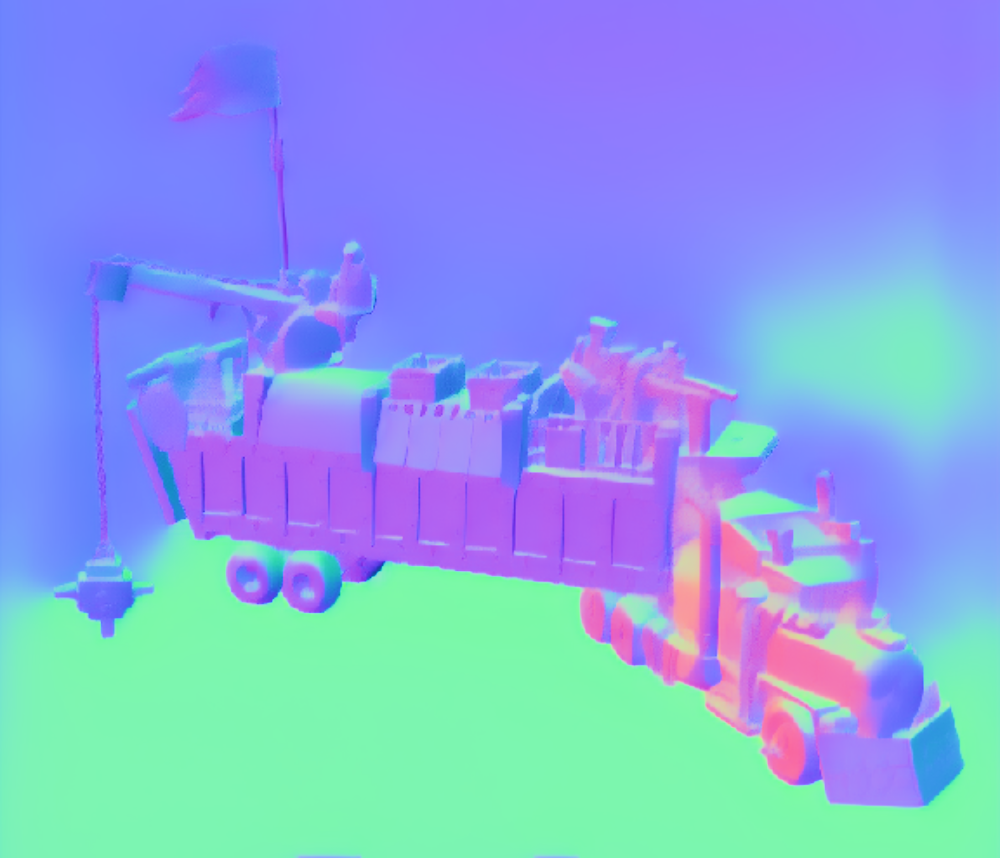} &
        \wiwcell{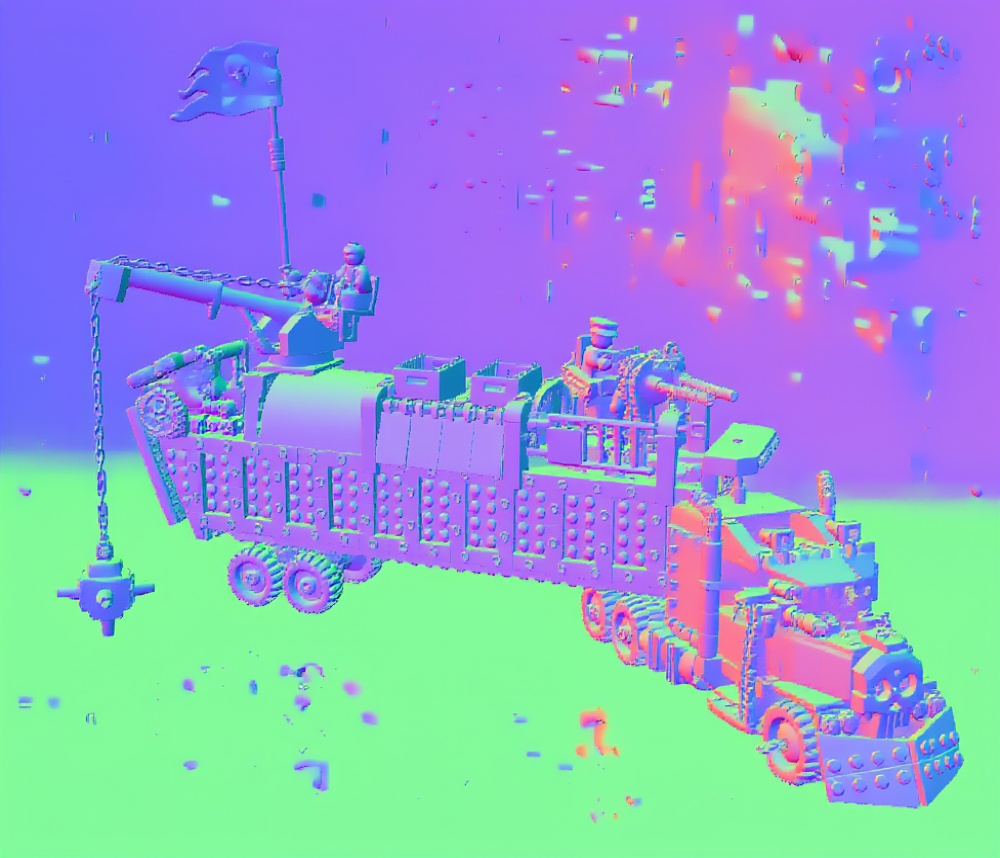} &
        \wiwcell{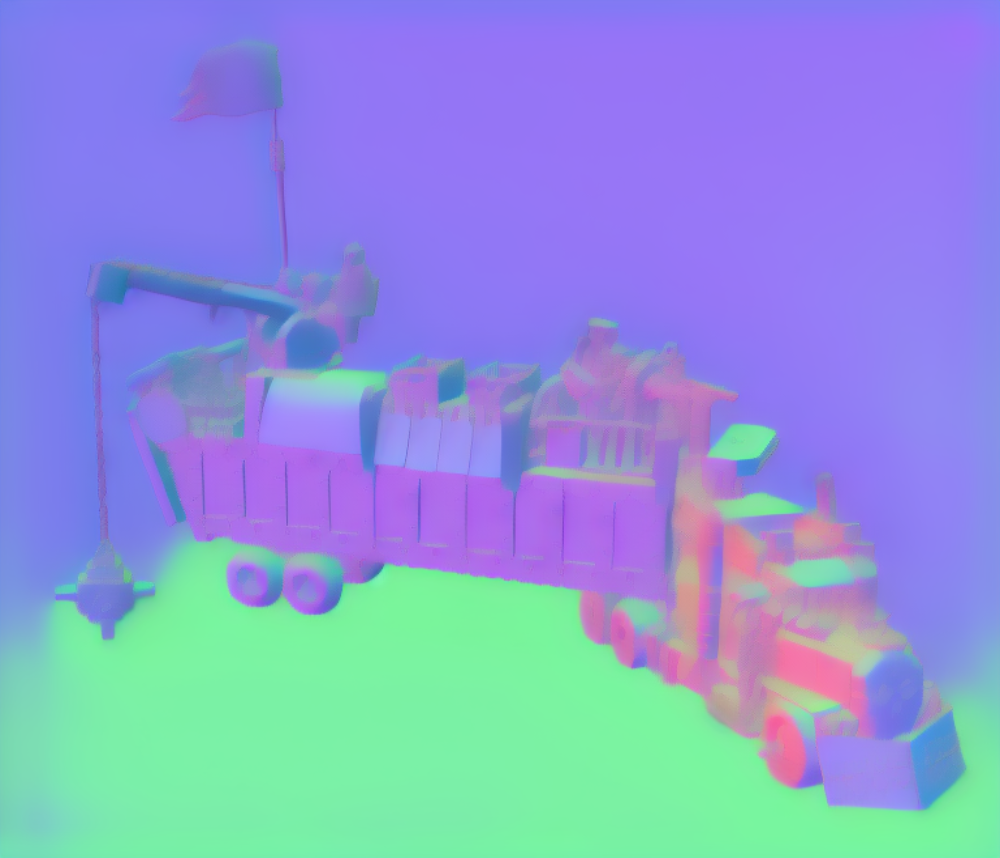} &
        \wiwcell{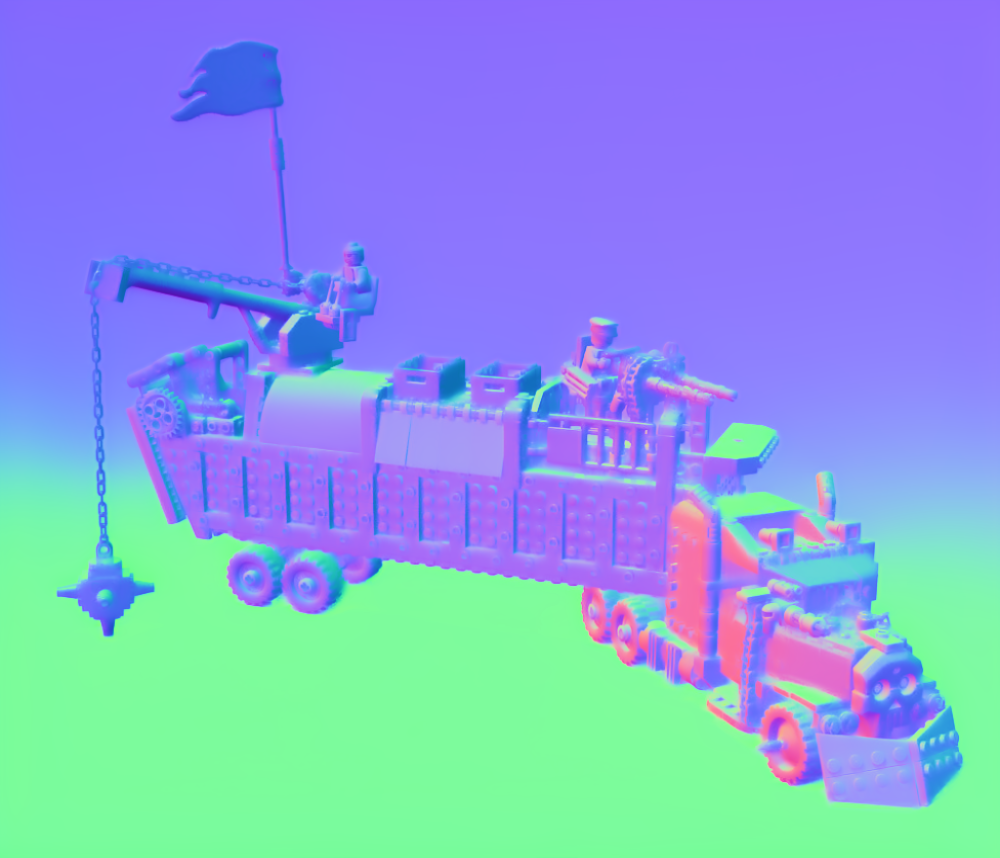} \\[0pt]
        \wiwcell{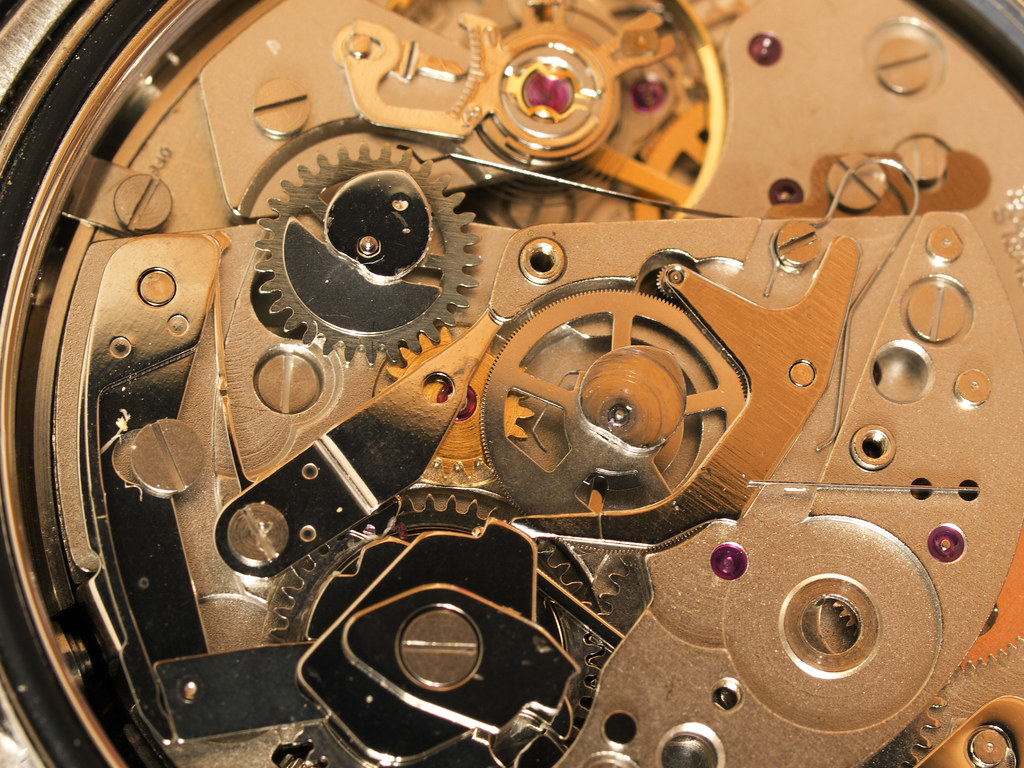} &
        \wiwcell{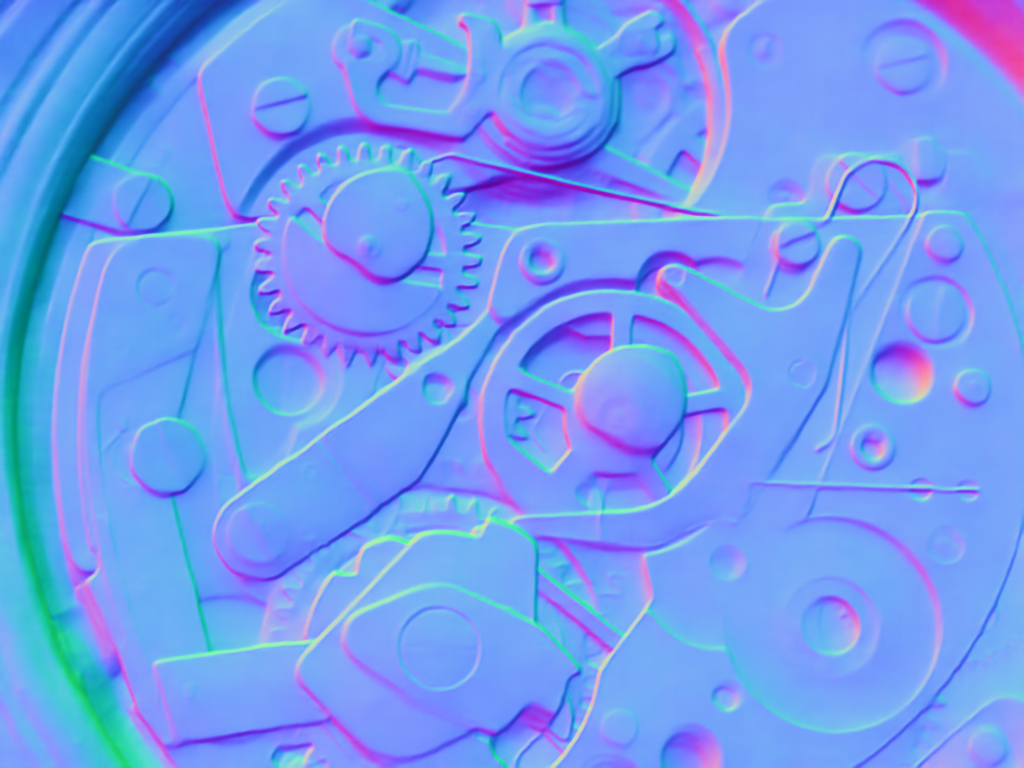} &
        \wiwcell{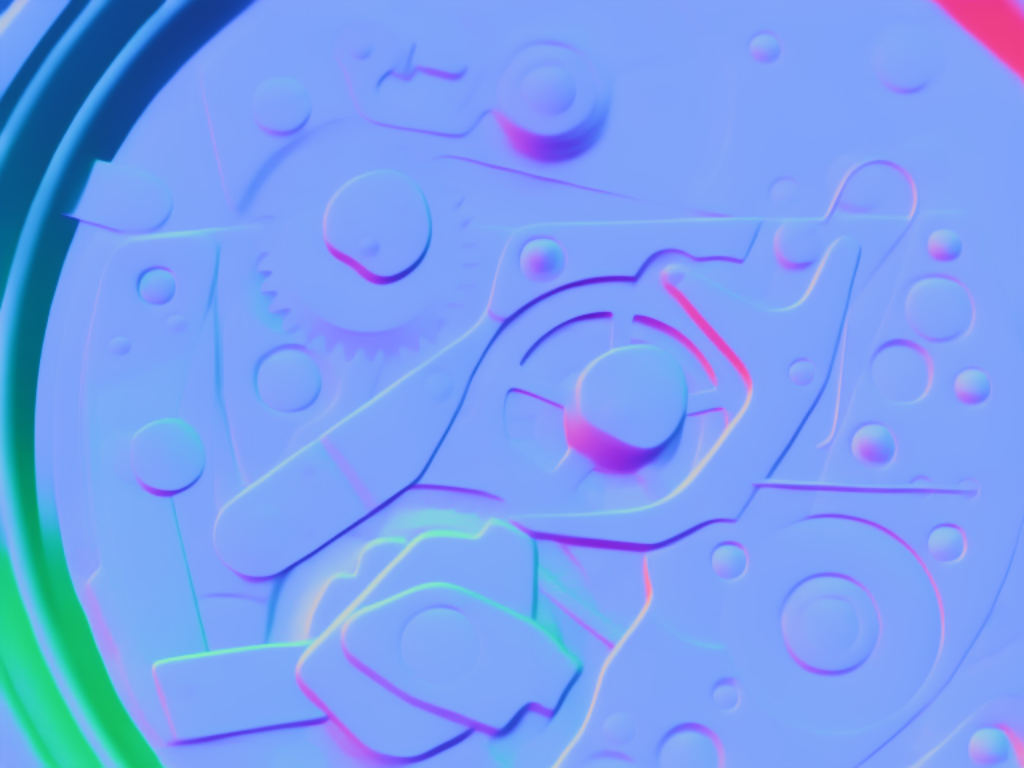} &
        \wiwcell{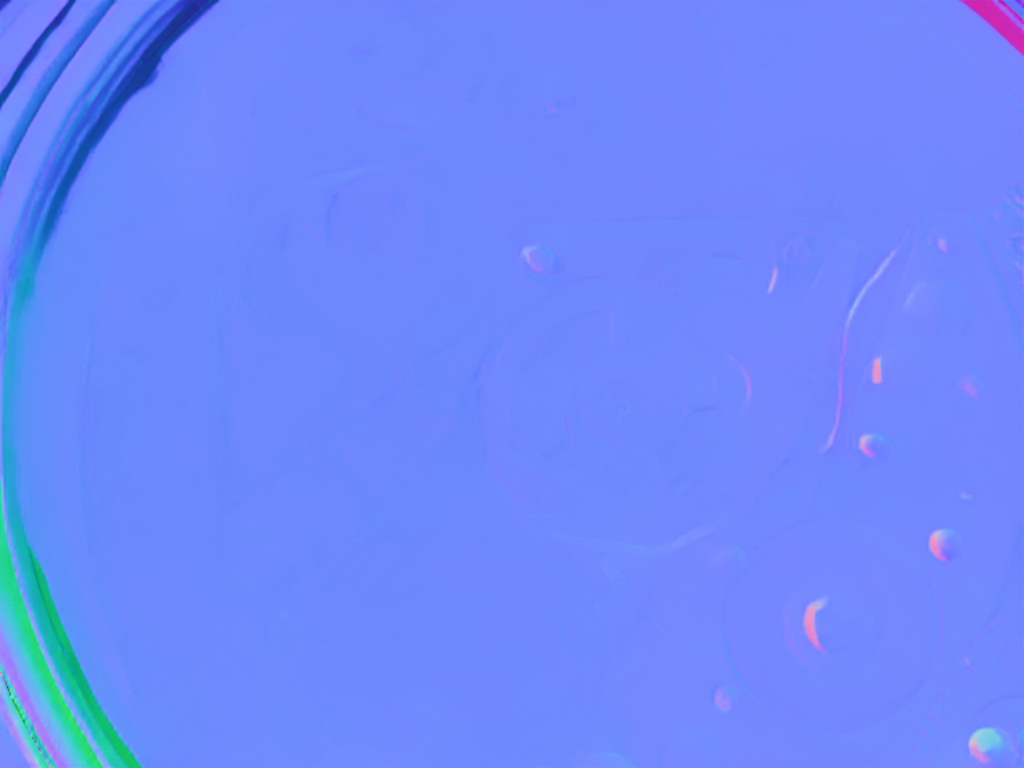} &
        \wiwcell{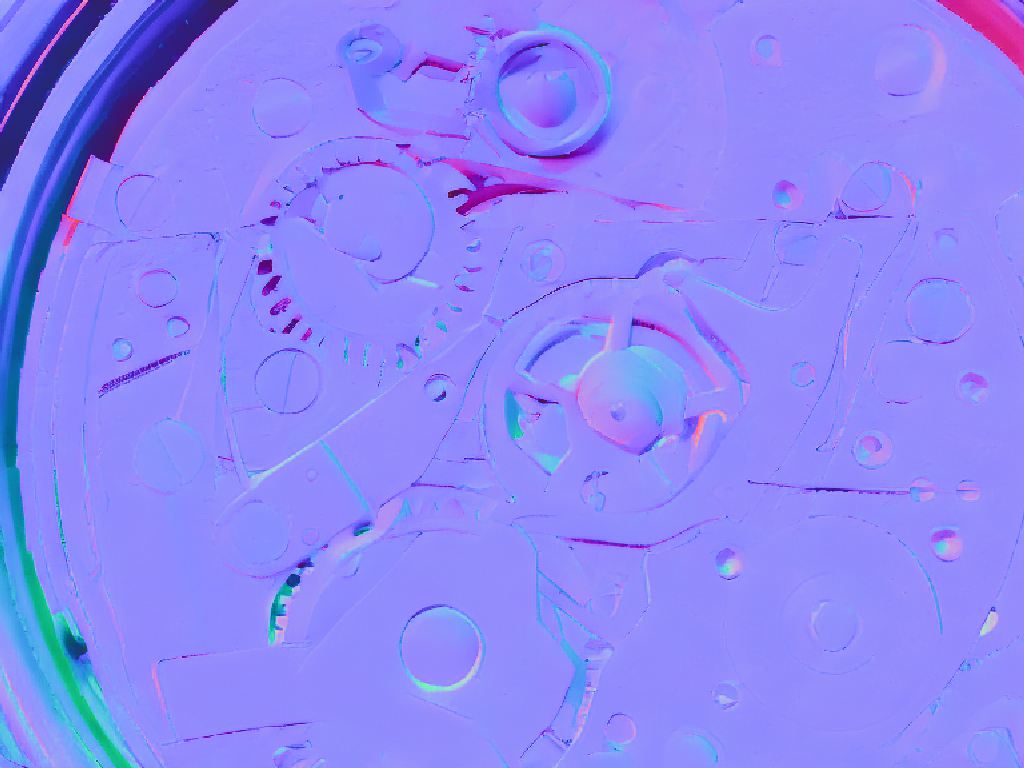} &
        \wiwcell{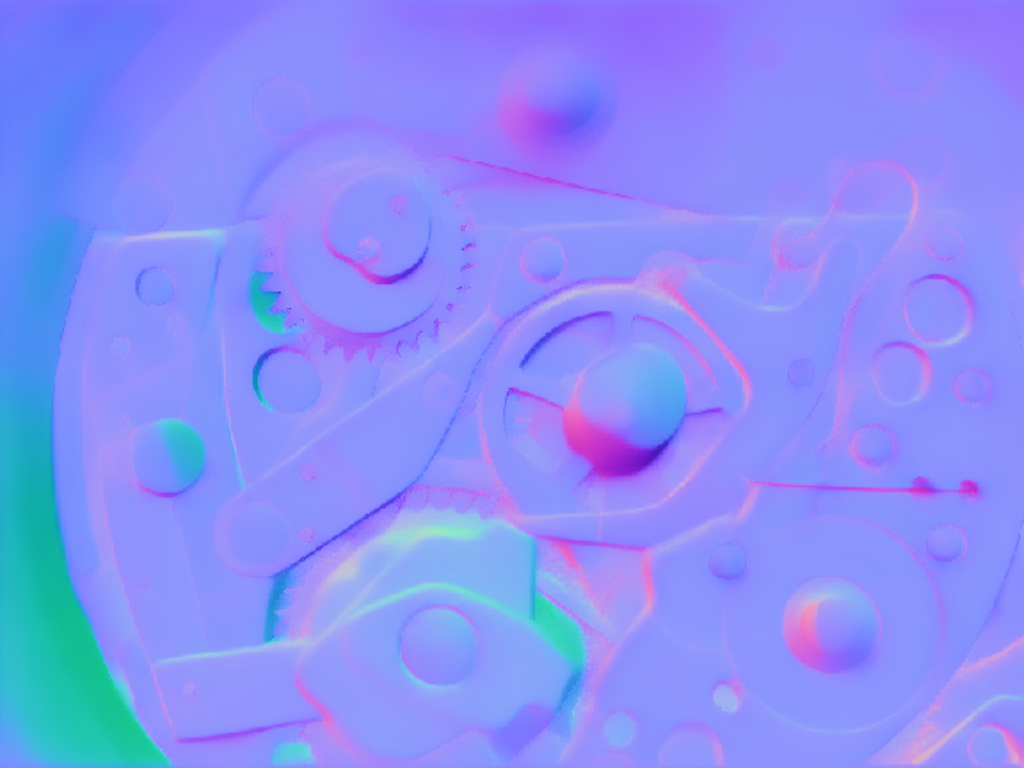} &
        \wiwcell{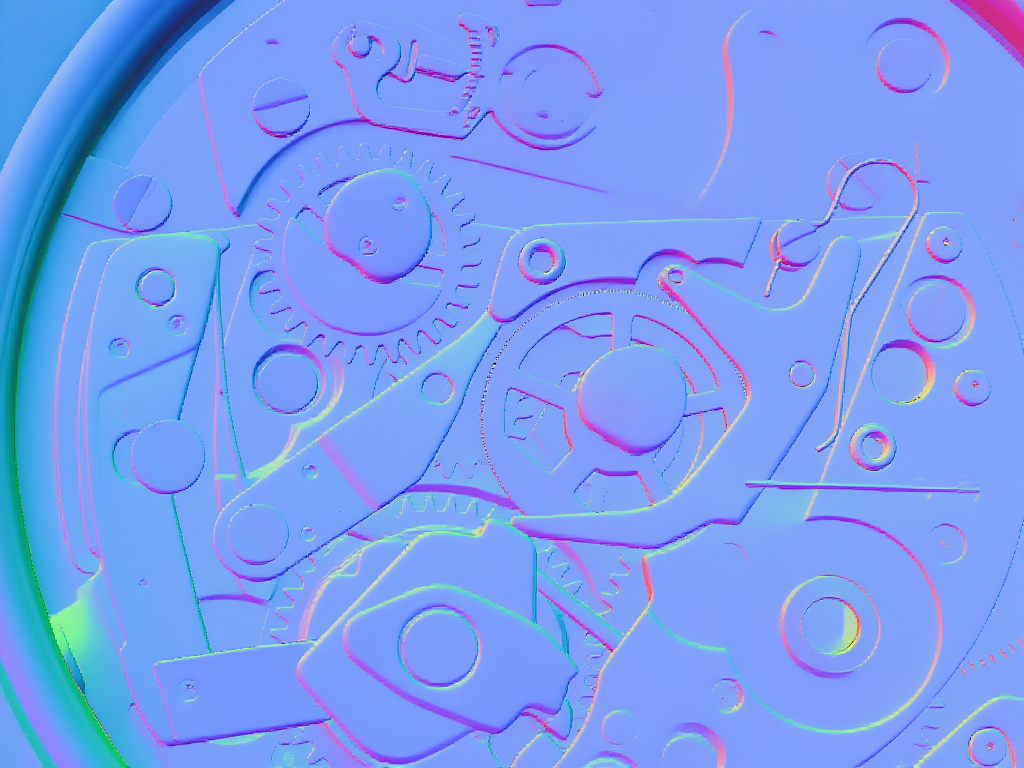} &
        \wiwcell{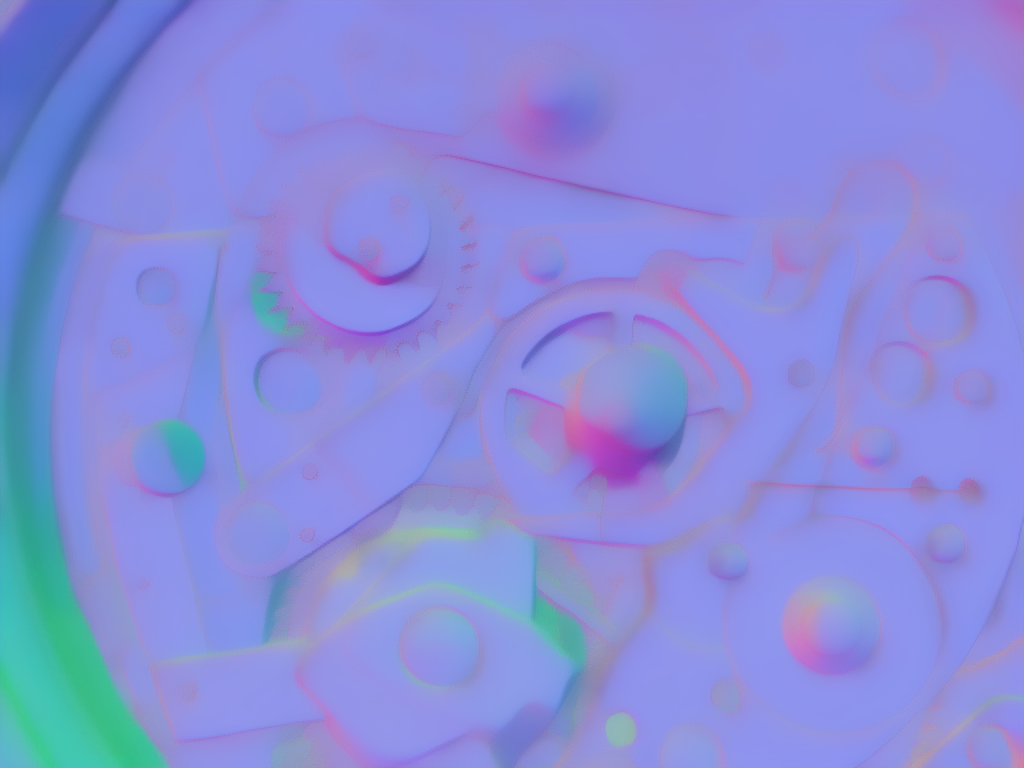} &
        \wiwcell{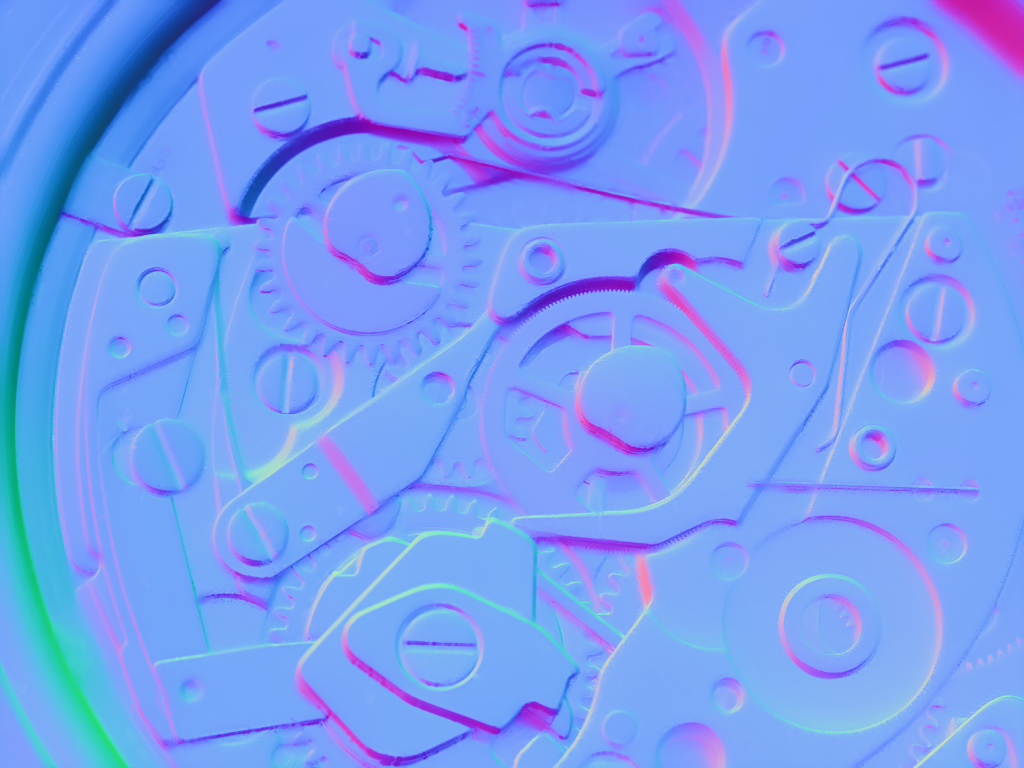}
    \end{tabular}
    \caption{\textbf{Qualitative results on unlabeled in-the-wild general objects.}
    We compare TransNormal-2 with representative baselines on examples from in-the-wild images.
    TransNormal-2 produces coherent object-scale geometry while preserving fine structures on curved ceramic surfaces and mechanical parts. Please zoom in \faSearchPlus~for details. (\S~\ref{par:qual_comparison})}
    \label{fig:in_the_wild_general}
    \vspace{-2mm}
\end{figure*}

\noindent\textbf{Comparison with Baselines.}\label{par:qual_comparison}
\Figref{fig:in_the_wild_general} presents qualitative comparisons on unlabeled in-the-wild general objects, including ceramic surfaces and mechanical assemblies. Because these images do not provide ground-truth normal annotations, we show only predicted normal maps without error rows. The examples test whether a model preserves object-scale shape while resolving fine geometric structures such as gear teeth, watch mechanisms, and curved pottery surfaces.

\begin{figure*}[t]
    \centering
    \setlength{\tabcolsep}{1pt}
    \renewcommand{\arraystretch}{0.6}
    \newcommand{\imgw}{0.105\textwidth}
    \begin{tabular}{@{}c@{\hspace{1pt}}cc|ccccccc@{}}
        & \makebox[\imgw][c]{\scriptsize Input/Mask}
        & \makebox[\imgw][c]{\scriptsize GT}
        & \makebox[\imgw][c]{\scriptsize Lotus}
        & \makebox[\imgw][c]{\scriptsize Lotus-2}
        & \makebox[\imgw][c]{\scriptsize MoGe-2}
        & \makebox[\imgw][c]{\scriptsize E2E-FT}
        & \makebox[\imgw][c]{\scriptsize GenPercept}
        & \makebox[\imgw][c]{\scriptsize TransNormal}
        & \makebox[\imgw][c]{\scriptsize \textbf{Ours}} \\[2pt]
        \multirow{2}{*}[3.5ex]{\rotatebox{90}{\small TN-Syn}} &
        \includegraphics[width=\imgw]{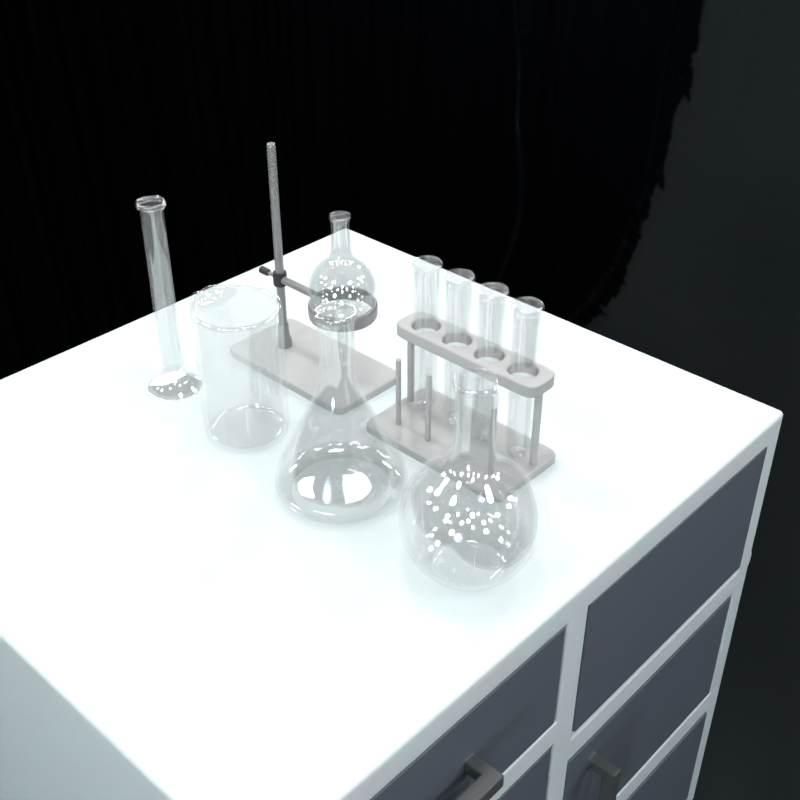} &
        \includegraphics[width=\imgw]{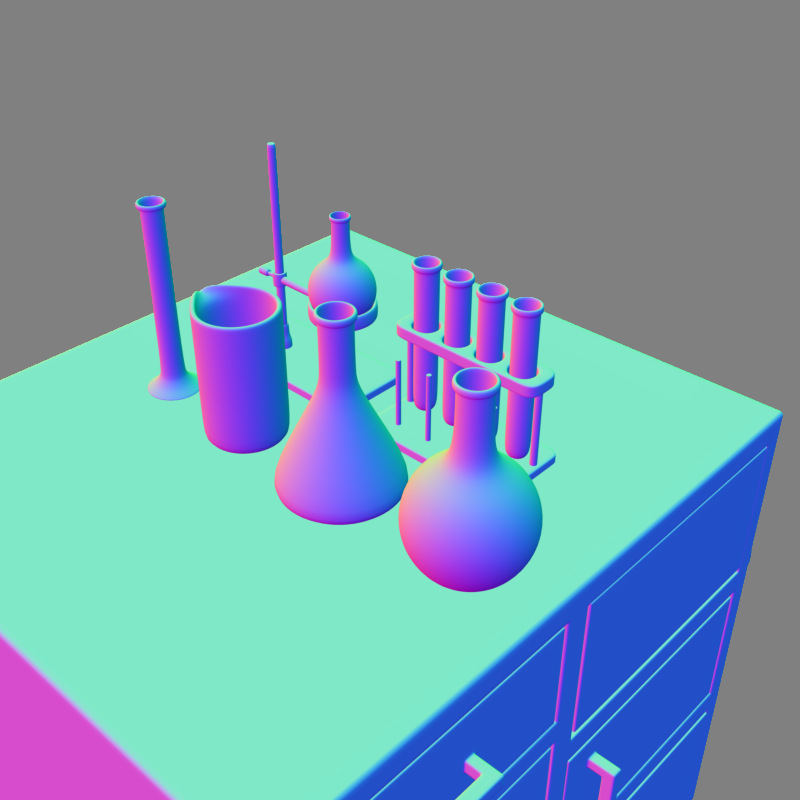} &
        \includegraphics[width=\imgw]{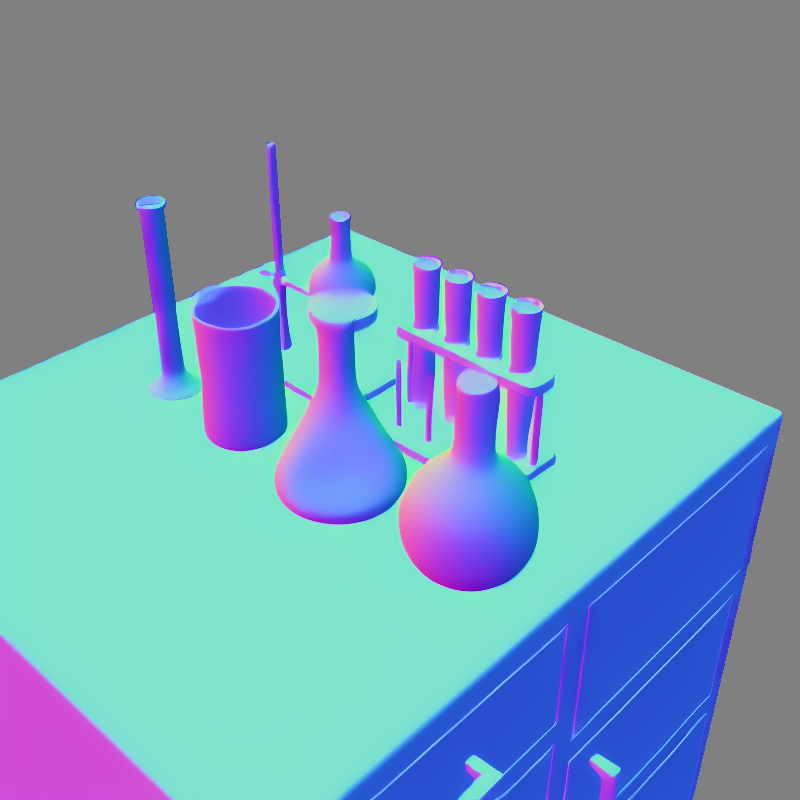} &
        \includegraphics[width=\imgw]{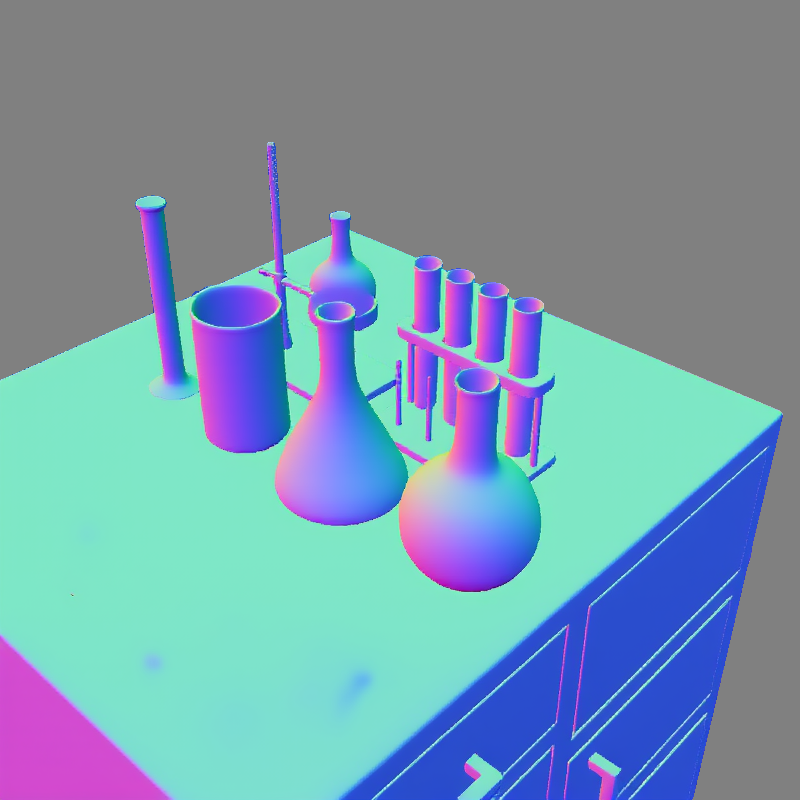} &
        \includegraphics[width=\imgw]{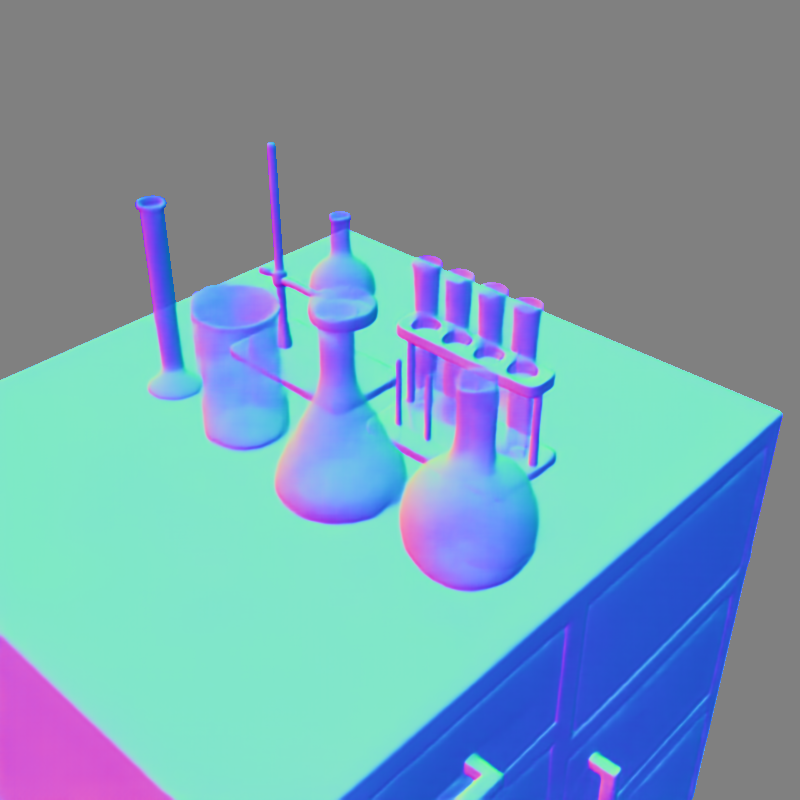} &
        \includegraphics[width=\imgw]{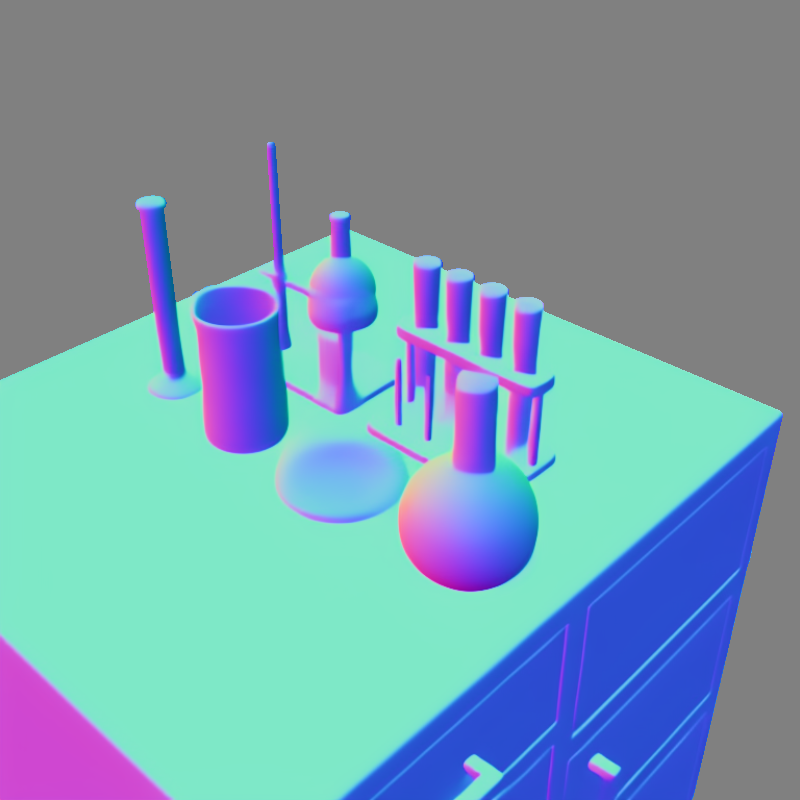} &
        \includegraphics[width=\imgw]{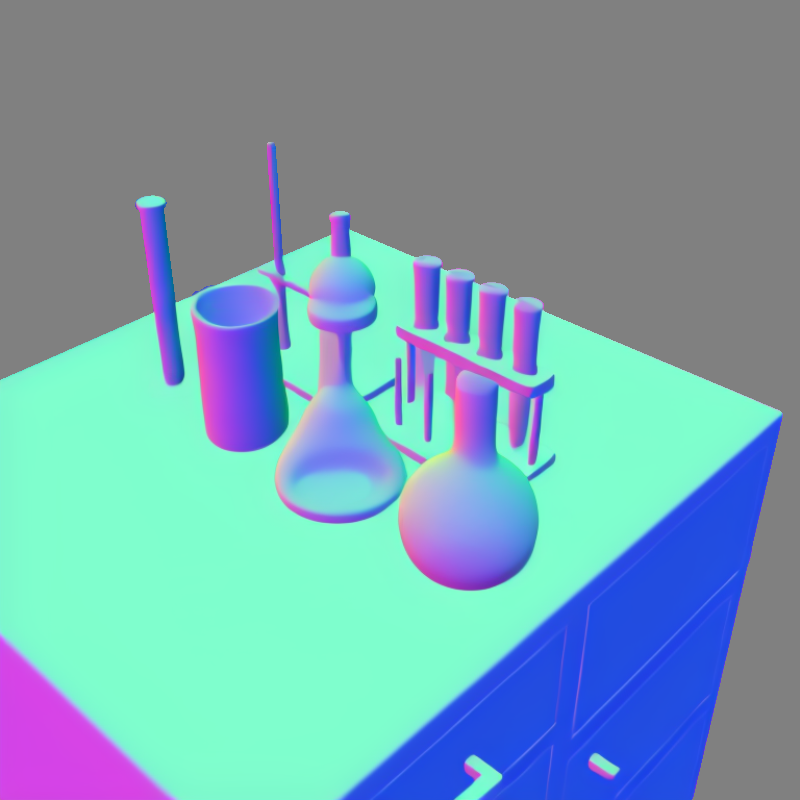} &
        \includegraphics[width=\imgw]{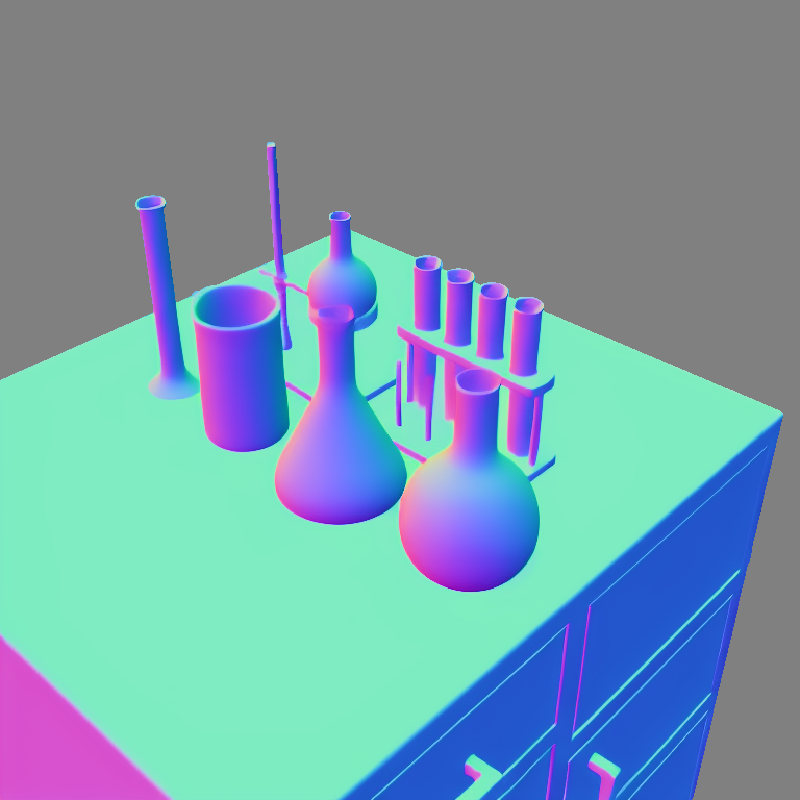} &
        \includegraphics[width=\imgw]{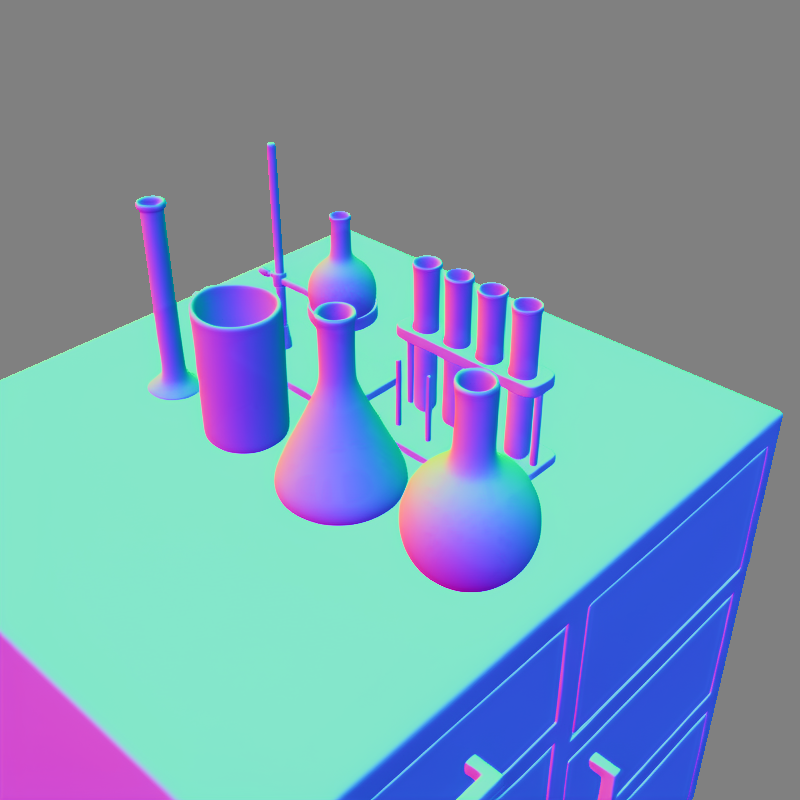} \\
        &
        \includegraphics[width=\imgw]{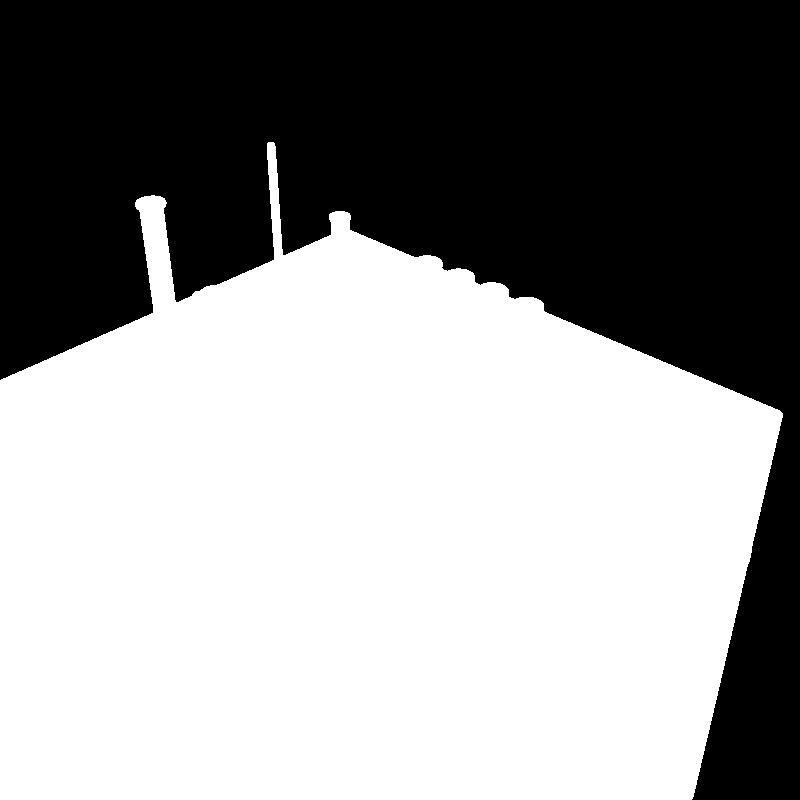} &
        \includegraphics[width=\imgw]{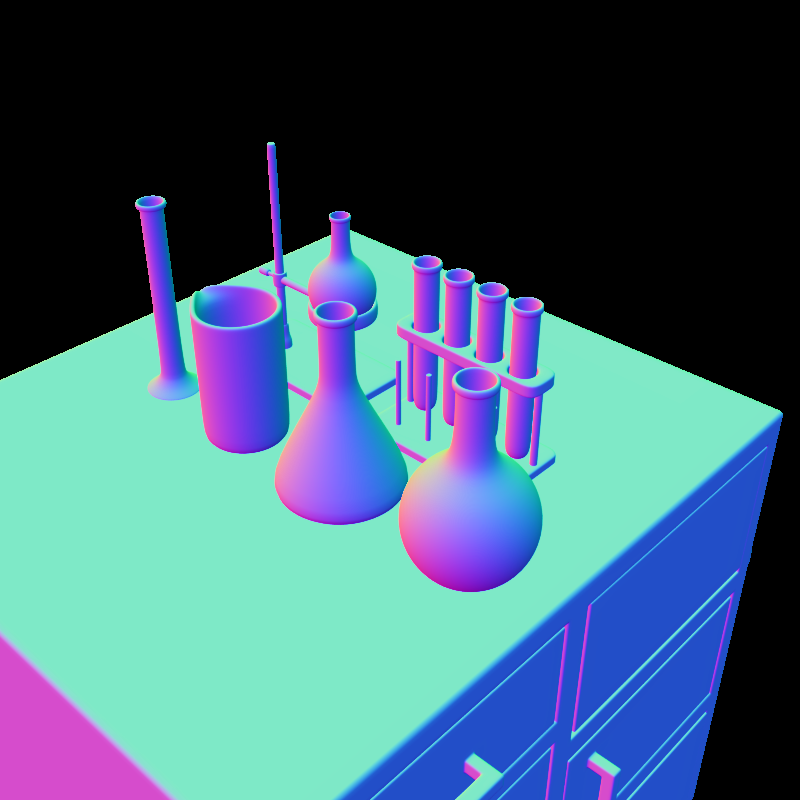} &
        \includegraphics[width=\imgw]{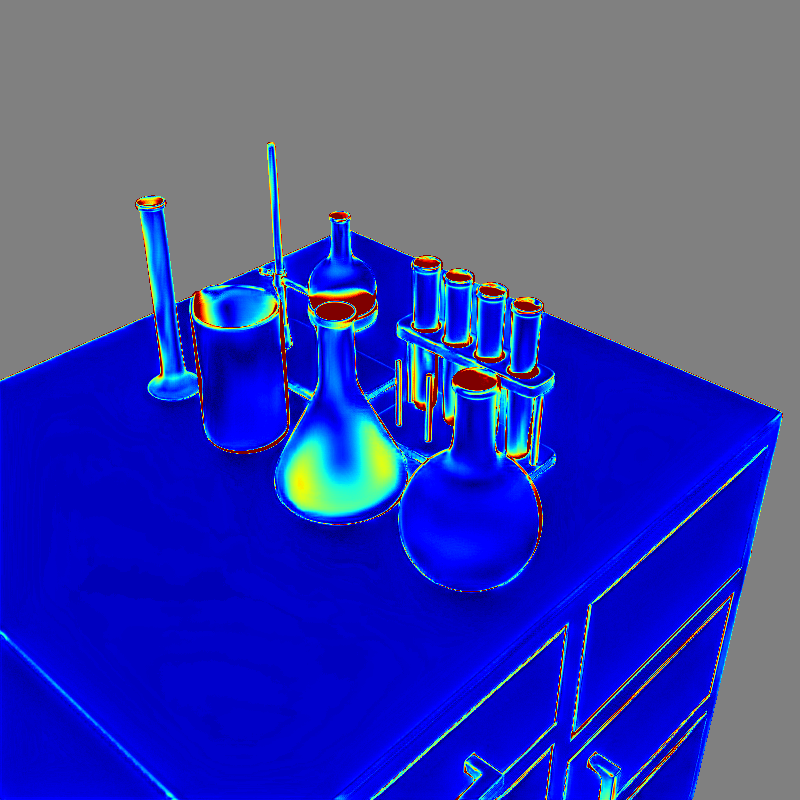} &
        \includegraphics[width=\imgw]{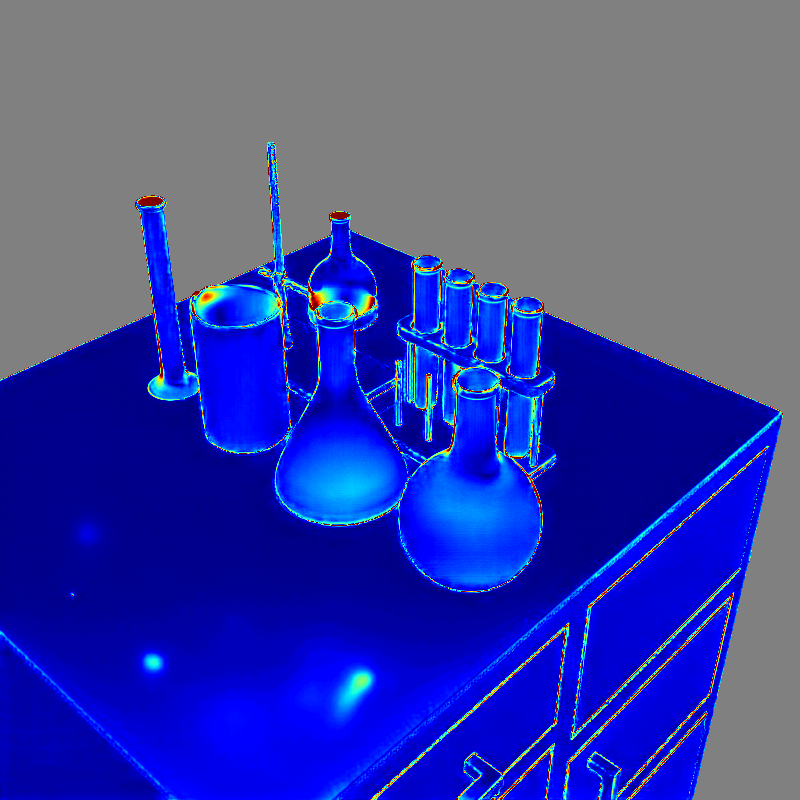} &
        \includegraphics[width=\imgw]{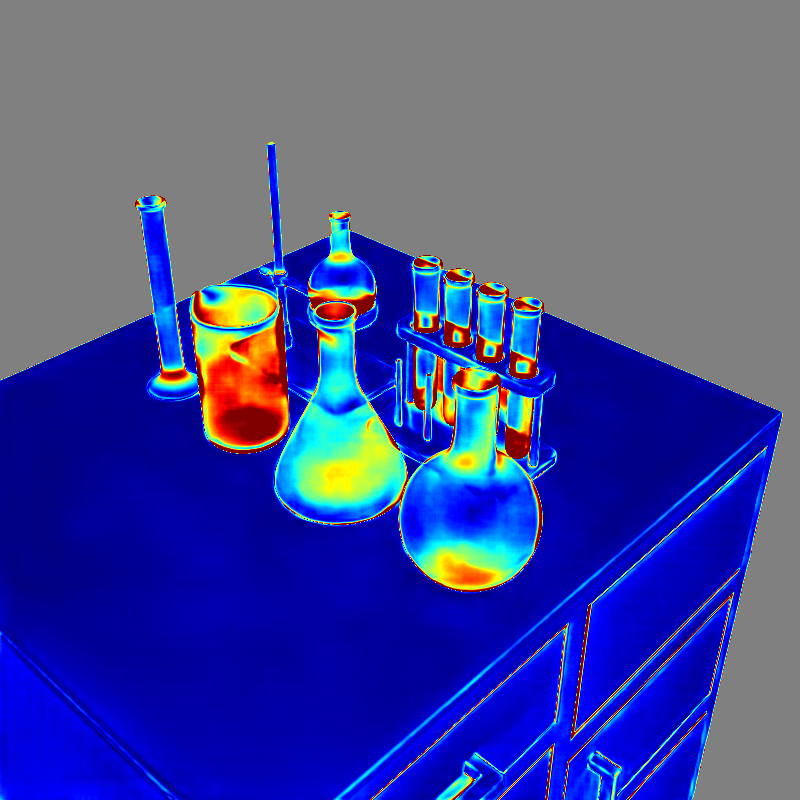} &
        \includegraphics[width=\imgw]{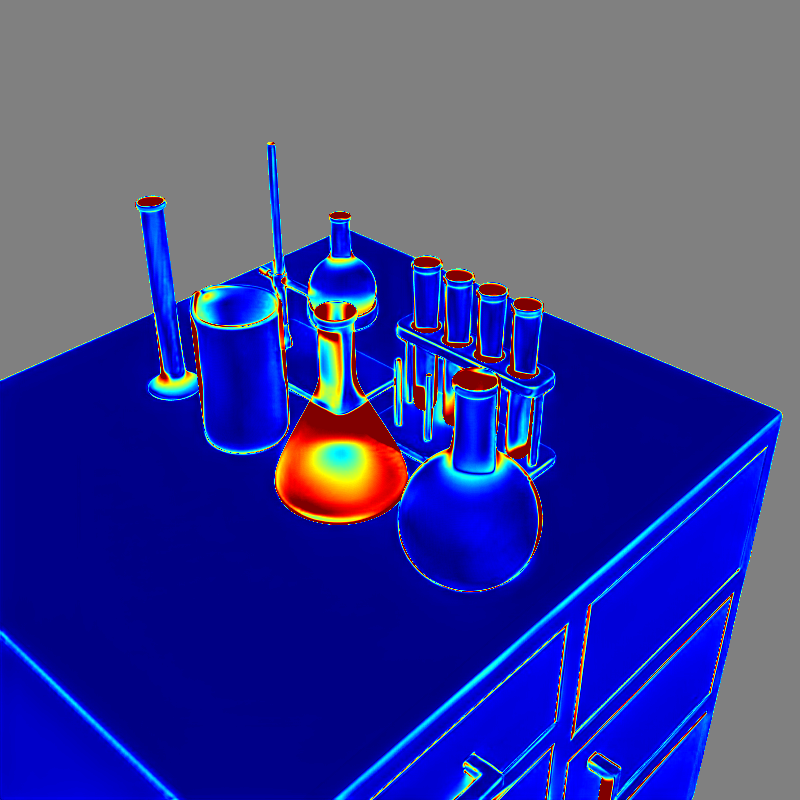} &
        \includegraphics[width=\imgw]{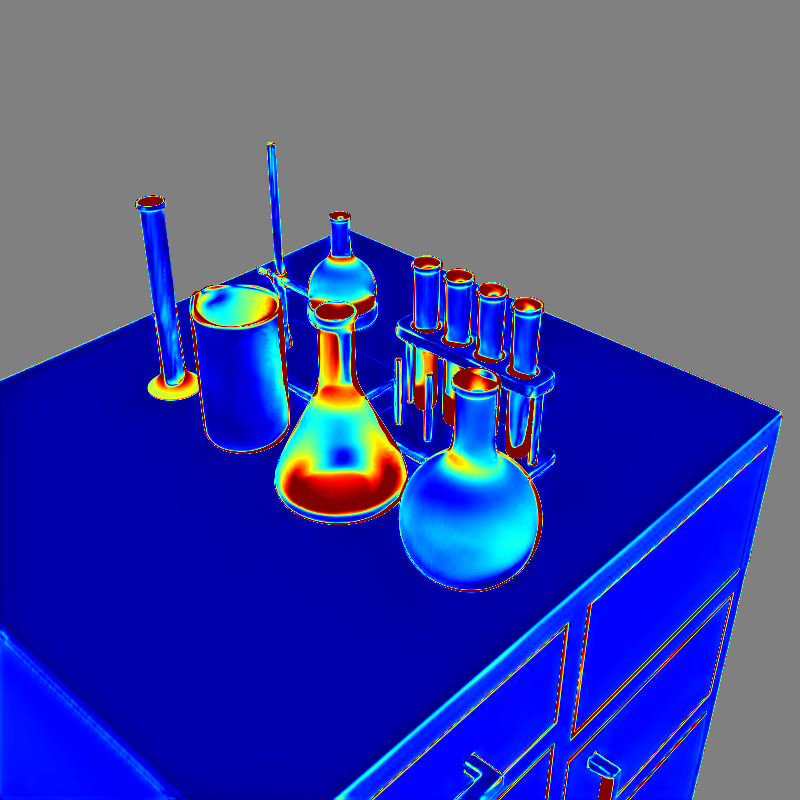} &
        \includegraphics[width=\imgw]{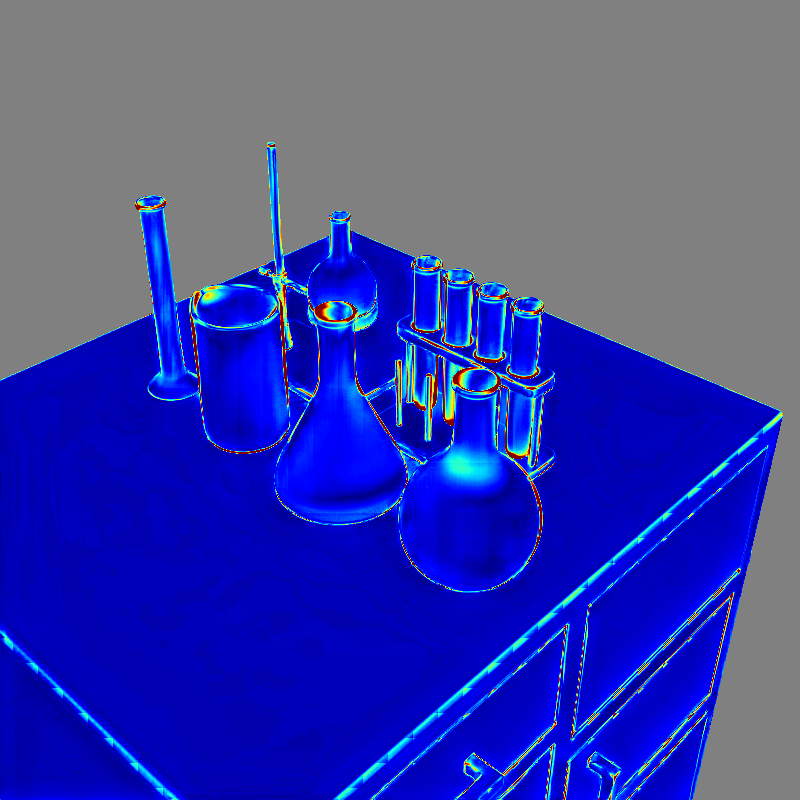} &
        \includegraphics[width=\imgw]{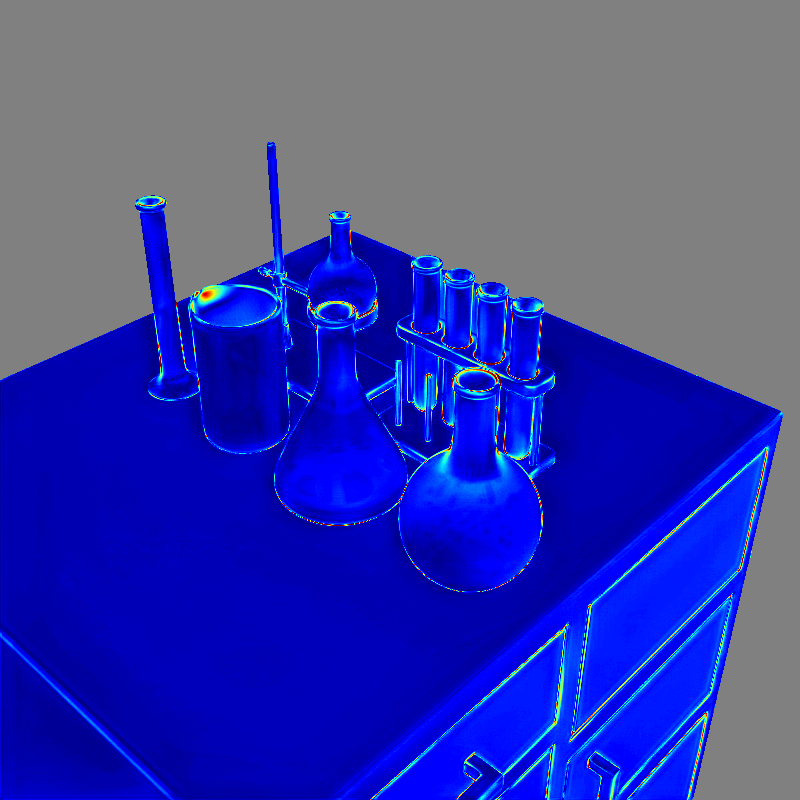} \\[2pt]
        \multirow{2}{*}[3.5ex]{\rotatebox{90}{\small ClearGrasp}} &
        \includegraphics[width=\imgw]{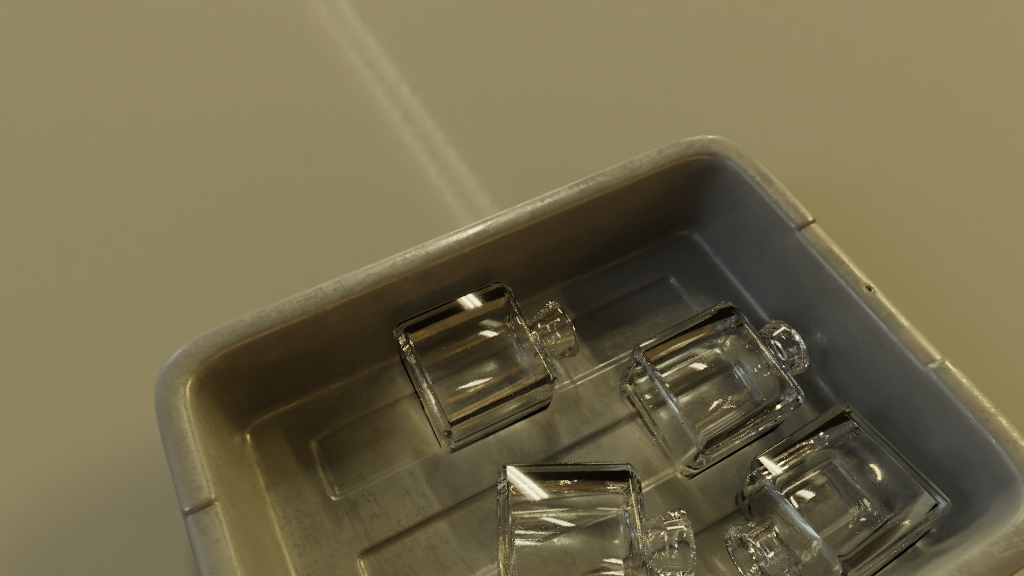} &
        \includegraphics[width=\imgw]{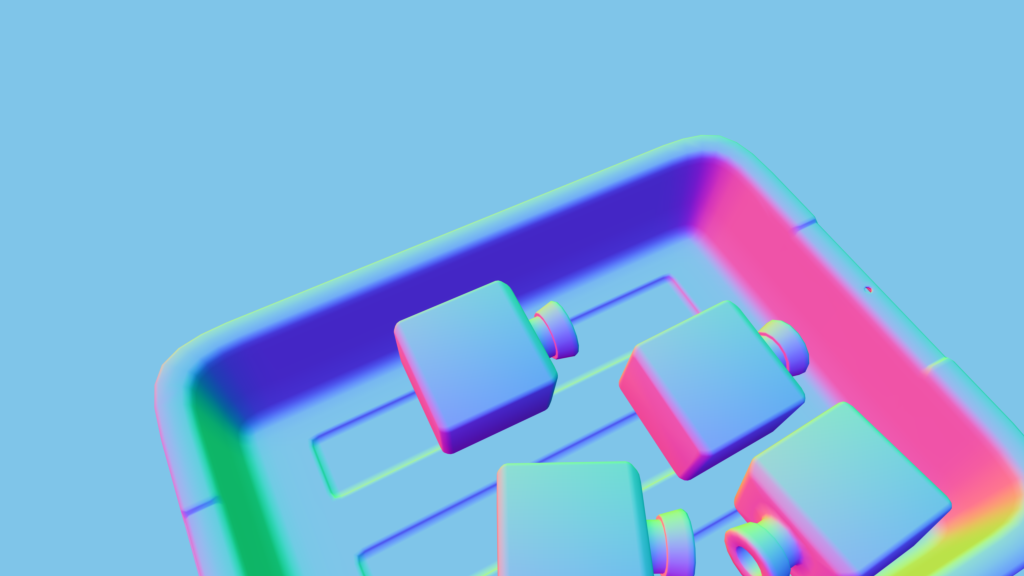} &
        \includegraphics[width=\imgw]{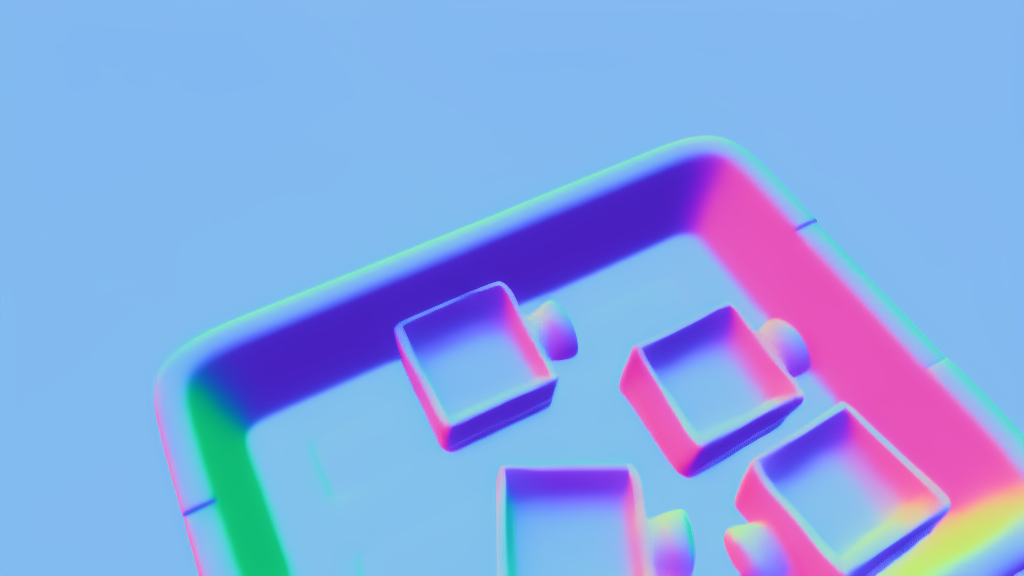} &
        \includegraphics[width=\imgw]{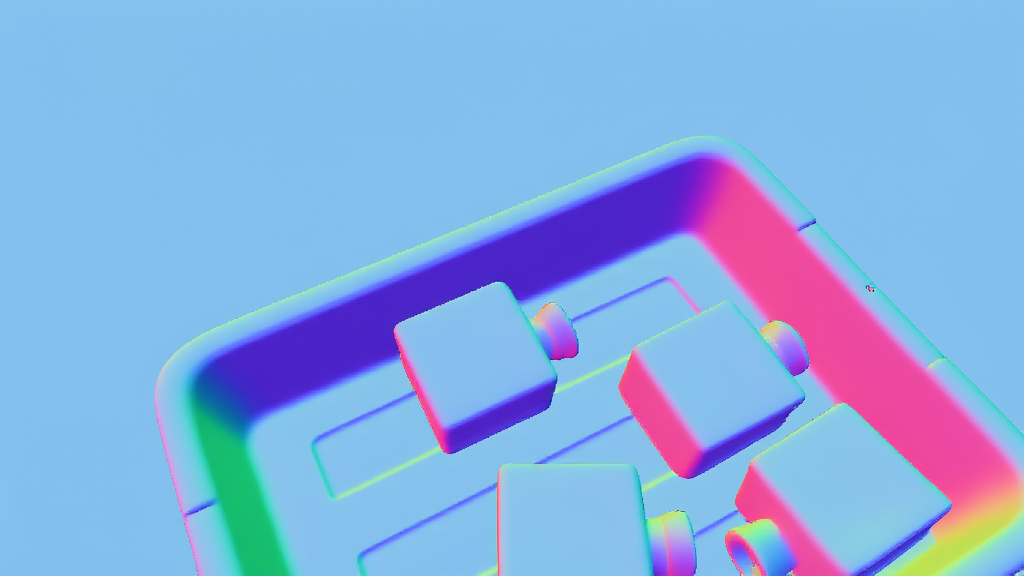} &
        \includegraphics[width=\imgw]{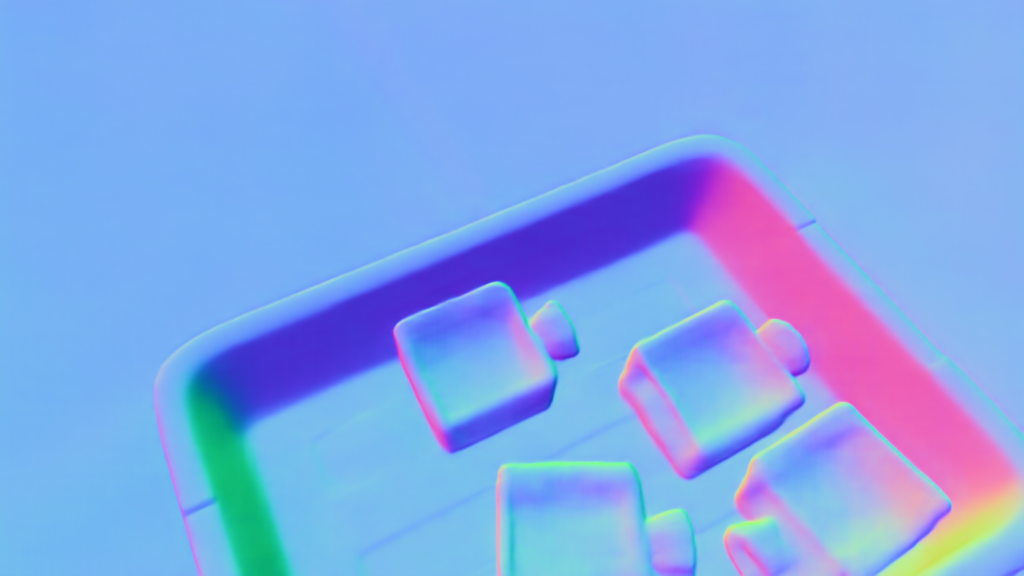} &
        \includegraphics[width=\imgw]{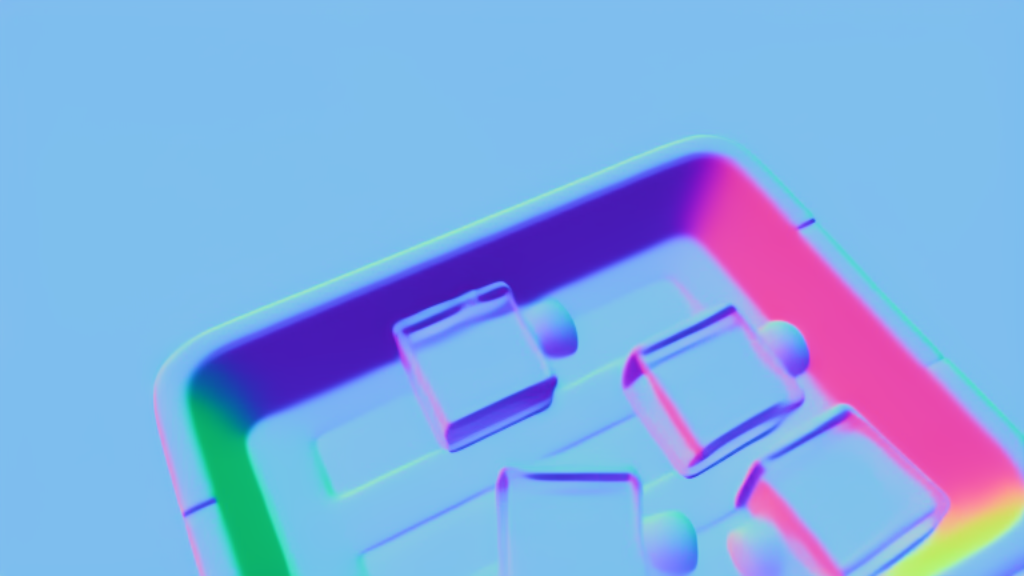} &
        \includegraphics[width=\imgw]{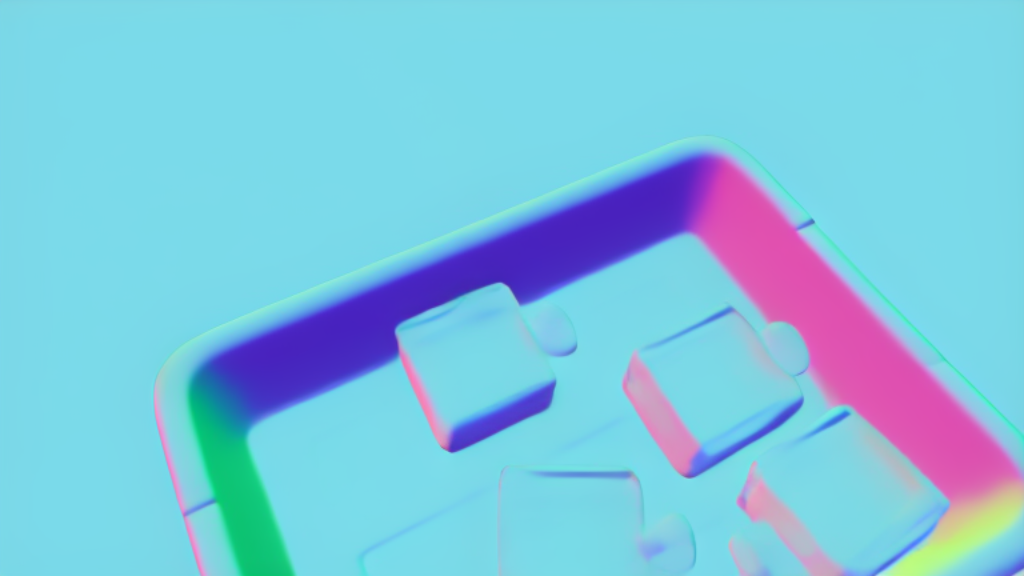} &
        \includegraphics[width=\imgw]{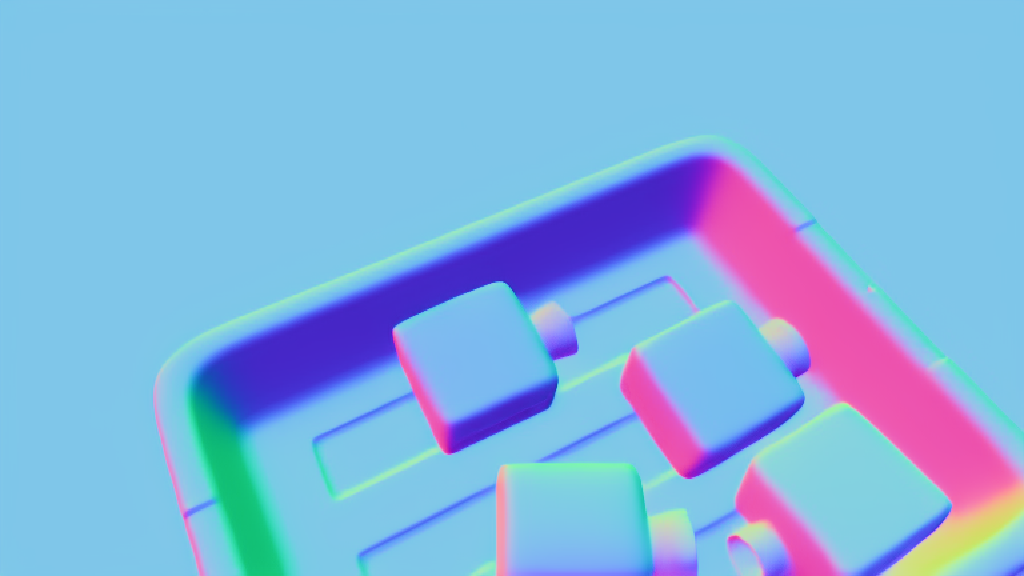} &
        \includegraphics[width=\imgw]{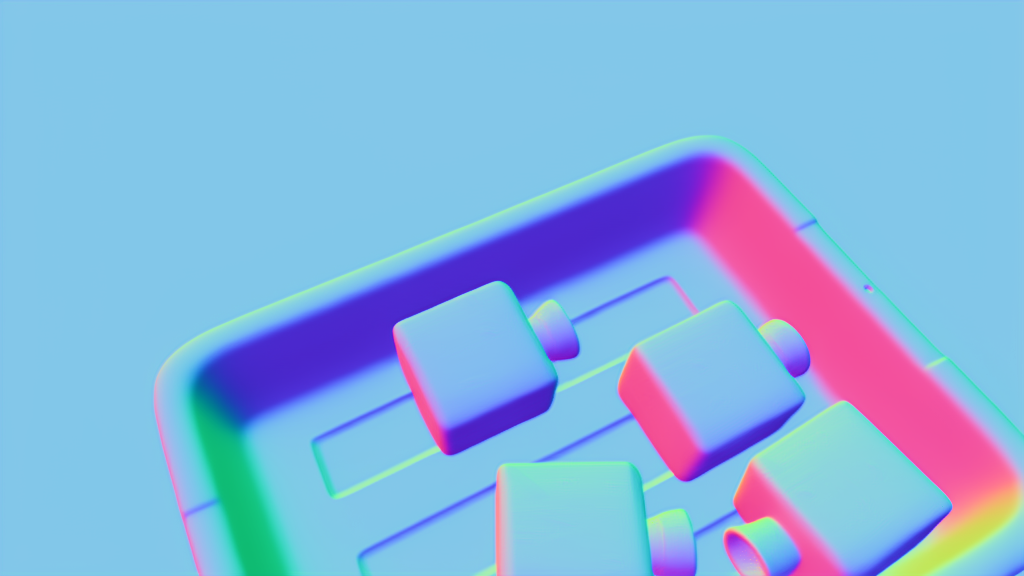} \\
        &
        \includegraphics[width=\imgw]{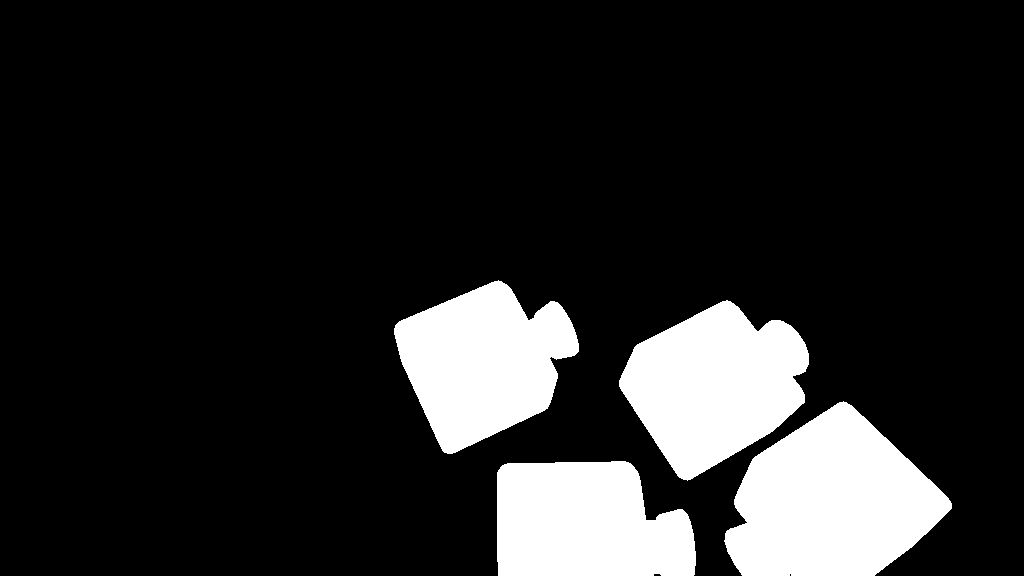} &
        \includegraphics[width=\imgw]{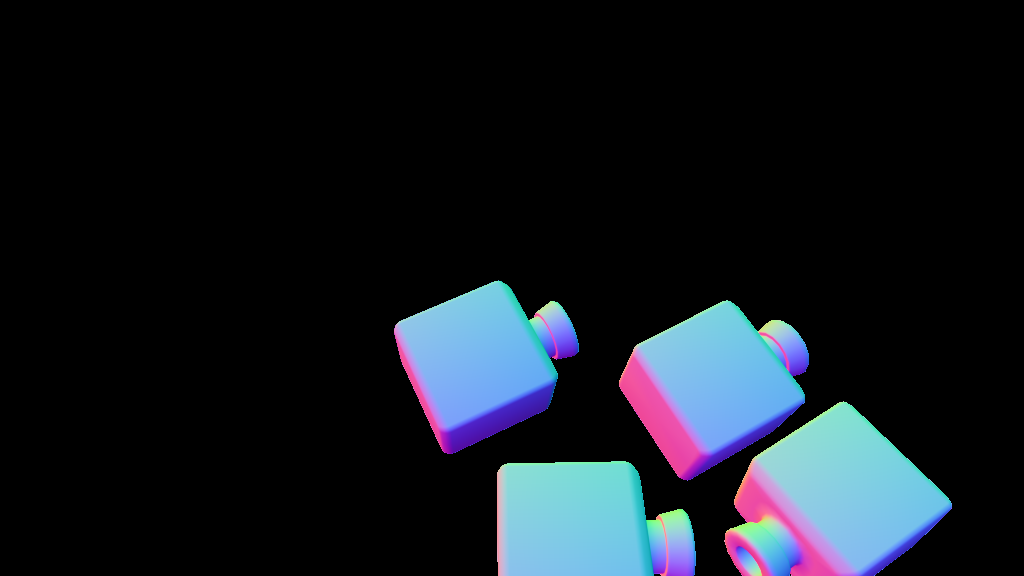} &
        \includegraphics[width=\imgw]{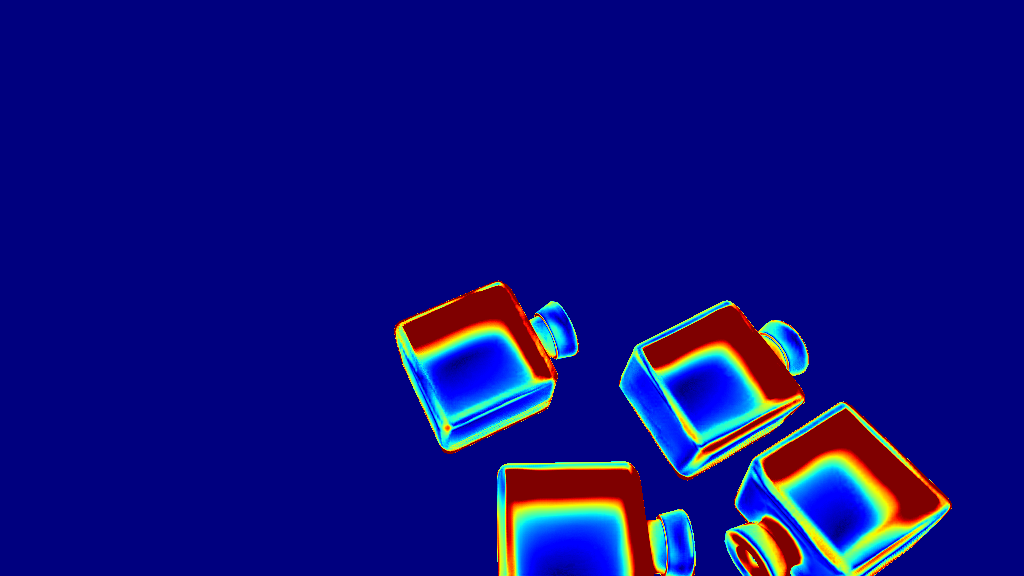} &
        \includegraphics[width=\imgw]{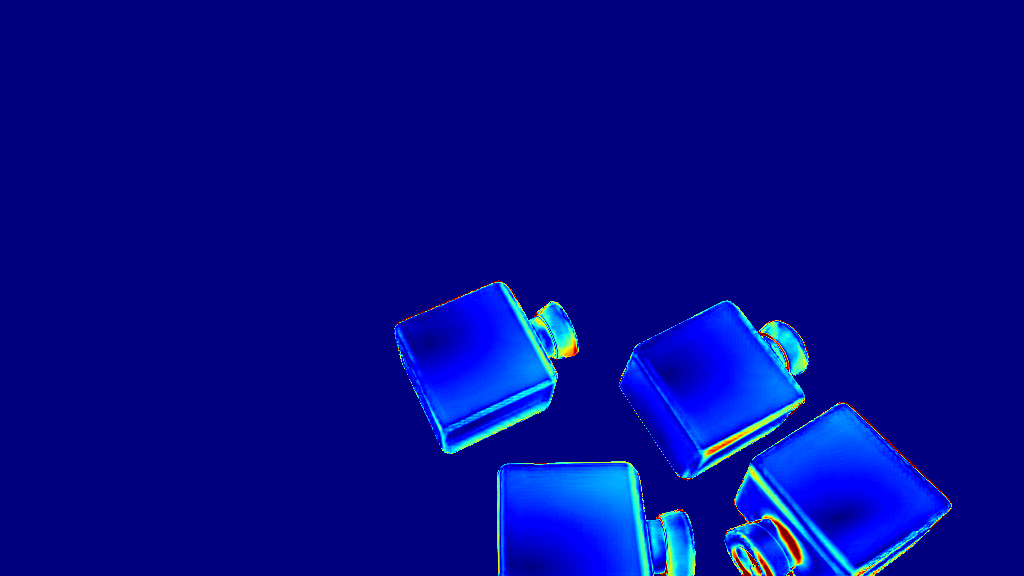} &
        \includegraphics[width=\imgw]{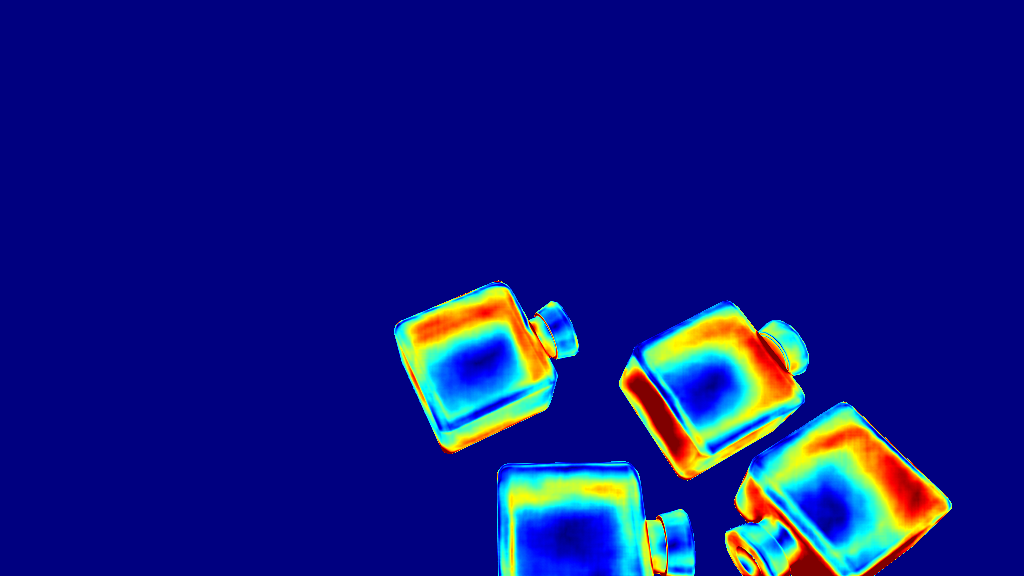} &
        \includegraphics[width=\imgw]{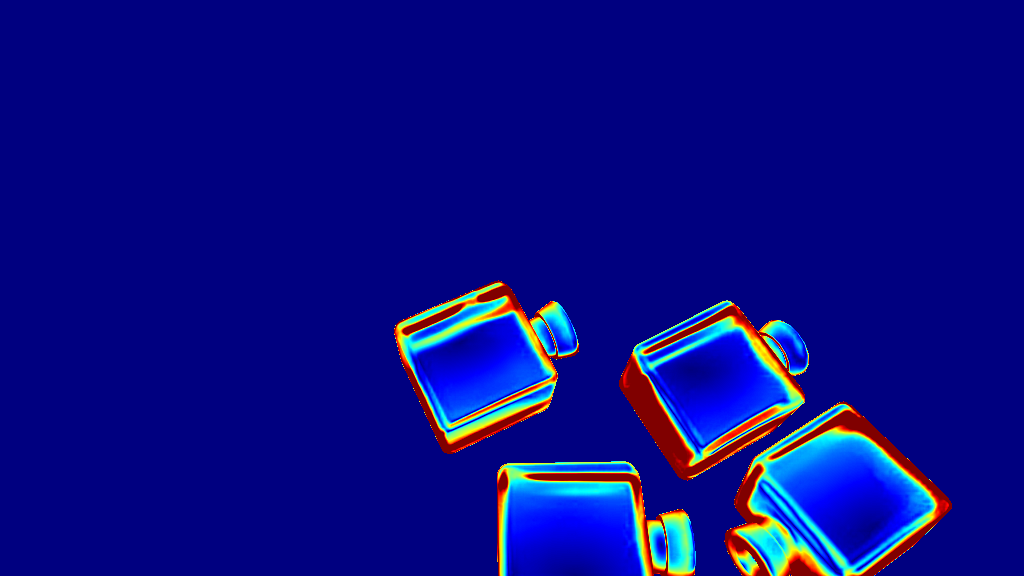} &
        \includegraphics[width=\imgw]{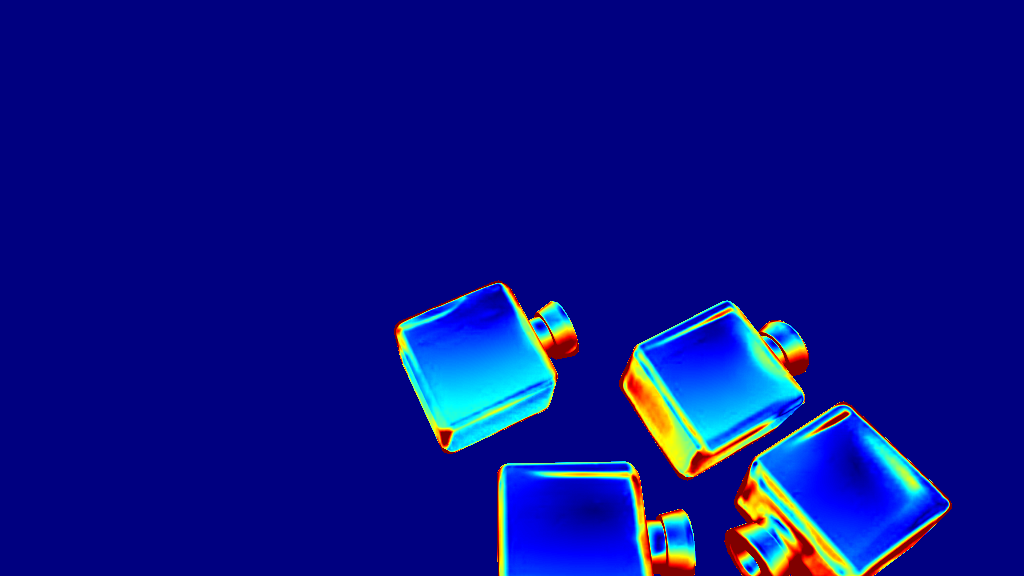} &
        \includegraphics[width=\imgw]{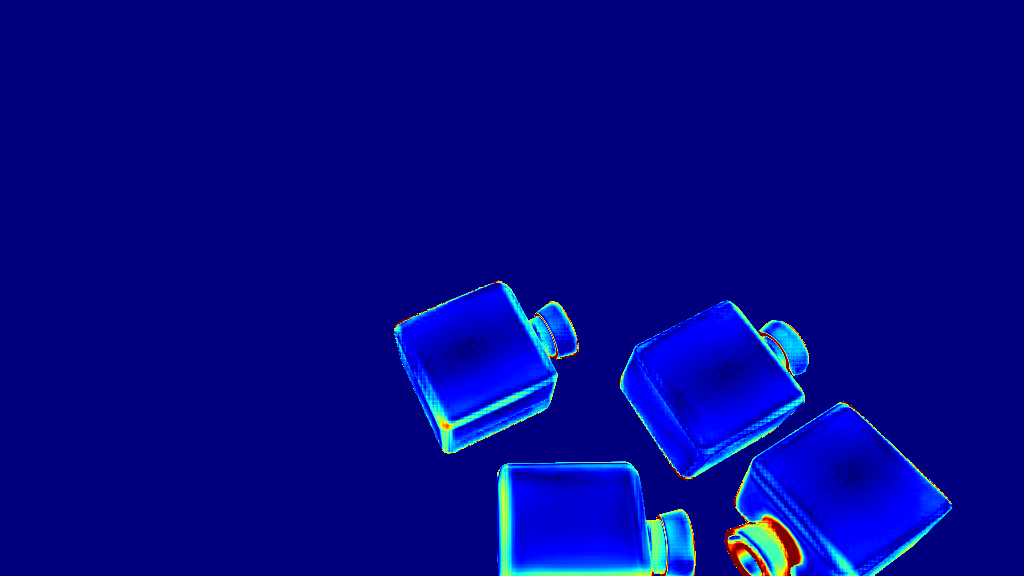} &
        \includegraphics[width=\imgw]{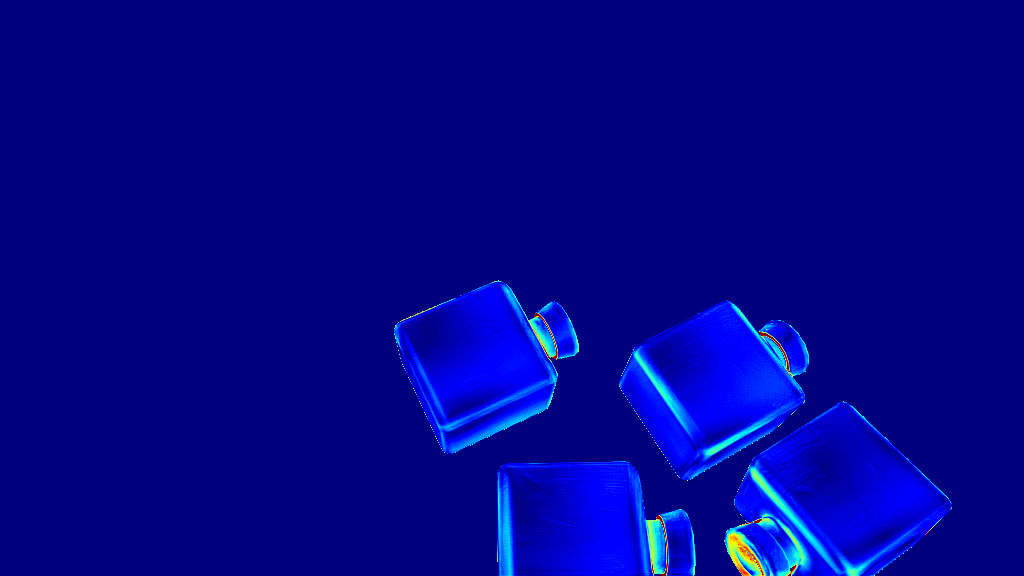} \\[2pt]
        \multirow{2}{*}[2.5ex]{\rotatebox{90}{\small ClearPose}} &
        \includegraphics[width=\imgw]{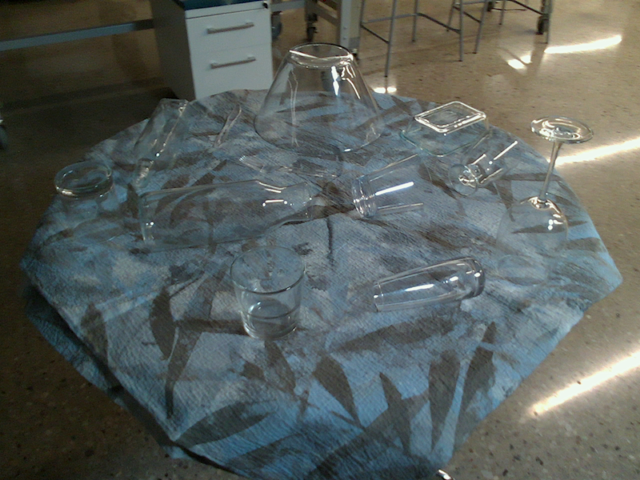} &
        \includegraphics[width=\imgw]{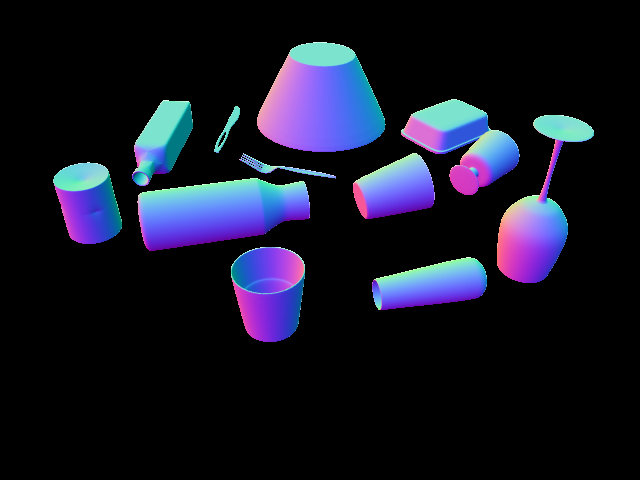} &
        \includegraphics[width=\imgw]{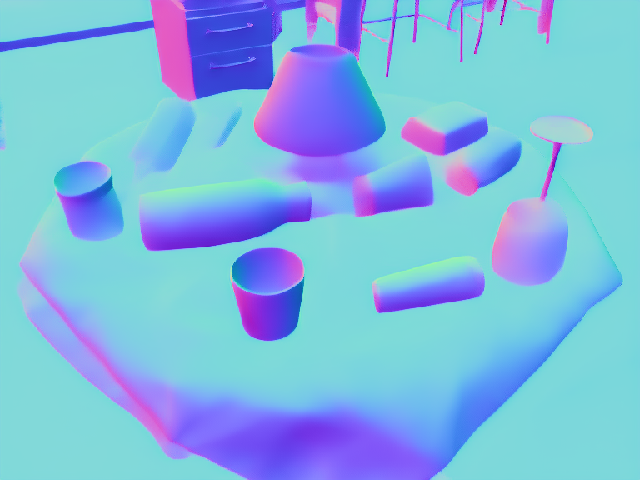} &
        \includegraphics[width=\imgw]{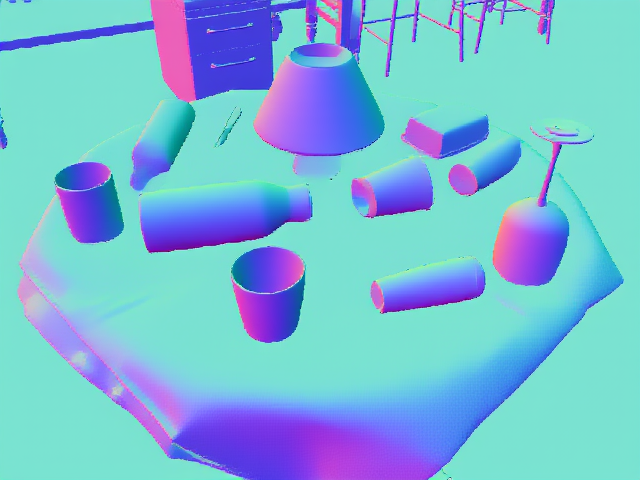} &
        \includegraphics[width=\imgw]{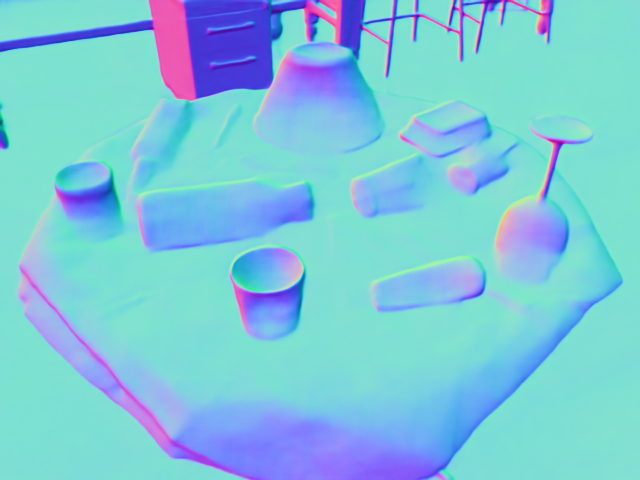} &
        \includegraphics[width=\imgw]{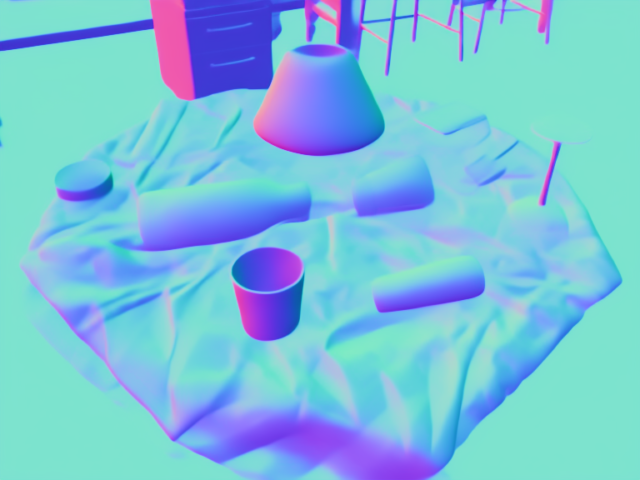} &
        \includegraphics[width=\imgw]{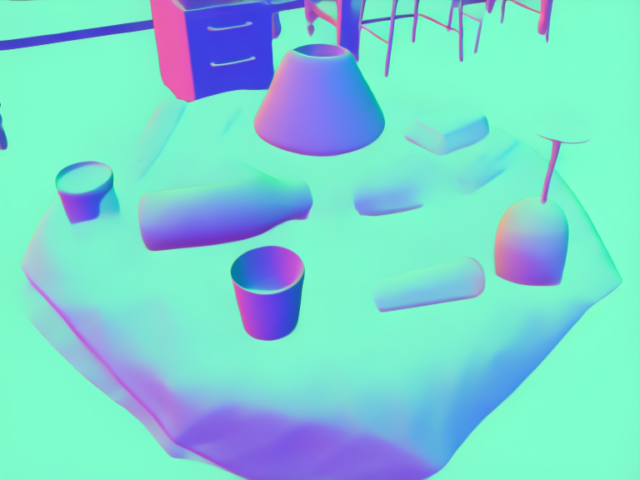} &
        \includegraphics[width=\imgw]{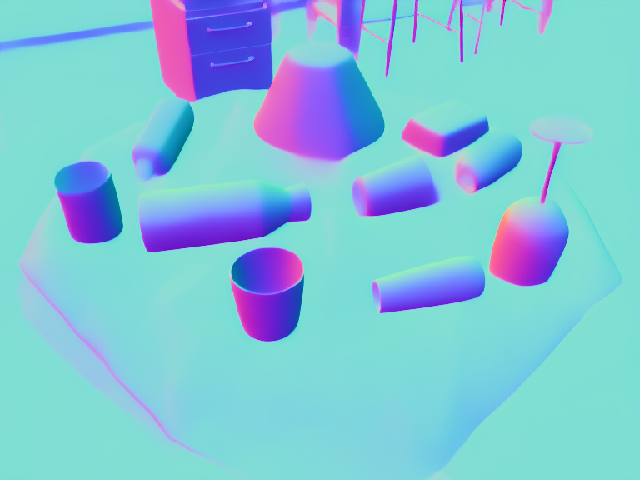} &
        \includegraphics[width=\imgw]{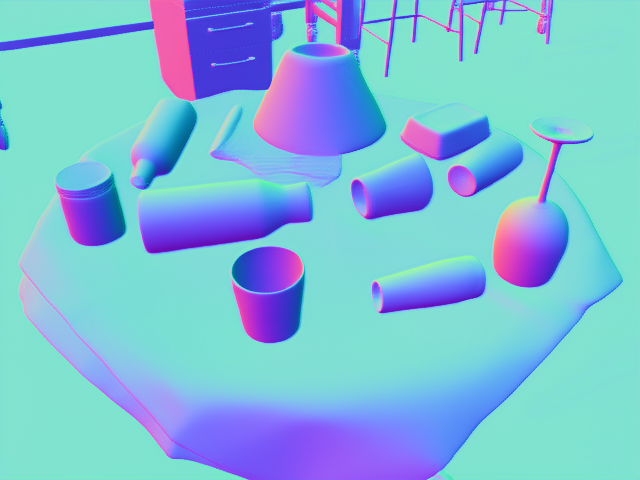} \\
        &
        \includegraphics[width=\imgw]{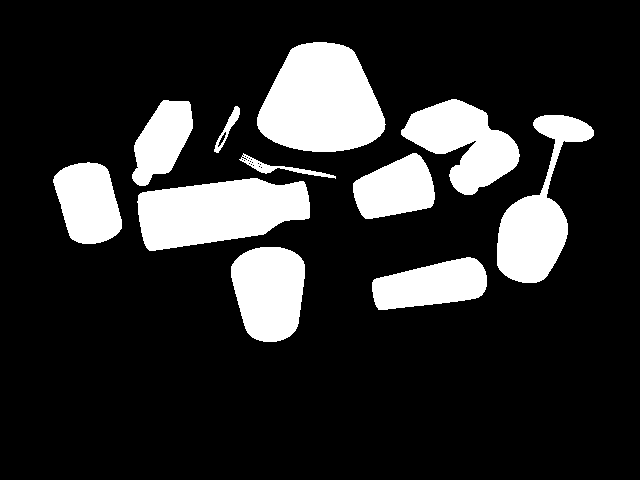} &
        \includegraphics[width=\imgw]{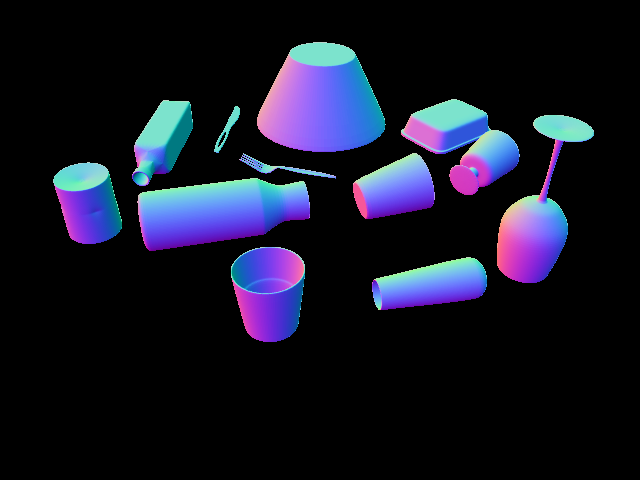} &
        \includegraphics[width=\imgw]{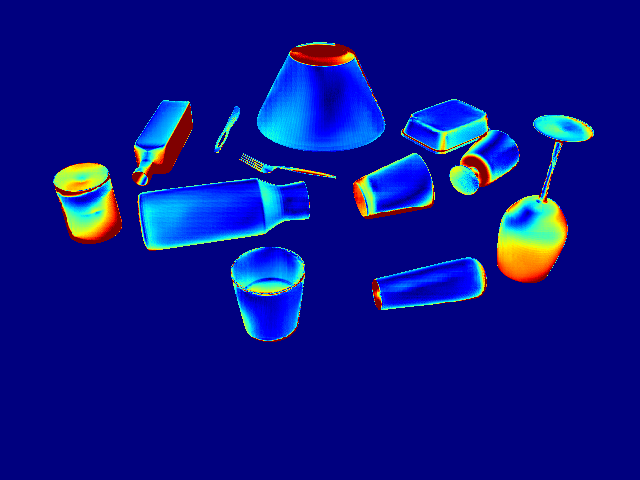} &
        \includegraphics[width=\imgw]{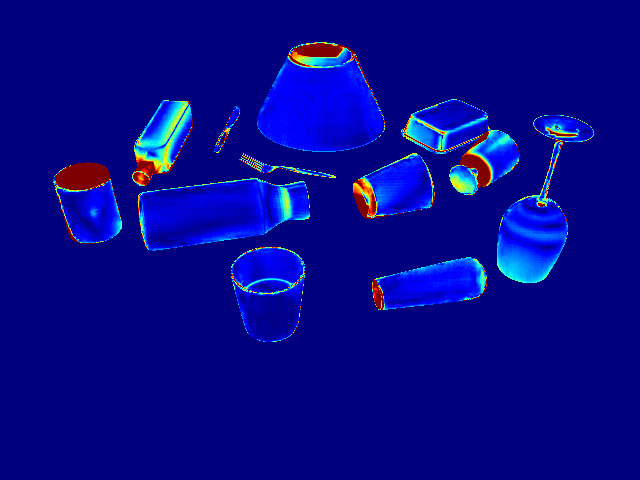} &
        \includegraphics[width=\imgw]{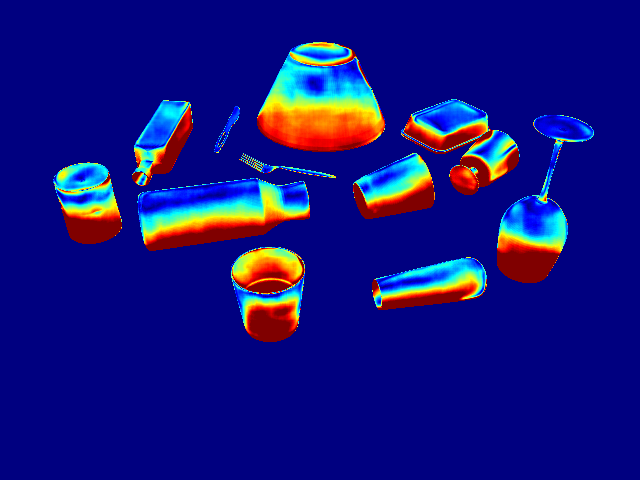} &
        \includegraphics[width=\imgw]{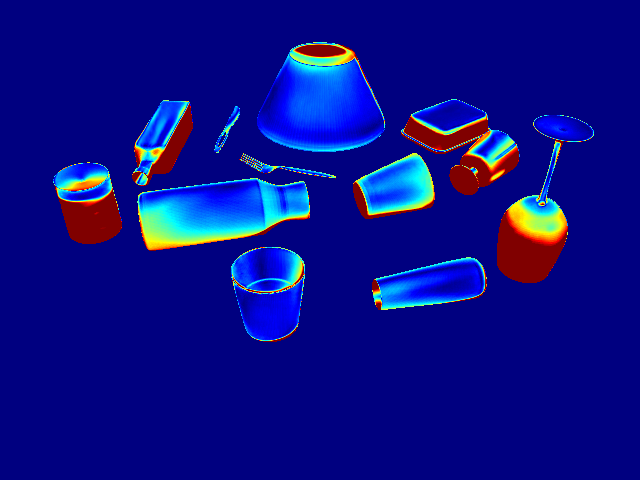} &
        \includegraphics[width=\imgw]{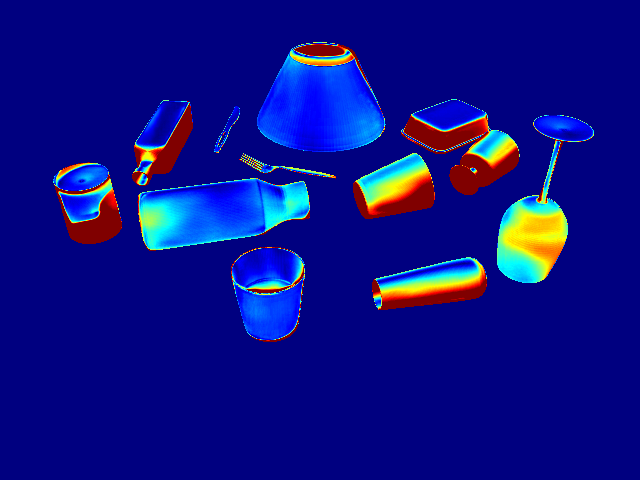} &
        \includegraphics[width=\imgw]{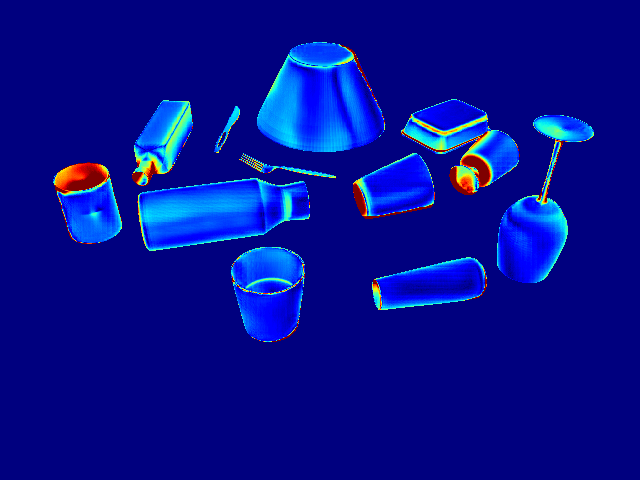} &
        \includegraphics[width=\imgw]{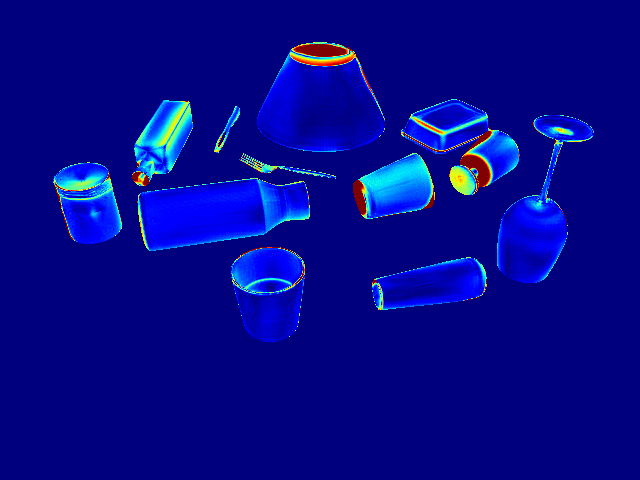} \\
    \end{tabular}
    \caption{\textbf{Qualitative comparison on transparent object normal estimation.}
    We compare our method against state-of-the-art approaches across TN-Syn, ClearGrasp, and ClearPose datasets.
    For each dataset, the top row shows predicted normals and the bottom row shows angular error maps (blue: low, red: high).
    Even on ClearPose, an extremely challenging real-world dataset with diverse transparent objects under cluttered scenes, our method produces lower-error zero-shot predictions than the compared approaches.
    Existing methods often produce blurry or incorrect normals on transparent regions due to refraction, while our method preserves more coherent surface geometry. Please zoom in \faSearchPlus~for details. (\S~\ref{par:qual_comparison})}
    \label{fig:comparison}
\end{figure*}

\Figref{fig:comparison} further evaluates three transparent object benchmarks. For each dataset, the first row shows predicted normal maps, and the second row displays error maps within the transparent object mask (blue: low error, red: high error). Existing methods often produce distorted normal predictions in transparent regions, as they are misled by refracted background textures. In contrast, TransNormal-2 produces lower-error geometry under challenging refractive conditions, benefiting from geometry-aware training losses and post-decode correction. Additional qualitative results are provided in the Supplementary Material.

\subsection{Quantitative Results}
\label{ssec:transparent_exp}

\noindent\textbf{Metrics.}
\label{sec:metrics}
Following prior works~\cite{bae2024dsine,ye2024stablenormal,he2024lotus}, we measure the \emph{mean angular error} (MAE; reported as Mean$\downarrow$ in the tables) and the percentage of pixels within $11.25^\circ$ and $30^\circ$ thresholds ($\uparrow$).
The Avg.~Rank is computed by ranking each method on every metric across all evaluation datasets within each table (four for general scenes, three for transparent objects), with tied values sharing their average rank, then averaging per-metric ranks.
For Edit2Perceive, we exclude Sintel from the average because it is trained on Sintel.

\noindent\textbf{General Scene Normal Estimation.}\label{ssec:general_normal} As shown in \tabref{tab:general}, TransNormal-2 matches or exceeds MoGe-2, a discriminative transformer trained with about $73{\times}$ as much labeled data, on all eight general-scene metrics.  It matches MoGe-2 on NYUv2 and iBims mean angular error ($14.7^\circ$), improves ScanNet ($12.7^\circ$ vs.\ $12.8^\circ$) and Sintel ($29.2^\circ$ vs.\ $29.3^\circ$), and improves the $11.25^\circ$ accuracy on all four general-scene benchmarks.  Overall, TransNormal-2 attains Avg.\ Rank 1.4, ahead of MoGe-2 (2.3); a simple horizontal-flip TTA ($2{\times}$ inference) further improves most benchmarks (Supplementary Material).

\noindent\textbf{Transparent Object Normal Estimation.}\label{ssec:transparent_normal} \tabref{tab:transparent} shows that TransNormal-2 leads all metrics on CG-Syn, TN-Syn, and zero-shot ClearPose.  It reduces the strongest prior mean angular error by $4.2^\circ$ on ClearGrasp ($11.3^\circ$ vs.\ Lotus-2's $15.5^\circ$) and by $3.1^\circ$ on ClearPose ($19.1^\circ$ vs.\ FE2E's $22.2^\circ$), while TN-Syn is already near saturation ($3.6^\circ$, $99.1\%$ at $30^\circ$).

\noindent\textbf{Comparison with TransNormal.}\label{ssec:comparison_transnormal} TransNormal-2 is a substantial redesign of the original TransNormal~\cite{li2026transnormal}.  As summarized in \tabref{tab:transnormal_comparison}, it replaces the SD~2.0 U-Net full fine-tuning pipeline with FLUX.2 LoRA adaptation, adds vMF and inverse-rendering losses, and uses the GRM to reduce VAE-decoded boundary errors, improving every head-to-head benchmark with the largest gain on zero-shot ClearPose ($-6.4^\circ$).

\begin{table}[!t]
\caption{\textbf{TransNormal vs.\ TransNormal-2: Key differences and head-to-head mean angular error ($^\circ$, $\downarrow$).}  TransNormal denotes our prior model~\cite{li2026transnormal}; both methods are evaluated under identical protocols. (\S~\ref{ssec:comparison_transnormal})}
\label{tab:transnormal_comparison}
\centering
\footnotesize
\setlength{\tabcolsep}{2pt}
\begin{tabular}{@{}>{\raggedright\arraybackslash}p{0.24\columnwidth}>{\raggedright\arraybackslash}p{0.32\columnwidth}>{\raggedright\arraybackslash}p{0.36\columnwidth}@{}}
\toprule
\textbf{Component} & \textbf{TransNormal} & \textbf{TransNormal-2} \\
\midrule
Backbone & SD~2.0 (U-Net, 865M) & FLUX.2[klein] (DiT, 9B) \\
Adaptation & Full fine-tune & LoRA ($r{=}256$) \\
Inference & Single-step prediction & Single-step prediction \\
Training loss & MSE + Wavelet & MSE + Wavelet + vMF + Rendering \\
Decode fix & None & GRM ($\sim$0.4M params) \\
Training data & CG + TN-Syn + HS + VK & CG + TN-Syn + HS + VK \\
\midrule
\multicolumn{3}{@{}l}{\textbf{Mean Angular Error ($^\circ$, $\downarrow$)}} \\
\midrule
CG-Syn    & 16.1 & \textbf{11.3} ($-4.8$) \\
TN-Syn    & 3.9  & \textbf{3.6}  ($-0.3$) \\
ClearPose & 25.5 & \textbf{19.1} ($-6.4$) \\
NYUv2     & 16.6 & \textbf{14.7} ($-1.9$) \\
ScanNet   & 15.3 & \textbf{12.7} ($-2.6$) \\
\bottomrule
\end{tabular}
\end{table}

\subsection{Training and Refinement Recipe Ablation}\label{par:ablation_studies}

\begin{table}[!t]
\scriptsize
\caption{\textbf{Ablation on key design choices.} Group~A: cumulative training-recipe ablation; rows~(a)--(b) are 20K-step from-scratch controls, and row~(c) uses the full staged Phase-1 schedule. Group~B: decode refinement. Row~(f) is full TransNormal-2. (\S~\ref{par:ablation_studies})}
\label{tab:ablation_main}
\centering
\setlength{\tabcolsep}{2.5pt}
\begin{tabular}{cl|cc|cc}
\toprule
& \multirow{2}{*}{Configuration}
& \multicolumn{2}{c|}{NYUv2}
& \multicolumn{2}{c}{ScanNet} \\
& & Mean$\downarrow$ & $11.25^\circ\!\uparrow$ & Mean$\downarrow$ & $11.25^\circ\!\uparrow$ \\
\midrule
\multicolumn{6}{l}{\textit{Group A: Training recipe (cumulative)}} \\
\midrule
(a) & MSE only & 16.9 & 58.8 & 14.6 & 64.9 \\
(b) & \quad + Wavelet + vMF & 16.6 & 58.5 & 14.3 & 65.8 \\
(c) & \quad + Rendering loss & 16.0 & 61.1 & 13.7 & 68.2 \\
\midrule
\multicolumn{6}{l}{\textit{Group B: Decode refinement (from row c)}} \\
\midrule
(d) & + Decoder LoRA & \cellcolor{best2}15.9 & \cellcolor{best2}61.5 & \cellcolor{best2}13.6 & 68.3 \\
(e) & + GRM (joint train) & 16.0 & \cellcolor{best2}61.5 & \cellcolor{best2}13.6 & \cellcolor{best2}68.8 \\
(f) & + GRM (decoupled) & \cellcolor{best}\textbf{14.7} & \cellcolor{best}\textbf{62.5} & \cellcolor{best}\textbf{12.7} & \cellcolor{best}\textbf{69.2} \\
\bottomrule
\end{tabular}
\end{table}

\tabref{tab:ablation_main} presents the main ablation results on NYUv2 and ScanNet, where edge-aligned indoor geometry is most sensitive to the VAE reconstruction bottleneck.

\noindent\textbf{Training recipe (Group A).} Starting from the MSE-only baseline (a), adding wavelet and vMF losses (b) improves mean angular error by $0.3^\circ$ on both NYUv2 and ScanNet, confirming that pixel-space angular supervision complements latent-space MSE. Completing the staged Phase~1 recipe, which extends the schedule and enables the inverse rendering self-consistency loss (c), yields a further $0.6^\circ$ gain on NYUv2 and $0.6^\circ$ on ScanNet, consistent with the rendering term acting as an additional diffuse-scene geometry cue.

\noindent\textbf{Decode refinement (Group B).} Decoder LoRA (d) improves over row~(c), but it remains weaker than the decoupled GRM on both NYUv2 and ScanNet. Joint training of the GRM with the DiT (e) underperforms the decoupled GRM fitted to the frozen core predictor, suggesting that gradient leakage from the refinement module destabilizes the converged transformer. In contrast, the final decoupled GRM (f) achieves the best results on NYUv2 ($14.7^\circ$, $62.5\%$) and ScanNet ($12.7^\circ$, $69.2\%$).

\begin{figure*}[!t]
    \centering
    \setlength{\tabcolsep}{1pt}
    \begin{tabular}{@{}ccccc@{}}
        \includegraphics[width=0.195\textwidth]{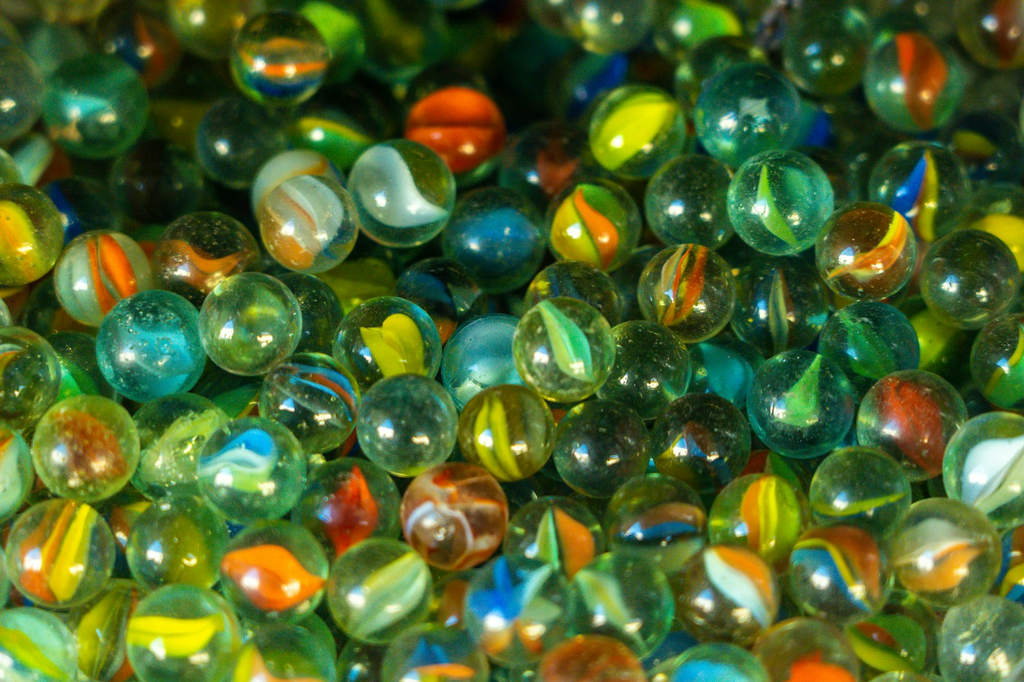} &
        \includegraphics[width=0.195\textwidth]{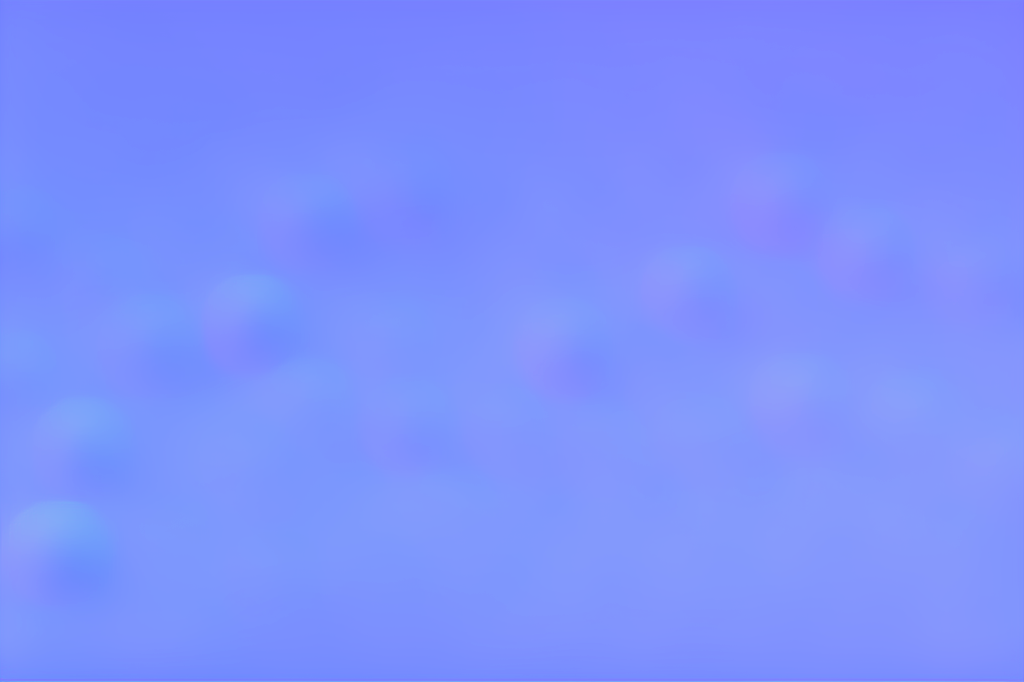} &
        \includegraphics[width=0.195\textwidth]{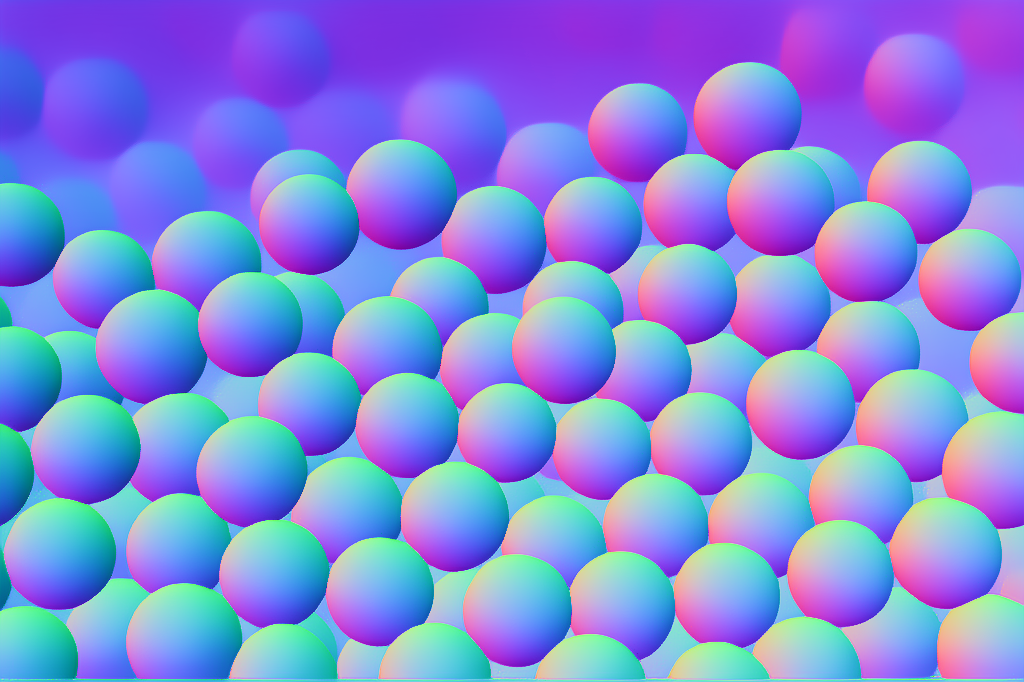} &
        \includegraphics[width=0.195\textwidth]{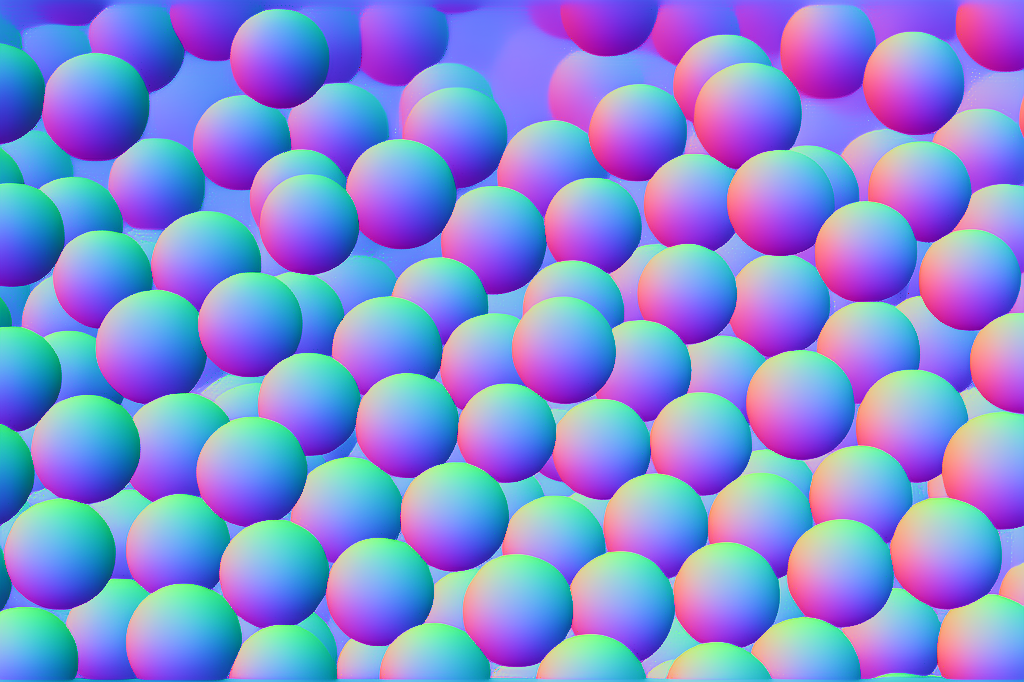} &
        \includegraphics[width=0.195\textwidth]{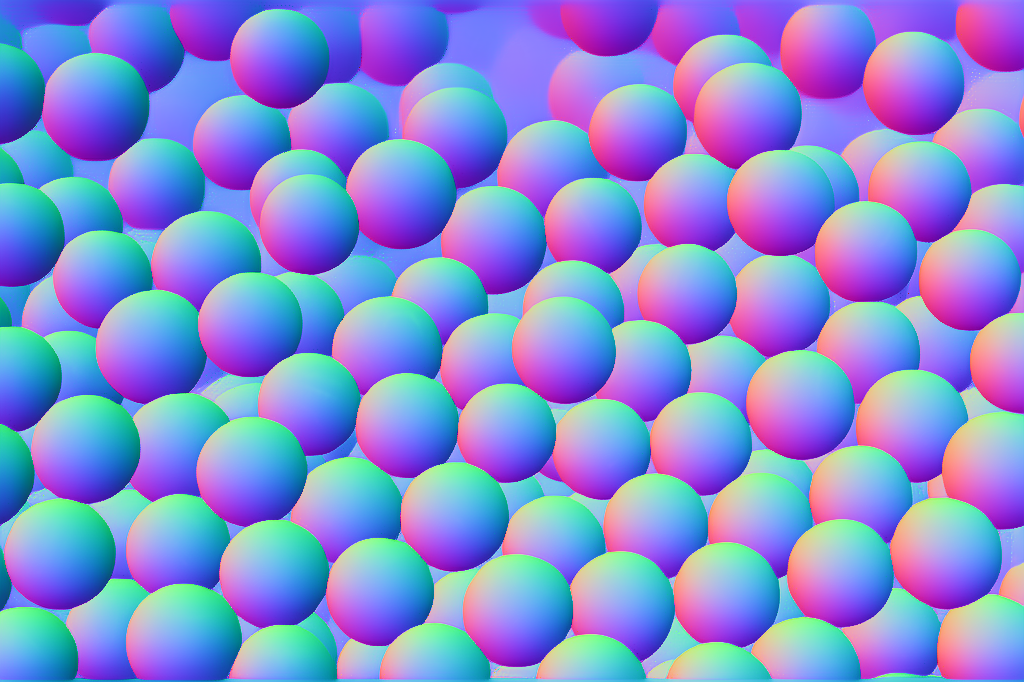} \\[-2pt]
        \footnotesize (a) Input &
        \footnotesize (b) MSE only &
        \footnotesize (c) +W+vMF &
        \footnotesize (d) +Rendering &
        \footnotesize (e) Full + GRM \\[-2pt]
    \end{tabular}
    \vspace{-4pt}
    \caption{\textbf{Qualitative ablation on in-the-wild objects.}
    Starting from the MSE-only baseline, wavelet and vMF supervision recover more coherent local geometry, inverse-rendering self-consistency further improves structure, and the full model with the GRM reduces residual boundary artifacts.
    Panel letters (b)--(e) correspond to rows (a), (b), (c), and (f) of \tabref{tab:ablation_main}, respectively.
    (\S~\ref{par:ablation_studies})}
    \label{fig:ablation}
\end{figure*}

\subsection{GRM Design Ablation}
\label{ssec:ablation_grm_arch}

The GRM uses $K$ residual blocks with edge feature injection at each stage.  We ablate its depth ($K \in \{2, 4, 8\}$), width (hidden channels $\in \{64, 128\}$), and input design (with vs.\ without RGB edge guidance) while keeping the final core predictor, GRM training losses, two-stage GRM training protocol, and inference settings fixed.  \Tabref{tab:ablation_grm_arch} shows that the GRM design is stable across depth and width: among the RGB-guided variants, average mean angular error spans only $0.03^\circ$.  Increasing capacity to $K{=}8$ or hidden$=$128 does not provide a systematic gain, while the compact $K{=}2$ variant nearly matches the default.  We therefore use $K{=}4$, hidden$=$64 as a conservative operating point that balances accuracy, parameter count, and stability rather than as a per-dataset optimum.

\noindent\textbf{RGB guidance mainly protects general-scene robustness.}  Removing RGB guidance altogether (the edge encoder, the RGB input channels, and the RGB-guided anchor, so that every image uses the coarse normal as anchor) increases mean error on the two large indoor benchmarks (about $+1.0^\circ$ on NYUv2 and $+0.7^\circ$ on ScanNet), but remains competitive on synthetic and transparent benchmarks, matching or slightly improving CG-Syn, TN-Syn, and ClearPose.  This indicates that RGB guidance is most important as a cross-domain image-edge prior for general scenes, while geometry-only refinement can still reduce residual errors in synthetic and transparent settings.

\begin{table*}[!t]
\centering
\caption{\textbf{GRM design ablation.}  We ablate the depth ($K$), width (hidden channels), and input design of the Geometric Refinement Module while keeping the final core predictor, GRM training losses, two-stage GRM training protocol, and inference settings fixed.  Metrics: mean angular error (Mean$\downarrow$) and percentage of pixels within $11.25^\circ$ ($\uparrow$).  The RGB-guided depth and width variants are highly stable, with average mean angular error varying by only $0.03^\circ$ across $K{=}2$, $K{=}4$, $K{=}8$, and hidden$=$128.  We use $K{=}4$, hidden$=$64 as a conservative accuracy-parameter-stability tradeoff.  Removing RGB guidance mainly hurts NYUv2 and ScanNet, while remaining competitive on synthetic and transparent benchmarks. (\S~\ref{ssec:ablation_grm_arch})}
\label{tab:ablation_grm_arch}
\setlength{\tabcolsep}{3pt}
\resizebox{\textwidth}{!}{%
\begin{tabular}{l|r|cc|cc|cc|cc|cc|cc|cc}
\toprule
\multirow{2}{*}{Configuration}
& \multirow{2}{*}{\#Params}
& \multicolumn{2}{c|}{NYUv2}
& \multicolumn{2}{c|}{ScanNet}
& \multicolumn{2}{c|}{iBims}
& \multicolumn{2}{c|}{Sintel}
& \multicolumn{2}{c|}{CG-Syn}
& \multicolumn{2}{c|}{TN-Syn}
& \multicolumn{2}{c}{ClearPose} \\
& & Mean$\downarrow$ & $11.25^\circ\!\uparrow$ & Mean$\downarrow$ & $11.25^\circ\!\uparrow$ & Mean$\downarrow$ & $11.25^\circ\!\uparrow$ & Mean$\downarrow$ & $11.25^\circ\!\uparrow$ & Mean$\downarrow$ & $11.25^\circ\!\uparrow$ & Mean$\downarrow$ & $11.25^\circ\!\uparrow$ & Mean$\downarrow$ & $11.25^\circ\!\uparrow$ \\
\midrule
$K{=}2$, hidden$=$64 & 234K
& \cellcolor{best}\textbf{14.7} & \cellcolor{best}\textbf{62.5} & \cellcolor{best}\textbf{12.7} & \cellcolor{best}\textbf{69.2} & 14.8 & 70.6 & 29.2 & 27.6
& \cellcolor{best}\textbf{11.2} & \cellcolor{best2}65.1 & \cellcolor{best2}3.6 & \cellcolor{best}\textbf{95.8} & \cellcolor{best}\textbf{19.1} & \cellcolor{best}\textbf{52.3} \\
$K{=}8$, hidden$=$64 & 716K
& \cellcolor{best}\textbf{14.7} & \cellcolor{best}\textbf{62.5} & \cellcolor{best}\textbf{12.7} & \cellcolor{best2}69.1 & 14.8 & 70.5 & 29.2 & 27.4
& \cellcolor{best2}11.3 & 65.0 & \cellcolor{best2}3.6 & \cellcolor{best}\textbf{95.8} & \cellcolor{best}\textbf{19.1} & \cellcolor{best}\textbf{52.3} \\
$K{=}4$, hidden$=$128 & 1.47M
& \cellcolor{best2}14.8 & \cellcolor{best2}62.3 & \cellcolor{best2}12.8 & 68.9 & \cellcolor{best2}14.7 & \cellcolor{best2}70.9 & \cellcolor{best2}29.1 & \cellcolor{best2}27.7
& \cellcolor{best2}11.3 & 64.9 & \cellcolor{best}\textbf{3.5} & \cellcolor{best}\textbf{95.8} & \cellcolor{best}\textbf{19.1} & \cellcolor{best2}52.2 \\
$K{=}4$, hidden$=$64, w/o RGB & 356K
& 15.7 & 61.5 & 13.4 & 68.6 & \cellcolor{best}\textbf{14.6} & \cellcolor{best}\textbf{72.1} & \cellcolor{best}\textbf{28.8} & \cellcolor{best}\textbf{28.8}
& \cellcolor{best}\textbf{11.2} & \cellcolor{best}\textbf{65.2} & \cellcolor{best}\textbf{3.5} & \cellcolor{best}\textbf{95.8} & \cellcolor{best}\textbf{19.1} & \cellcolor{best}\textbf{52.3} \\
\midrule
$K{=}4$, hidden$=$64 \textbf{(default)} & $\sim$0.4M
& \cellcolor{best}\textbf{14.7} & \cellcolor{best}\textbf{62.5} & \cellcolor{best}\textbf{12.7} & \cellcolor{best}\textbf{69.2} & \cellcolor{best2}14.7 & 70.6 & 29.2 & 27.5
& \cellcolor{best2}11.3 & 65.0 & \cellcolor{best2}3.6 & \cellcolor{best}\textbf{95.8} & \cellcolor{best}\textbf{19.1} & \cellcolor{best}\textbf{52.3} \\
\bottomrule
\end{tabular}%
}
\end{table*}

\subsection{LoRA Rank Ablation}
\label{ssec:ablation_lora_rank}

\begin{table}[t]
\centering
\caption{\textbf{LoRA rank ablation.}  We vary the LoRA rank $r$ (with $\alpha{=}r$) and evaluate the core predictor with GRM refinement disabled, isolating the capacity of the DiT LoRA adapter.  Rank 256 gives the best average error and matches or leads the best rounded value on every dataset, with no further gains at rank 512. (\S~\ref{ssec:ablation_lora_rank})}
\label{tab:ablation_lora_rank}
\scriptsize
\setlength{\tabcolsep}{1.5pt}
\begin{tabular}{@{}r|r|cccc|ccc|c@{}}
\toprule
\multirow{2}{*}{Rank}
& \multirow{2}{*}{\#Params}
& \multicolumn{4}{c|}{General}
& \multicolumn{3}{c|}{Transparent}
& \multirow{2}{*}{Avg$\downarrow$} \\
& & NYUv2 & ScanNet & iBims & Sintel & CG-Syn & TN-Syn & ClearPose & \\
\midrule
64  & $\sim$175M
& 16.6 & 14.6 & 15.9 & 30.6 & 12.0 & 4.6 & 21.1 & 16.5 \\
128 & $\sim$350M
& \textbf{16.0} & 14.1 & 15.7 & 30.6 & 12.3 & 5.4 & 21.0 & 16.5 \\
\rowcolor{best}\textbf{256} & $\sim$\textbf{700M}
& \textbf{16.0} & \textbf{13.7} & \textbf{14.8} & \textbf{28.9} & \textbf{11.4} & \textbf{3.9} & \textbf{19.3} & \textbf{15.4} \\
512 & $\sim$1.4B
& 16.8 & 14.6 & 16.0 & 31.2 & 12.6 & 4.0 & 20.6 & 16.5 \\
\bottomrule
\end{tabular}
\end{table}

We ablate the LoRA rank $r \in \{64, 128, 256, 512\}$ (with $\alpha{=}r$) by evaluating the core predictor with GRM refinement disabled, so the comparison isolates the DiT LoRA capacity.  \Tabref{tab:ablation_lora_rank} shows that rank 256 gives the best average error and matches or leads the best rounded value on every benchmark.  Both under-parameterized (rank 64, 128) and over-parameterized (rank 512) variants converge to similar average errors ($\sim$16.5$^\circ$), while rank 256 achieves 15.4$^\circ$, a clear sweet spot.  The rank 512 result is notable: doubling the trainable parameters from 700M to 1.4B provides \emph{no improvement}, suggesting that the FLUX.2 backbone's representational capacity is already well-captured at rank 256.

\subsection{GRM vs.\ Classical Image-Guided Filters}
\label{ssec:grm_vs_classical}

A natural alternative to the learned GRM is a classical image-guided filter.  We apply Guided Filter~\cite{he2013guided} with $r_{\text{GF}} \in \{4, 8, 16\}$ ($r_{\text{GF}}{=}4$ is the anchor setting of Eq.~\eqref{eq:grm_readout}), Joint Bilateral Filter, and Bilateral Filter to the same coarse normal maps and evaluate them with the same harness as the main tables.  \Tabref{tab:grm_vs_classical} shows that no filter improves all four benchmarks, whereas the GRM gives the lowest mean angular error on each of them and the best average ($10.6^\circ$ vs.\ $11.0^\circ$ for the best filter).

\begin{table}[t]
\centering
\scriptsize
\caption{%
  \textbf{Learned GRM vs.\ classical image-guided filters.}
  All rows refine the \emph{same} coarse normal predictions and are evaluated with the harness of the main tables, so ``No refinement'' matches the core-predictor results reported there.
  Mean angular error ($^\circ$, $\downarrow$); the GRM is best on every dataset and on average. (\S~\ref{ssec:grm_vs_classical})
}
\label{tab:grm_vs_classical}
\setlength{\tabcolsep}{4pt}
\begin{tabular}{l|cccc|c}
\toprule
\textbf{Method} & NYUv2 & ScanNet & CG-Syn & TN-Syn & Avg. \\
\midrule
No refinement            & 16.0 & 13.7 & 11.4 & 3.9 & 11.3 \\
Guided Filter ($r_{\text{GF}}{=}4$)  & 14.8 & 12.8 & 12.4 & 4.7 & 11.2 \\
Guided Filter ($r_{\text{GF}}{=}8$)  & 15.0 & 12.9 & 14.5 & 5.6 & 12.0 \\
Guided Filter ($r_{\text{GF}}{=}16$) & 16.3 & 13.8 & 18.8 & 7.1 & 14.0 \\
Joint Bilateral          & 15.0 & 13.0 & 11.7 & 4.2 & 11.0 \\
Bilateral                & 15.4 & 13.4 & 11.4 & 3.9 & 11.0 \\
\midrule
\textbf{GRM (Ours)}      & \cellcolor{best}\textbf{14.7} & \cellcolor{best}\textbf{12.7} & \cellcolor{best}\textbf{11.3} & \cellcolor{best}\textbf{3.6} & \cellcolor{best}\textbf{10.6} \\
\bottomrule
\end{tabular}
\end{table}

\subsection{Edge-Band Error Analysis}
\label{ssec:edge_analysis}

\begin{table}[t]
\centering
\scriptsize
\caption{\textbf{Edge vs.\ non-edge MAE analysis.}  Pixels are separated into edge (Canny on RGB, dilated $3{\times}3$) and non-edge regions.  We compare the final core predictor with and without the GRM on general-scene benchmarks, where RGB edges provide a reliable proxy for geometric discontinuities.  Negative $\Delta$ means lower MAE after GRM.  (\S~\ref{ssec:edge_analysis})}
\label{tab:edge_analysis}
\setlength{\tabcolsep}{3pt}
\begin{tabular}{l|cc|cc|cc}
\toprule
& \multicolumn{2}{c|}{Edge ($^\circ$)} & \multicolumn{2}{c|}{Non-Edge ($^\circ$)} & \multicolumn{2}{c}{$\Delta$ ($^\circ$)} \\
Dataset & w/o & w/ & w/o & w/ & Edge & Non-E \\
\midrule
NYUv2      & 26.1  & 23.5  & 13.7  & 12.9  & $-2.6$  & $-0.8$  \\
ScanNet    & 26.0  & 23.4  & 12.5  & 11.9  & $-2.6$  & $-0.6$  \\
\bottomrule
\end{tabular}
\end{table}

To quantify that the GRM specifically targets the VAE's spatially non-uniform degradation, we separate pixels into edge and non-edge regions (via Canny on the input RGB, dilated by $3{\times}3$).  \Tabref{tab:edge_analysis} reports this diagnostic on general-scene benchmarks, where RGB edges are a reliable proxy for geometric discontinuities.  Before GRM, edge MAE is $1.9$--$2.1{\times}$ the non-edge MAE.  After GRM refinement, edge pixels improve by about $2.6^\circ$, whereas non-edge pixels improve by $0.6^\circ$--$0.8^\circ$.  Thus the absolute gain on edge regions is $3.3$--$4.4{\times}$ that on non-edge regions, matching the VAE's spatial error profile observed in \figref{fig:vae_roundtrip}.  We therefore use this table as an opaque-scene mechanism diagnostic; transparent-object performance is covered by the main benchmark tables, since RGB edges on refractive objects do not always correspond to true geometric discontinuities.

\subsection{Frequency-Domain Error Analysis}
\label{ssec:frequency_analysis}

\begin{figure}[t]
\centering
\includegraphics[width=0.70\linewidth]{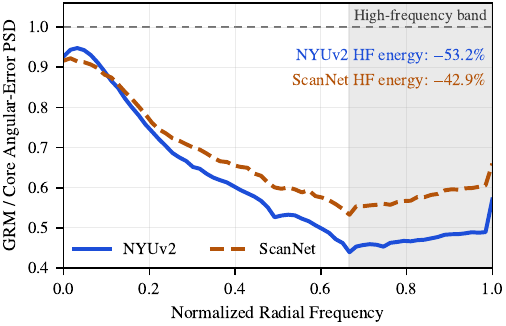}
\caption{\textbf{Frequency-wise ratio of angular-error PSD after and before GRM refinement.}  Values below $1$ indicate lower spatial error energy after refinement.  The shaded region marks the highest third of radial-frequency bins; integrated energy in this band decreases by $53.2\%$ on NYUv2 and $42.9\%$ on ScanNet. (\S~\ref{ssec:frequency_analysis})}
\label{fig:frequency_analysis}
\end{figure}

We further analyze the spatial frequency content of angular-error maps.  For each sample, we compare the core predictor with and without the GRM, subtract the mean error within the valid normal mask, and compute the radially-averaged PSD of the remaining error field.  This PSD is computed on \emph{angular-error maps}, not on the predicted normal maps.  \Figref{fig:frequency_analysis} plots the frequency-wise ratio of the angular-error PSD after and before GRM refinement; the ratio stays below $1$ across the spectrum on both general-scene benchmarks.  The integrated error energy in the high-frequency band (the highest third of radial-frequency bins) decreases by $53.2\%$ on NYUv2 and $42.9\%$ on ScanNet.  This frequency-domain view agrees with the edge/non-edge analysis: the GRM primarily suppresses spatially localized boundary errors rather than uniformly shifting all pixels.

Model complexity, per-method inference speed, a per-component runtime breakdown, and inference-optimization results (including a real-time-FPS configuration) are reported in the Supplementary Material.

\section{Conclusion}
\label{sec:conclusion}

We have identified and quantified a previously overlooked source of systematic error in diffusion-based geometry estimation: VAE reconstruction introduces up to $8.5^\circ$ of angular error even on ground-truth normals, with edge MAE reaching up to $2.8{\times}$ the global MAE.  TransNormal-2 addresses this bottleneck with geometry-aware pixel-space supervision and post-decode correction: vMF, wavelet, and inverse-rendering losses constrain the decoded normal field, while a lightweight RGB-guided GRM reduces residual boundary-localized decoding errors without freely rewriting the coarse prediction.  TransNormal-2 matches or exceeds MoGe-2 on all eight general-scene metrics and achieves especially large margins where boundary precision is critical ($-4.2^\circ$ on ClearGrasp and $-3.1^\circ$ on zero-shot ClearPose over the strongest prior baselines).

\noindent\textbf{Limitations and Future Work.}
The inverse-rendering loss assumes diffuse Lambertian reflectance, so refractive and strongly specular image formation remains only weakly modeled.  In addition, the GRM refines fine-scale geometric structure around boundaries; when the core predictor has already mispredicted a large region, the refinement cannot compensate for it.  Future work may explore material-aware rendering constraints, larger curated normal annotations, and higher-resolution latent spaces for finer geometry supervision.

\bibliographystyle{IEEEtran}
\bibliography{main}

\end{document}